%% file: main-arxiv.tex
\documentclass[10pt,twocolumn,letterpaper]{article}

\usepackage{cvpr}              
\input{preamble}
\definecolor{cvprblue}{rgb}{0.21,0.49,0.74}
\usepackage[pagebackref,breaklinks,colorlinks,allcolors=cvprblue]{hyperref}

\def\paperID{xxxxx} 
\def\confName{CVPR}
\def\confYear{2026}

\title{Generalizable Friction Coefficient Estimation via Material Embedding and Proxy Interaction Modeling}

\author{Zhendong Wang\\
Style3D Research\\
Hangzhou, China\\
{\tt\small wang.zhendong.619@gmail.com}
\and
Huamin Wang\\
Style3D Research\\
Hangzhou, China\\
{\tt\small wanghmin@gmail.com}
}

\begin{document}
\maketitle


\begin{abstract}
    Accurately estimating friction coefficients between arbitrary material pairs is critical for robotics, digital fabrication, and physics-based simulation, but exhaustive pairwise testing scales quadratically with the number of materials. We introduce a proxy-based modeling framework that approximates any pairwise friction $f(A,B)$ from a small, fixed set of proxy materials $C=[c_1,\dots,c_k]$ by learning a per-material embedding $z_A = g(f(A,c1),\dots,f(A,ck))$ and a fusion function $p$ such that $f(A,B)\approx p\big(z_A,z_B\big)$. We present deterministic and probabilistic realizations of $g$ and $p$, procedures for selecting diverse proxy sets, and mechanisms for handling missing or noisy proxy measurements. The learned embeddings are compact, interpretable, and enable calibrated uncertainty estimates for downstream decision making. On simulated and measured friction datasets, our approach achieves high predictive accuracy, robust performance with partial observations, and substantial experimental savings by significantly reducing pairwise testing.
\end{abstract}


\section{Introduction}
\label{sec:intro}

Accurate characterization of inter-material friction is critical to mechanics, robotics, and physically based simulation. Conventional practice requires exhaustive pairwise testing with quadratic cost in the size of the material library, making comprehensive characterization impractical and slowing iteration and deployment, particularly for unseen material pairs. Existing empirical and analytical models generalize poorly beyond measured combinations and typically lack calibrated uncertainty, limiting safe, data-efficient decision making.

The key challenge lies in developing a \emph{scalable and generalizable representation} that predicts friction coefficients \(f(A, B)\) for arbitrary material pairs \(A, B\in\mathcal{M}\) without exhaustive pairwise experiments. Friction depends on complex, nonlinear interactions among surface roughness, elasticity, adhesion, and contact mechanics. Thus, \(f(A, B)\) cannot be reliably inferred from the independent features of \(A\) and \(B\). Naïve mappings from per-material descriptors to pairwise coefficients fail because the relevant interaction terms emerge only through coupled surface behaviors.

We address this challenge by learning \emph{latent friction embeddings} that capture each material's intrinsic frictional characteristics through its interaction profile with other materials. Specifically, a feature extraction function \(g\) maps a material's interaction vector $u_A=[f(A, m_1), \dots, f(A, m_n)]$ to a latent embedding \(z_A=g(u_A)\). A fusion function \(p\) then predicts pairwise friction via $f(A, B)\approx p(z_A, z_B)$. Jointly learning \(g\) and \(p\) constitutes an end-to-end supervised task equivalent to learning a kernel embedding of the friction operator defined over the material space.

Because the full interaction matrix $\mathbf{F}$ are infeasible to measure, we introduce a \emph{proxy-based representation}: each material is characterized by its interactions with a small subset of proxy materials \(C=\{c_1, \dots, c_k\} \), where \( k \ll |\mathcal{M}| \). Empirical analysis of the full friction matrix reveals a low-rank structure, suggesting that frictional behavior can be efficiently approximated within a low-dimensional basis spanned by the proxies. The proxy interaction vector $v_A=[f(A, c_1), \dots, f(A, c_k)]$ thus provides a compact descriptor from which \(g\) learns the embedding \(z_A\), while \(p\) predicts pairwise coefficients from latent embeddings. This formulation reduces the required measurements from \( O(|\mathcal{M}|^2) \) to \(O(|\mathcal{M}|k)\), achieving near-linear scaling while preserving predictive accuracy.

To enhance robustness, \(g\) is implemented as an attention-based encoder capable of handling partial or noisy proxy observations through masking and imputation. For calibrated uncertainty, \(p\) is modeled using Gaussian Processes or Bayesian Neural Networks, providing confidence estimates critical for safe deployment in simulation and robotics. We train \(g\) and \(p\) end-to-end using a supervised metric-learning objective with symmetry and calibration regularization. This yields \emph{compact and transferable} embeddings that generalize to unseen material pairs, \emph{robustness} to incomplete proxy data via attention-based masking, and \emph{calibrated uncertainty} through probabilistic prediction. Our approach serves as a scalable surrogate model for friction, converting quadratic experimental complexity into a linear proxy-based process while maintaining interpretability through proxy contribution analysis. Beyond reducing experimental cost, this framework provides interpretable insights into the latent structure of frictional interactions, enabling informed material selection and predictive modeling across robotics, fabrication, and physically based simulation. 


\section{Related Work}
Friction coefficient measurement spans classical tribology to modern robotics. Canonical benchtop methods include inclined-plane tests for static friction (originating with Coulomb) and instrumented tribometers such as sled/drag (ASTM D1894, ISO 8295), pin-on-disk (ASTM G99), and reciprocating rigs (ASTM G133), enabling sweeps of load, speed, temperature, and humidity to probe Coulomb and Stribeck regimes \cite{Coulomb1785,Stribeck1902,ASTM_D1894,ISO_8295,ASTM_G99,ASTM_G133}. At micro/nano scales, friction force microscopy and nano-tribometers interrogate adhesion, plowing, and interfacial chemistry, complemented by surface profilometry and spectroscopy \cite{Carpick1997,Bhushan2013,Mate2008}. Despite standardization, measurements remain protocol- and surface-preparation dependent, complicating cross-material generalization and making exhaustive pairwise testing expensive \cite{Bhushan2013}.

Robotics-oriented estimation performs in situ identification using force–torque sensing, tactile skins, and vision-based tactile sensors, often via push–slide or incipient-motion experiments \cite{Mason2001,Goyal1991,Zhou2016}. Learning-based slip detection and contact-state classification on tactile images/signals (CNN/RNN) provide robust online indicators but are sensitive to contact conditions and yield device/task-specific coefficients \cite{Yuan2017,Donlon2018,Lambeta2020,Calandra2018}. While such approaches reduce per-material calibration, they do not address the combinatorial growth of pairwise measurements across large libraries.

Data-driven modeling links measurements to descriptors or exploits latent structure. Supervised regressors map composition, microstructure, or roughness statistics to friction-related properties, borrowing tools from materials informatics \cite{Ward2016,Butler2018}. Pairwise friction can be cast as a partially observed matrix and imputed via low-rank factorization or collaborative filtering, improving sample efficiency when the interaction matrix is near low rank \cite{Koren2009,Candes2009,Mazumder2010}. Probabilistic predictors—Gaussian processes, Bayesian neural networks, and deep ensembles—provide calibrated uncertainty to guide experiment design and safe deployment \cite{Rasmussen2006,Blundell2015,Lakshminarayanan2017,Guo2017}. However, models relying only on independent per-material features may miss emergent pair-dependent interactions, and full matrix-completion still assumes access to scattered pairwise labels at scale.

AI-based proxy strategies sidestep quadratic scaling by characterizing each material through interactions with a small, fixed set of references, then learning latent embeddings and a symmetric fusion for pairwise prediction—conceptually related to relational/knowledge-graph embeddings and metric learning \cite{Nickel2016,Bordes2013}. Attention-based encoders naturally handle partial proxy observations via masking \cite{Vaswani2017}. Active proxy selection using diversity and submodular objectives improves coverage of frictional regimes \cite{Krause2014,Kulesza2012}. 

Positioned within this landscape, our method unifies: (i) classical measurement principles via symmetry and range constraints, (ii) low-rank/relational structure through proxy-induced embeddings, and (iii) probabilistic prediction for calibrated uncertainty. This yields generalizable friction estimates for unseen material pairs while reducing measurements from $O(|\mathcal{M}|^2)$ to $O(|\mathcal{M}|k)$ and preserving interpretability via proxy contribution analysis.


\section{Problem Definition}

Let $\mathcal{M}$ be the set of materials with size of $|\mathcal{M}|=n$. For any $A\in\mathcal{M}$, define its interaction vector $u_A\in\mathbb{R}^n$ with entries $[u_A]_i=f(A,m_i)$, observed under a binary mask $s_A\in\{0,1\}^n$ indicating missing measurements. We seek a feature extractor $g:\mathbb{R}^n\times\{0,1\}^n\to\mathbb{R}^d$ and a symmetric fusion function $p:\mathbb{R}^d\times\mathbb{R}^d\to\mathbb{R}_{\ge 0}$ such that \[f(A,B)\approx p\big(z_A,z_B\big),\] where $z=g(u,s)\in\mathbb{R}^d$ represents friction-related latent features, approximates $f(A,B)$ for arbitrary pair $A,B\in\mathcal{M}$. To reduce the measurement cost from $O(n^2)$ to $O(nk)$, we also need to find a proxy set $C=\{c_1,\dots,c_k\}$ with $k\ll n$ and define the proxy interaction vector $v_A\in\mathbb{R}^k$ with entries $[v_A]_i=f(A,c_i)$. The feature extractor then becomes $g_c:\mathbb{R}^k\to\mathbb{R}^d$ with $g_c(v_A)\approx g(u_A,s_A)$. The fusion function $p$ should satisfy the symmetry property $p(a,b)=p(b,a)$ for any $a,b\in\mathbb{R}^d$ and adapt to both proxy and non-proxy interactions.


\section{Friction Embedding Model}

Given the complete friction coefficient matrix $\mathbf{F}\in\mathbb{R}^{n\times n}$ with entries $F_{AB}=f(A,B)$ for all $A,B\in\mathcal{M}$, define for each material $A$ its all-material interaction vector $u_A=[f(A,m_1),\dots,f(A,m_n)]$. A feature extractor $g$ maps $u_A$ to a latent embedding $z_A=g(u_A)$. A natural baseline is to learn $g$ via an autoencoder that jointly trains an encoder and decoder to minimize reconstruction error:
\begin{equation}
    \begin{aligned}
    & \min \sum_{A\in\mathcal{M}}\|u_A-\hat{u}_A\|_2^2,\\
    & z_A=\mathrm{Encoder}(u_A),\quad \hat{u}_A=\mathrm{Decoder}(z_A).
    \end{aligned}
\end{equation}
After training, the encoder is used as $g$ to produce embeddings for any material $A$. However, because $u_A$ aggregates interactions with all other materials, the resulting $z_A$ tends to entangle pair-specific effects rather than isolate intrinsic, friction-relevant properties. As a consequence, these embeddings (i) do not generalize reliably to unseen material pairs and (ii) are not recoverable from a small set of proxy interactions alone. Training the fusion function $p$ directly on such embeddings therefore often exhibits poor generalization, a behavior we observe empirically. 

To overcome these shortcomings, we jointly learn the feature extractor $g$ and fusion function $p$ directly from observed pairwise friction coefficients. The encoder $g$ maps each interaction vector $u_A$ to a latent representation $z_A=g(u_A)$. Rather than coupling $g$ with a decoder that reconstructs $u_A$, we supervise $g$ solely through the pairwise prediction objective by training $p$ to regress $f(A,B)$ from $(z_A,z_B)$. This end-to-end formulation aligns the embedding space with the downstream task, encouraging $z_A$ to capture intrinsic, friction-relevant structure that transfers to unseen material pairs and measurement regimes.

\subsection{Model Architecture}

\paragraph{Feature Extraction}

The feature extractor $g$ maps interaction vectors $u_A\in\mathbb{R}^n$ to latent embeddings $z_A\in\mathbb{R}^d$. We consider nonlinear encoding via MLP and Transformer attention. MLP encodes nonlinear relationships between material pair interactions. However, in the friction matrix $\mathbf{F}$, there are probably some missing values. To address this, we take advantages of transformer-style encoders to incorporate attention masking, enabling robustness to missing measurements. Therefore, we implement $z_A=g(u_A,m_A)$ as a Transformer encoder \cite{Vaswani2017} with $L$ layers and $H$ attention heads. Each input interaction vector $u_A$ is first embedded into a sequence of token vectors via a linear projection and positional encoding. The Transformer layers then apply multi-head self-attention with masking $m_A$ to handle missing entries, followed by feedforward networks and layer normalization. The final embedding $z_A$ is obtained by pooling the output token representations (e.g., via mean or max pooling). During training, we simulate missing data by randomly masking entries in the interaction vectors, enhancing the model's robustness to real-world scenarios where measurements may be incomplete.

\paragraph{Feature Fusion}

The fusion function $p$ serves as a predictive mapping from pairs of latent embeddings to a scalar friction estimate. A practical realization is an MLP applied to a symmetry-preserving pairwise feature map, e.g., 
\[p(z_A,z_B)=h\big([\,z_A{+}z_B,\ |z_A{-}z_B|,\ z_A\odot z_B\,]\big),\] which enforces $p(z_A,z_B)=p(z_B,z_A)$ by construction. To respect physical bounds, the network output is passed through a squashing function and scaled to the admissible range, e.g., $\hat f=\mu_{\max}\,\sigma(\cdot)$, ensuring nonnegativity and $0\le \hat f(A,B)\le \mu_{\max}$.


\paragraph{The Loss Objective}

The training objective is:
\begin{equation}
\mathcal{L}
= \alpha\sum_{(A,B)\in\mathcal{M}} \bigl(f_{AB}-\hat f_{AB}\bigr)^2
+ \beta\sum_{A\in \mathcal{M}} \left\|u_A-\phi(z_A)\right\|_2^2,
\label{eq:loss}
\end{equation}
where $z_A=g(u_A)$, $\hat f_{AB}=p(z_A,z_B)$, and $\phi:\mathbb{R}^d\!\to\!\mathbb{R}^k$ is an optional decoder promoting embedding informativeness via input‑space reconstruction. The first term enforces pairwise predictive fidelity; the second regularizes the latent space and mitigates degenerate solutions. Coefficients $\alpha,\beta>0$ satisfy $\alpha+\beta=1$ and are selected by validation.

To enhance embedding robustness against missing values and noise, we apply input corruption during training and enforce decoder consistency. Mini‑batch corruption can be obtained by sample mask $m_A$ and additive proxy noise $\eta\sim\mathcal{N}(0,\sigma_v^2 I)$, $v_A^{\text{noisy}}=(v_A+\eta)\odot m_A,\quad z_A=g\big(v_A^{\text{noisy}},m_A\big)$, decoder consistency penalty:
\[
\mathcal{L}_{\text{latent}}=\sum_A \| \phi(z_A)-v_A\|_1.
\]

For the probabilistic fusion module $p$, let the predictive mean and variance be $\mu_{AB}$ and $\sigma_{AB}^2$, respectively, the negative log-likelihood is
\[
\mathcal{L}_{\text{NLL}} = \tfrac12 \sum_{(A,B)\in\mathcal{S}} w_{AB}\left(r_{AB}^2/\hat\sigma_{AB}^2 + \log \hat\sigma_{AB}^2 \right),
\]
with residual $r_{AB}\!=\!\tilde f_{AB}\!-\!\mu_{AB}$ and residual‑adaptive weights $w_{AB}=\bigl(1+\exp\!\big(\kappa\,(|r_{AB}|-\tau)\big)\bigr)^{-1}$ to mitigate the influence of outliers. $\kappa>0$ controls the steepness, and $\tau$ is a running quantile (e.g., 90th) of $|r_{AB}|$ computed per epoch or mini‑batch. This scheme downweights large‑residual samples while leaving inliers near unit weight. This heteroscedastic formulation adaptively attenuates the influence of high-noise measurements by allocating larger predictive variance, improving robustness to experimental uncertainty.

The noise-aware training objective is:
\begin{equation}
\mathcal{L}_{\text{noise}}
=\mathcal{L}+\gamma\mathcal{L}_{\text{NLL}}+\theta\mathcal{L}_{\text{latent}},
\end{equation}
where $\gamma,\theta>0$ weight the heteroscedastic negative log-likelihood and the latent regularizer. This objective preserves aleatoric uncertainty, attenuates outlier influence, and supports active acquisition by prioritizing pairs with large epistemic variance (e.g., high $\sigma^{2}_{\mathrm{epi}}$ when using ensembles), thereby improving reliability under sparse, noisy measurements.


\section{Proxy Materials Selection}
\label{sec:proxy_selection}

Practical deployment hinges on selecting an informative and feasible proxy set $C$. Proxies should span diverse frictional regimes while remaining straightforward to measure. The full friction matrix $\mathbf{F}$ is empirically low rank, making exhaustive pairwise acquisition redundant and costly. A compact, well‑conditioned subset $C$ supplies an approximate basis for the interaction space, enabling accurate pairwise prediction from $O(nk)$ measurements and substantially reducing experimental effort.

Given an arbitrary material $A$, its embedding $z_A$ can be extracted by $z_A=g(u_A)$ where $u_A=[f(A,m_1),\dots,f(A,m_n)]$ is the interaction vector with all materials. The effect of the proxy materials can be interpreted as projecting $u_A$ onto the subspace spanned by the proxies, capturing the most relevant features for friction prediction, i.e. $z'_A=g(v_A)\approx z_A=g(u_A)$.

\subsection{Friction Matrix Analysis}

The friction matrix $\mathbf{F}$ is empirically low rank, enabling compact representation and efficient computation. We assess low-rank structure by inspecting the spectral decay. Under the symmetry assumption $f(A,B)=f(B,A)$, we use the eigendecomposition $\mathbf{F}=\mathbf{U}\boldsymbol{\Lambda}\mathbf{U}^\top$ with $\boldsymbol{\Lambda}=\operatorname{diag}(\lambda_1,\dots,\lambda_n)$, $\lambda_1\geq\cdots\geq\lambda_n\geq 0$ (coinciding with the singular values in this case). The decay of $\{\lambda_i\}$ quantifies the intrinsic dimensionality of the interaction space. Given a relative threshold $\tau$ (e.g., $1\%$ of $\lambda_1$), the effective rank is $r=\bigl|\{\,i:\lambda_i\geq \tau\lambda_1\,\}\bigr|$. In our experiments, the spectrum decays rapidly, with approximately $k\approx 10$ dominant modes capturing the majority of variance. This observation justifies proxy-based modeling: a small, well-chosen subset of materials can span the column space of $\mathbf{F}$, enabling accurate prediction from $O(nk)$ measurements while maintaining predictive performance. Standard matrix factorizations (low-rank approximations) and, when extending to multi-condition data, tensor decompositions, further expose the latent factors governing frictional behavior.

Guided by the spectral structure of $\mathbf{F}$, we select the proxy set $C$ via column–subset selection using rank‑revealing QR (RRQR). This procedure approximately maximizes the volume of $\mathbf{F}_{:,C}$ (max‑vol), yielding a well‑conditioned basis that spans the interaction space and increases the informativeness of proxy vectors $v_A$. To better align with the embedding–fusion pipeline, we subsequently refine $C$ to preserve embeddings, as detailed next. The proxy budget $k$ trades off informativeness against experimental cost. We therefore pose a bi‑criteria objective that (i) maximizes coverage of frictional behaviors and (ii) minimizes $k$, such as,
\[
\min_{k} \ k \quad \text{s.t.}\quad \sum_{A\in\mathcal{M}}\big\|g(u_A)-g(u_A\odot m)\big\|_2^2\ge \tau,
\]
where $\tau$ is a target alignment threshold, and the constraint quantifies the agreement between proxy‑restricted and full embeddings.

\subsection{Proxy Selection with Embedding Preservation}

The proxy selection strategy based on RRQR focuses on spanning the friction interaction space but does not directly consider the learned embedding space defined by $g$ and $p$. To address this, we propose to refine the proxy set $C$ by explicitly minimizing the discrepancy between full‑context embeddings $g(u_A)$ and proxy‑restricted embeddings $g_c(v_A)$. This can be formulated as seeking a proxy mask with minimal observation:
\[\min_{m}\ \operatorname{Sum}(m) \quad \text{s.t.}\quad \sum_{A\in\mathcal{M}}\big\|g(u_A)-g(u_A\odot m)\big\|_2^2\]
where $m\in\{0,1\}^n$ is a binary mask indicating selected proxies.

\section{Dataset and Experimental Setup}
\label{sec:dataset}

The primary dataset comprises friction measurements acquired using a custom tribometer across a library of over 50 materials, encompassing metals, polymers, composites, and a broad range of textile fabrics; textiles constitute the majority of specimens. Friction measurements are conducted under standardized conditions, with each material pair tested multiple times to account for variability. 

\subsection{Measurement Device}

We designed a custom tribometer for high-throughput friction characterization across a diverse material library. The apparatus comprises a slider block, optionally wrapped with textile specimens, that traverses a standardized grooved counterface with a $150^{\circ}$ included angle. The counterface can be covered with textile or non-textile materials to provide interchangeable counterparts. One end of the groove is actuated to form an adjustable incline; the slope angle is set and logged via an electronically controlled actuator. A pair of laser gates defines a measurement section along the groove, and the block’s transit time between gates is recorded for kinematic estimation. The mechanical layout and quick-release fixtures enable rapid sample exchange, supporting efficient acquisition of both proxy and pairwise friction measurements.

\begin{figure}[t]
    \centering
    \includegraphics[height=0.55\linewidth]{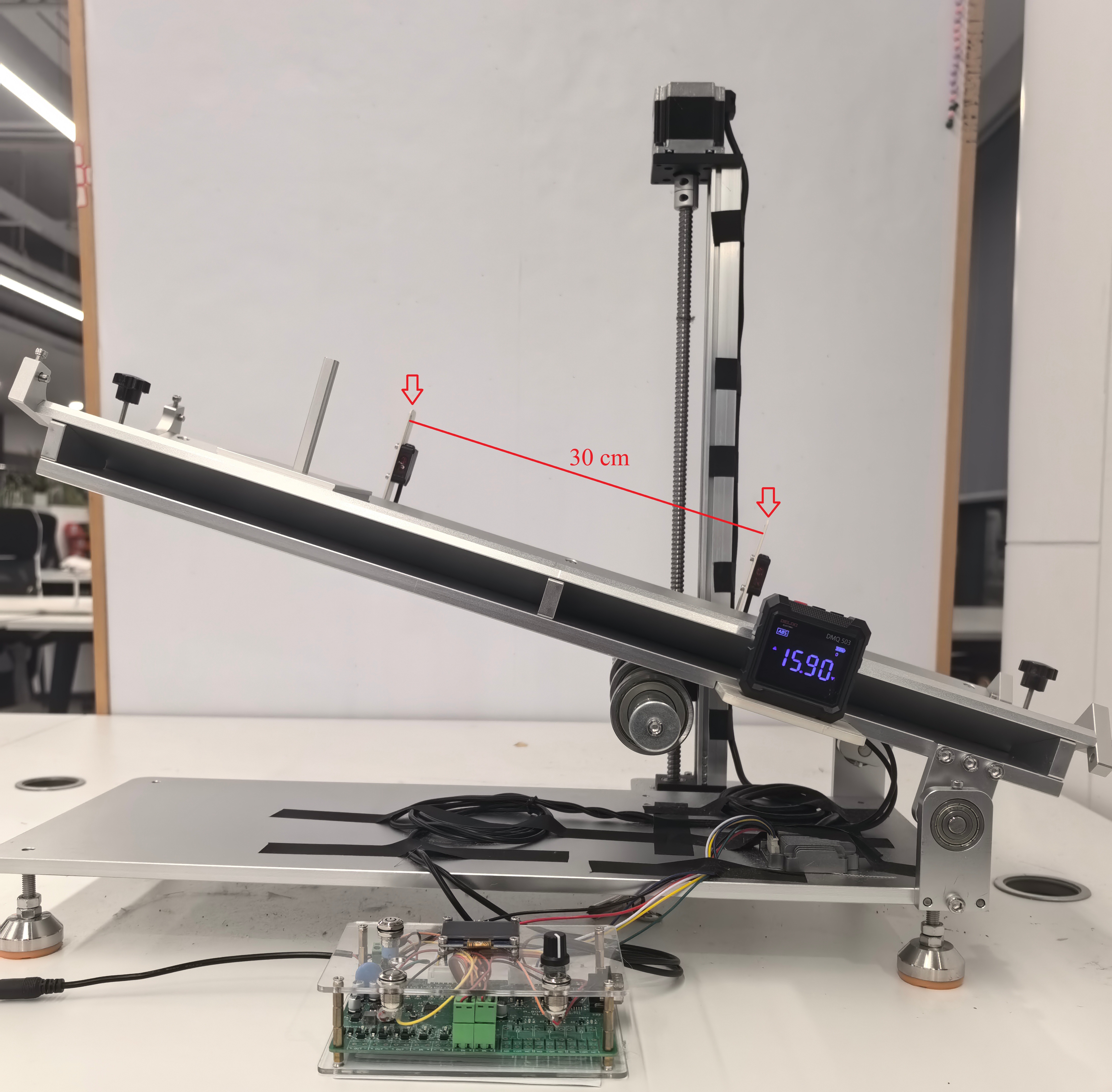}
    \includegraphics[height=0.55\linewidth]{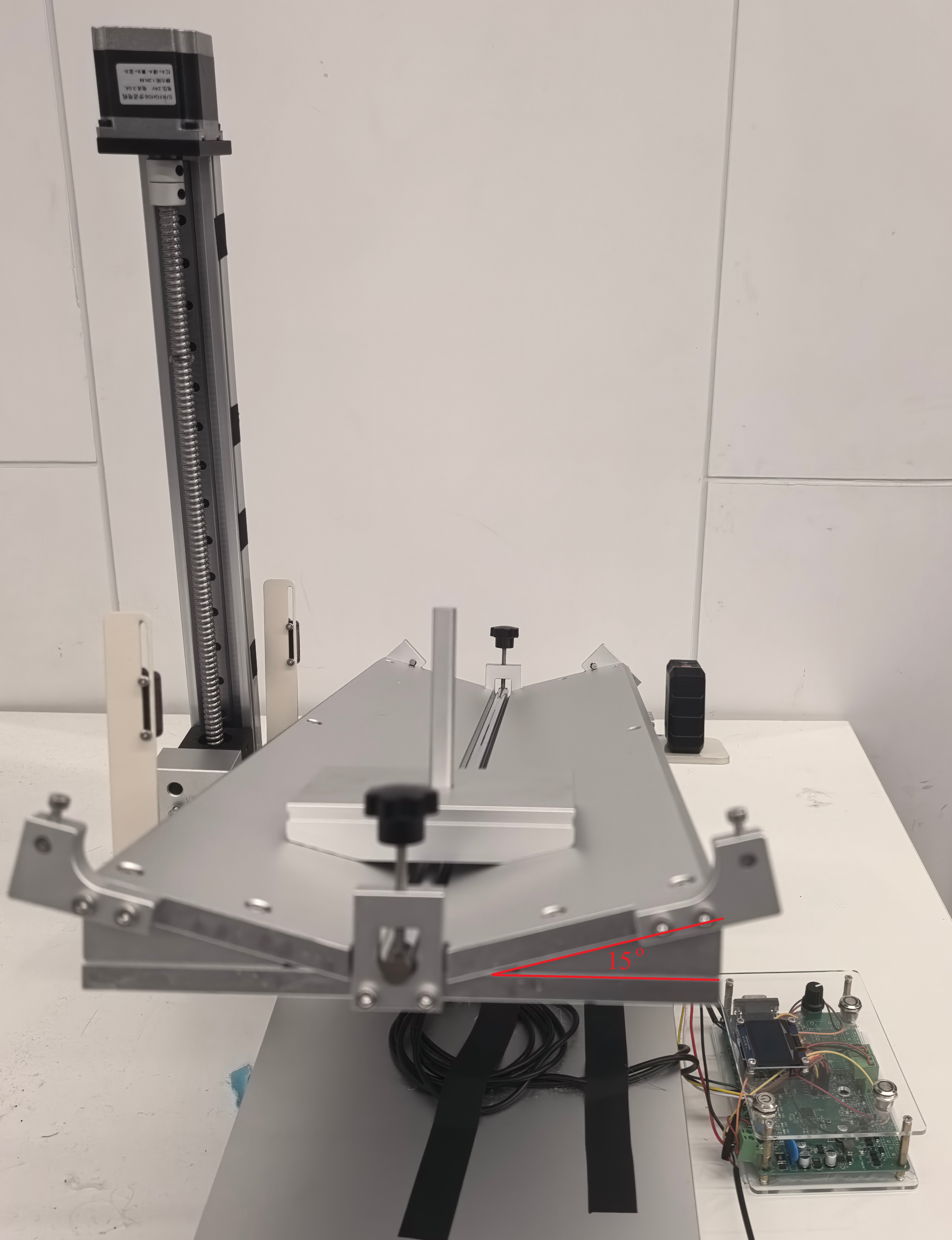}
    \caption{The custom tribometer used for friction measurements. The apparatus features an adjustable inclined groove, interchangeable counterface materials, and laser gates, as marked by red arrows in the left image for velocity measurement, enabling systematic evaluation of static and kinetic friction coefficients across a diverse material library.}
    \label{fig:tribometer}
\end{figure}

\subsection{Friction Measurement Protocols}

\paragraph{Static Friction Measurement}
When a block wrapped with material \(A\) is placed on an inclined plane covered with material \(B\), the static friction coefficient \(\mu_s\) is determined by gradually increasing the incline angle \(\theta\) until the block initiates motion. The coefficient is computed as \(\mu_s = \tan(\theta)\) at the onset of sliding.

\paragraph{Kinetic Friction Measurement}
When a block wrapped with material \(A\) slides down an inclined plane covered with material \(B\) at a constant speed, the kinetic friction coefficient \(\mu_k\) is calculated using the formula \(\mu_k = \frac{F_f}{N}\), where \(F_f\) is the measured friction force during sliding and \(N\) is the normal force acting on the block.

In our method, the target friction observable $f(A,B)$ may be instantiated as either the static coefficient $\mu_s(A,B)$ (incipient motion) or the kinetic coefficient $\mu_k(A,B)$ (steady sliding), selected according to the operational regime. Alternatively, a vector-valued formulation $f(A,B) = [\mu_s(A,B),\,\mu_k(A,B)]$ enables multi-task learning with a shared encoder $g$ and task-specific fusion heads $p_s, p_k$. This joint modeling exploits inductive transfer between regimes, permits enforcement of physically motivated constraints (e.g., $\mu_s(A,B)\ge \mu_k(A,B)$), and can improve sample efficiency under sparse measurements.

\subsection{Anisotropy Handling}
To account for the anisotropic nature of textile materials, we employ a systematic approach to measure friction in multiple orientations. Each textile material is tested with respect to its weft and warp directions, and the resulting friction coefficients are recorded as functions of the sliding direction. This allows us to capture the full frictional behavior of materials and incorporate this information into our modeling framework.


\section{Results}

\subsection{Implementation Details}

The implementation uses PyTorch with GPU acceleration. The feature extractor \(g\) is a 4‑layer Transformer encoder (4 attention heads, masked self‑attention for incomplete interaction vectors). The fusion module \(p\) is a symmetry-preserving MLP operating on concatenated pairwise features to produce a scalar friction estimate. Optimization uses Adam (learning rate \(1\times10^{-3}\), batch size 32) with early stopping on validation loss to mitigate overfitting.

\subsection{Training Evaluation}

\begin{figure}[t]
    \centering
    \includegraphics[width=\linewidth]{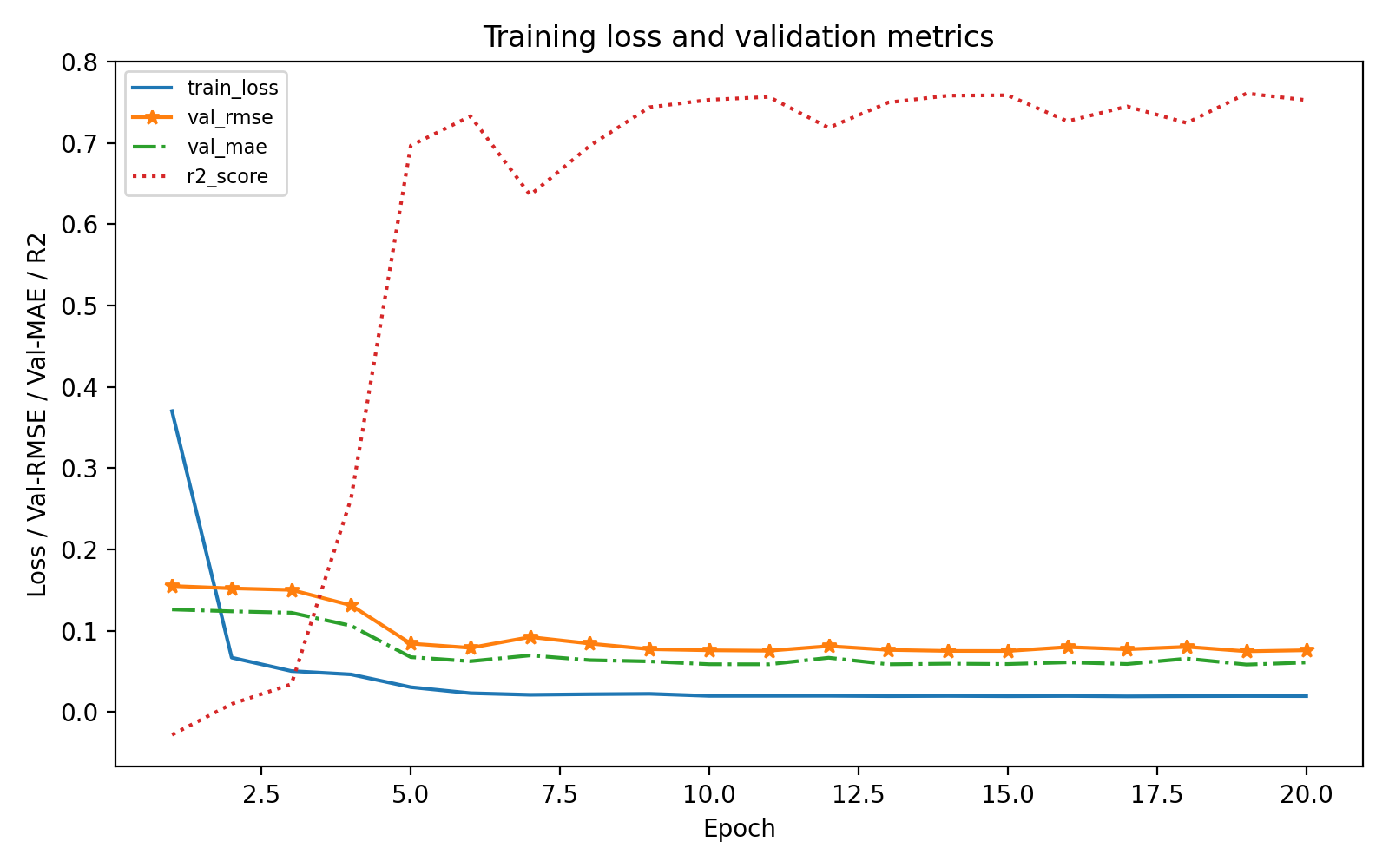}
    \caption{We show training and validation loss curves over epochs for one cross-validation fold. The model converges within 100 epochs, with validation loss closely tracking training loss, indicating effective generalization without overfitting.}
    \label{fig:loss_curves}
\end{figure}

\begin{figure*}
    \centering
    \begin{subfigure}[t]{0.24\linewidth}
        \centering
        \includegraphics[width=\linewidth]{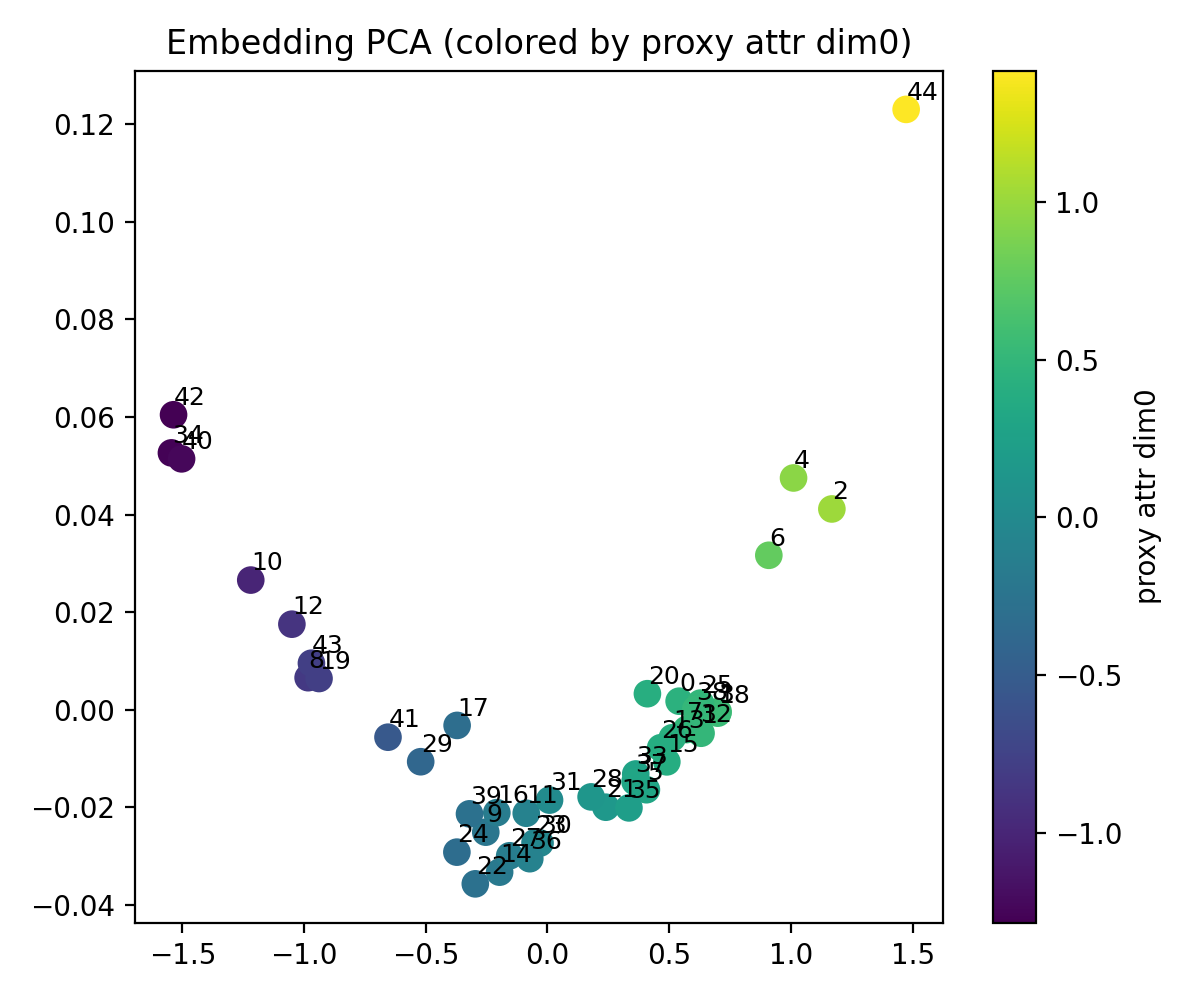}
        \subcaption{PCA embedding}
        \label{fig:emb_pca}
    \end{subfigure}
    \hfill
    \begin{subfigure}[t]{0.24\linewidth}
        \centering
        \includegraphics[width=\linewidth]{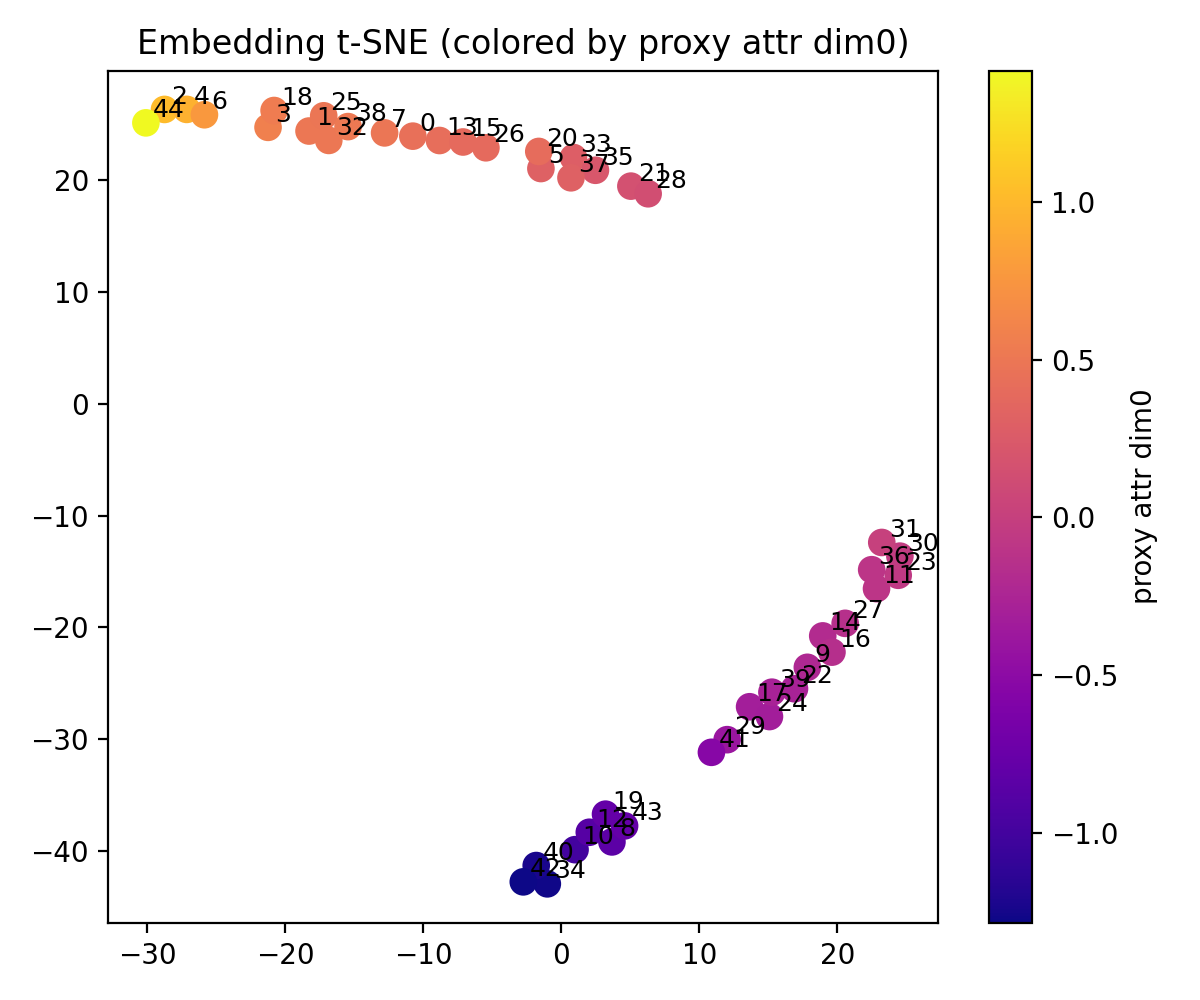}
        \subcaption{t-SNE embedding}
        \label{fig:emb_tsne}
    \end{subfigure}
    \hfill
    \begin{subfigure}[t]{0.2\linewidth}
        \centering
        \includegraphics[width=\linewidth]{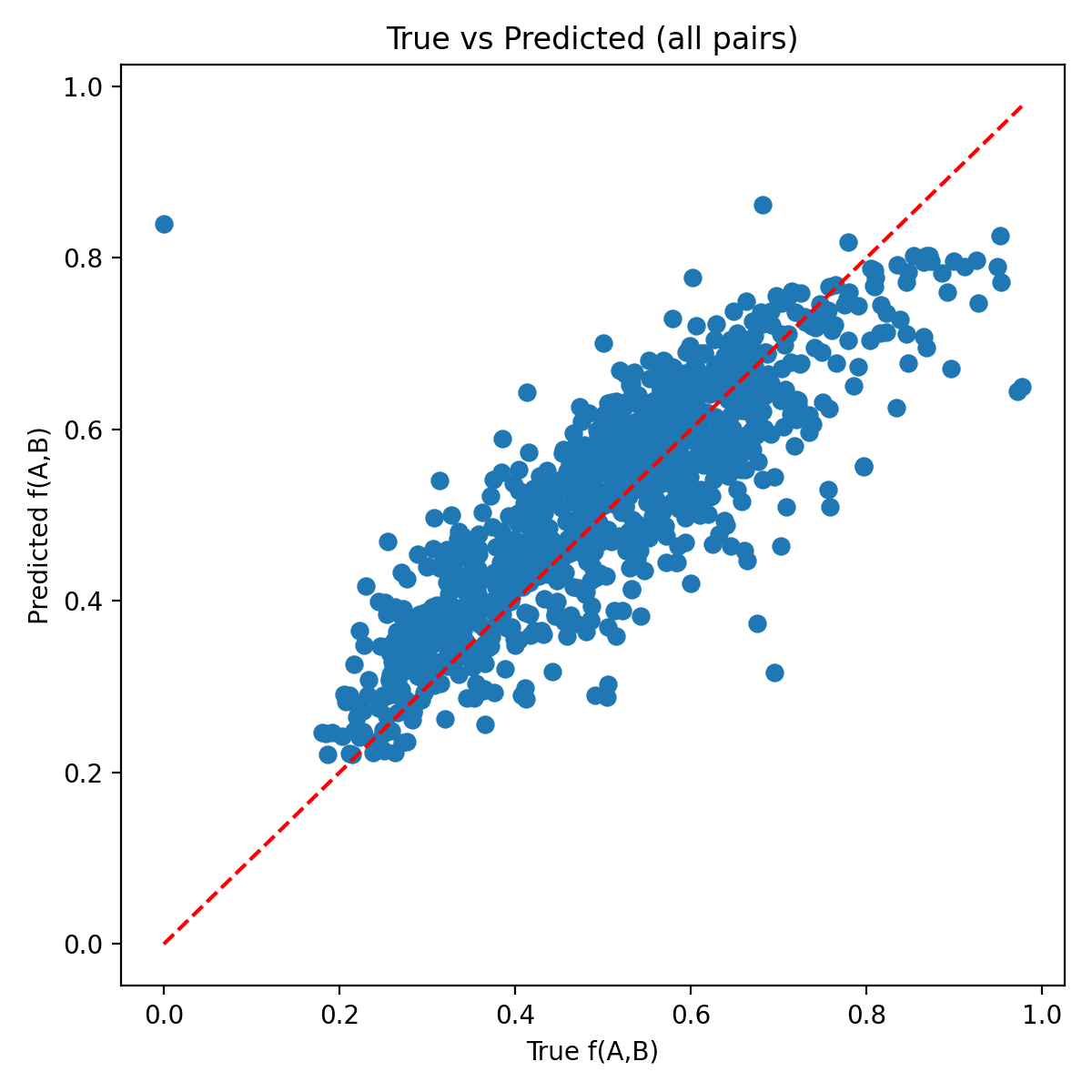}
        \subcaption{True vs. predicted}
        \label{fig:true_vs_pred}
    \end{subfigure}
    \begin{subfigure}[t]{0.3\linewidth}
        \centering
        \includegraphics[width=\linewidth]{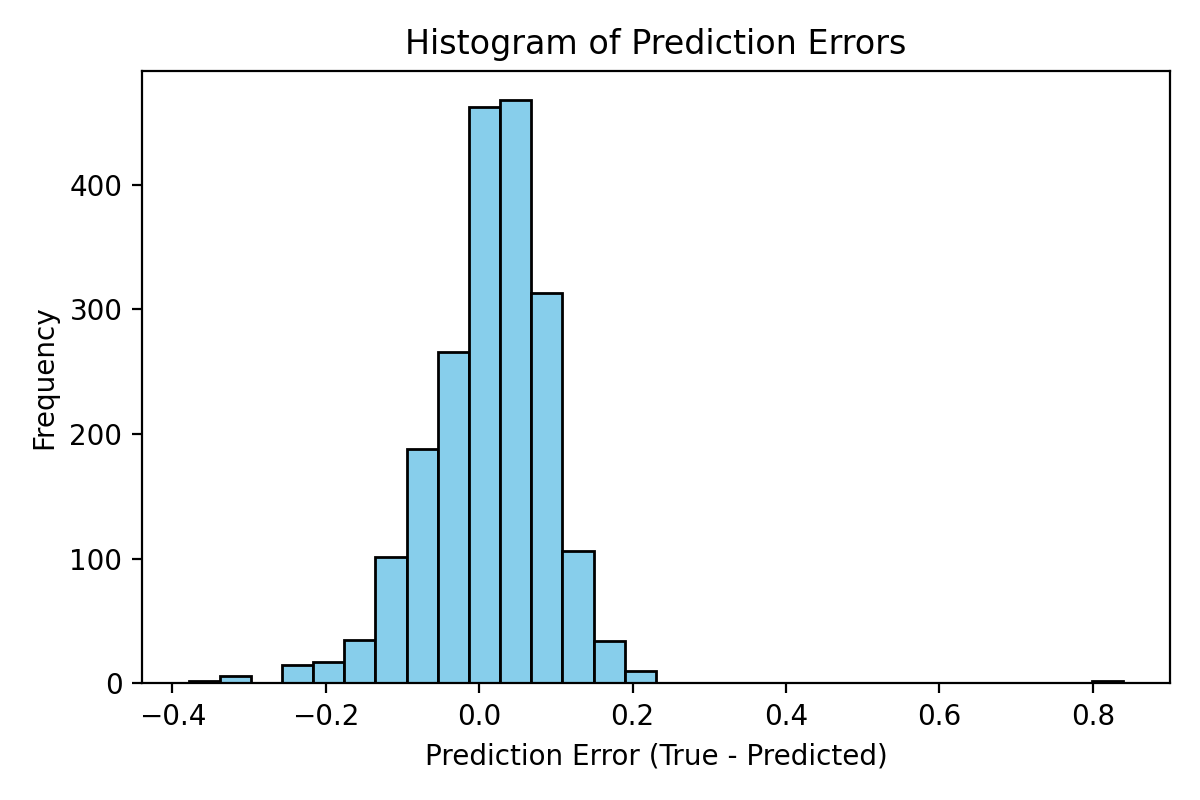}
        \subcaption{Error histogram}
        \label{fig:error_histogram}
    \end{subfigure}
    \caption{(a) PCA and (b) t-SNE visualizations of the learned latent embedding space, revealing clusters corresponding to different frictional regimes. (c) Scatter plot of true vs. predicted friction coefficients on the test set, demonstrating high predictive accuracy with points closely aligned along the diagonal. (d) Histogram of prediction errors, showing a concentrated distribution around zero, indicating low bias and variance in predictions.}
    \label{fig:embeddings_and_predictions}
\end{figure*}

Model performance is evaluated by partitioning the dataset into disjoint training, validation, and test splits, ensuring that all material pairs in the test split are unseen during training. We allocate $70\%/15\%/15\%$ to training/validation/testing and further assess robustness via K-fold cross-validation. As shown in Figure~\ref{fig:loss_curves}, the model converges within 20 epochs; the validation loss closely tracks the training loss and the coefficient of determination ($R^2$) approaches 1, indicating effective generalization without overfitting. The learned latent space can be further analyzed to identify meaningful subspaces corresponding to different friction regimes or material properties. Techniques such as t-SNE or PCA can be employed to visualize the embedding space and uncover underlying structures. As shown in Figure~\ref{fig:embeddings_and_predictions}, PCA and t-SNE visualizations of the learned latent embedding space reveal clusters corresponding to different frictional regimes. Additionally, the scatter plot of true vs. predicted friction coefficients on the test set demonstrates high predictive accuracy, with points closely aligned along the diagonal.

\subsection{Proxy-Based Evaluation}

\begin{figure*}
    \centering
    \begin{subfigure}[t]{0.24\linewidth}
        \centering
        \includegraphics[width=\linewidth]{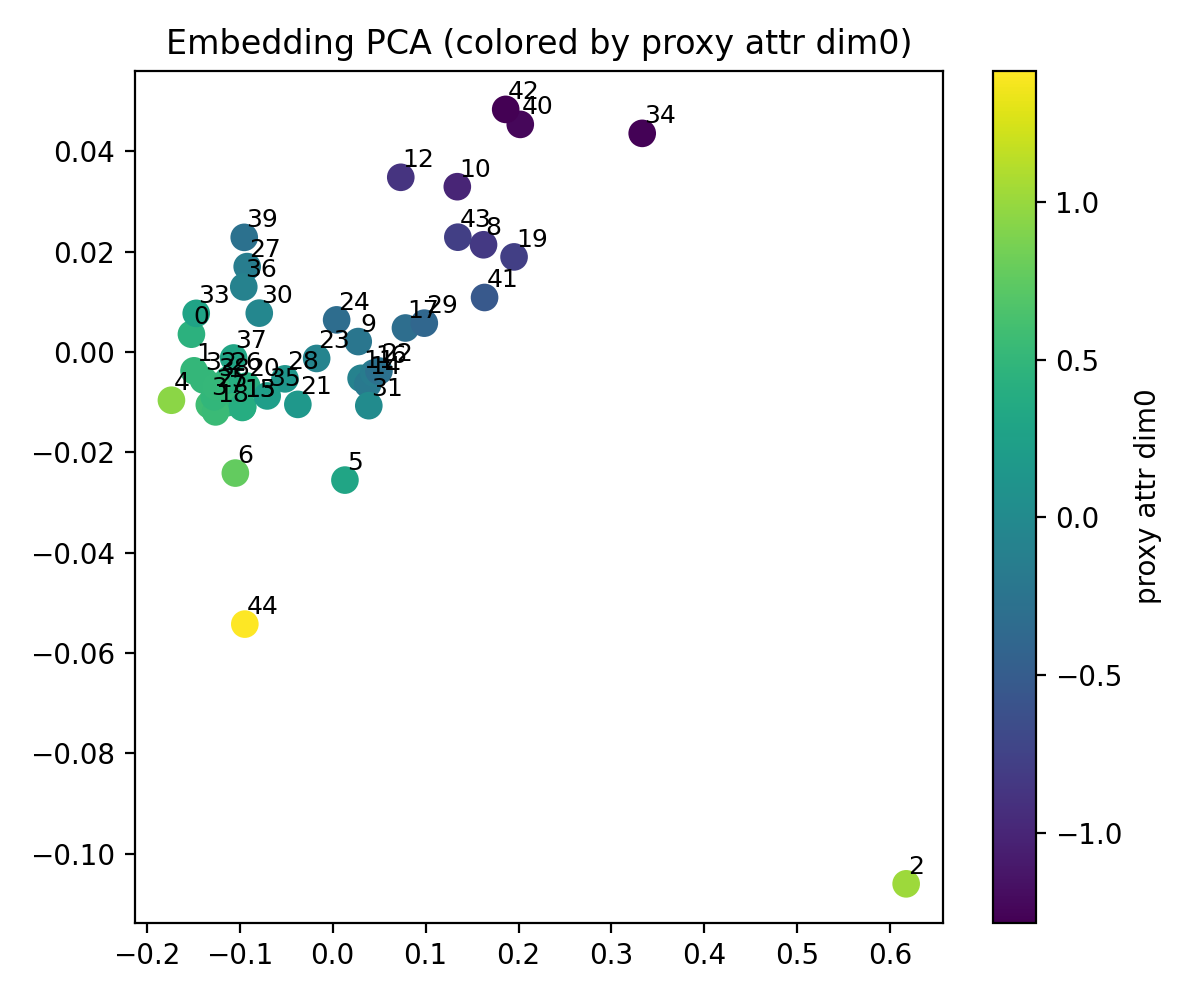}
        \includegraphics[width=\linewidth]{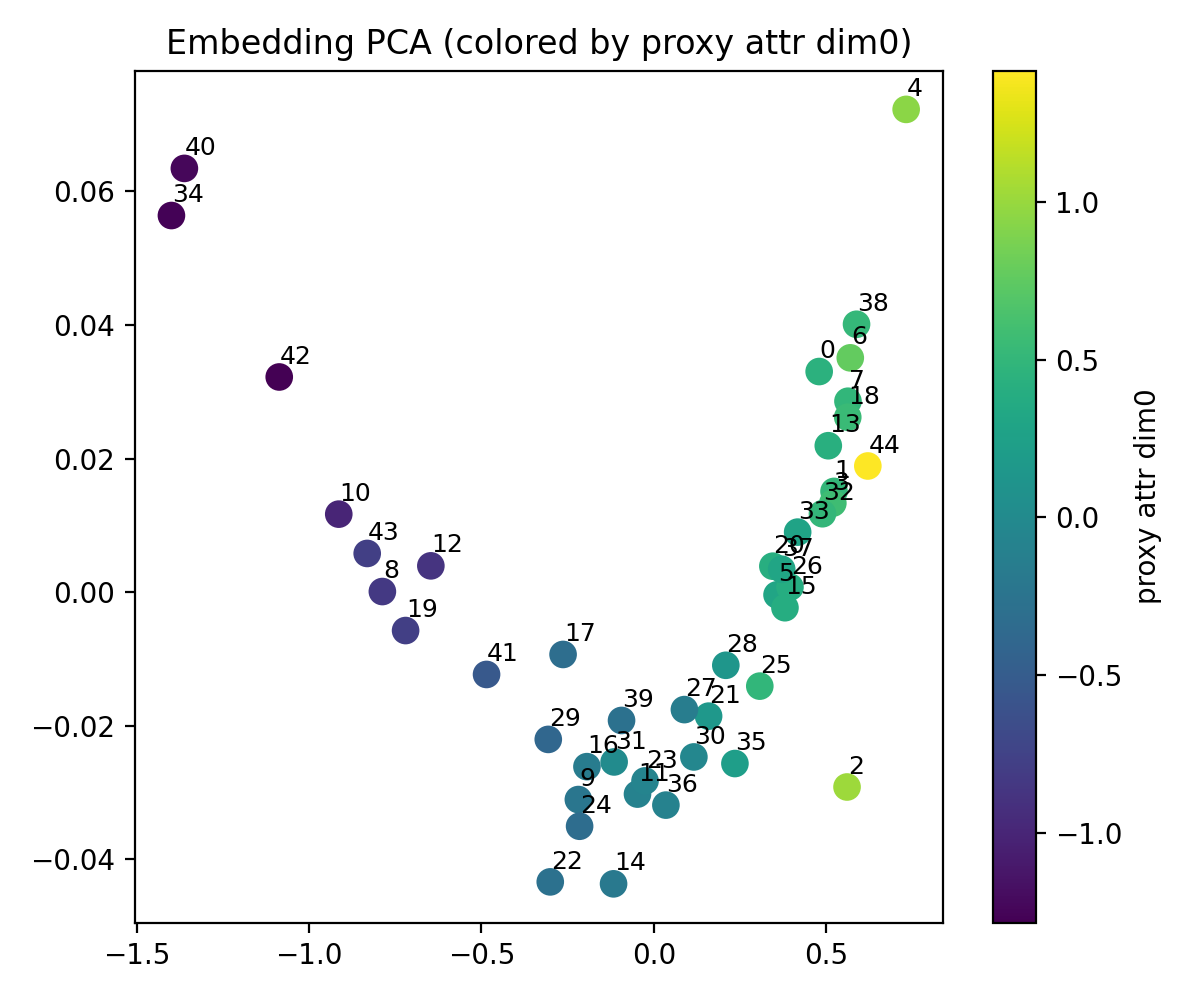}
        \includegraphics[width=\linewidth]{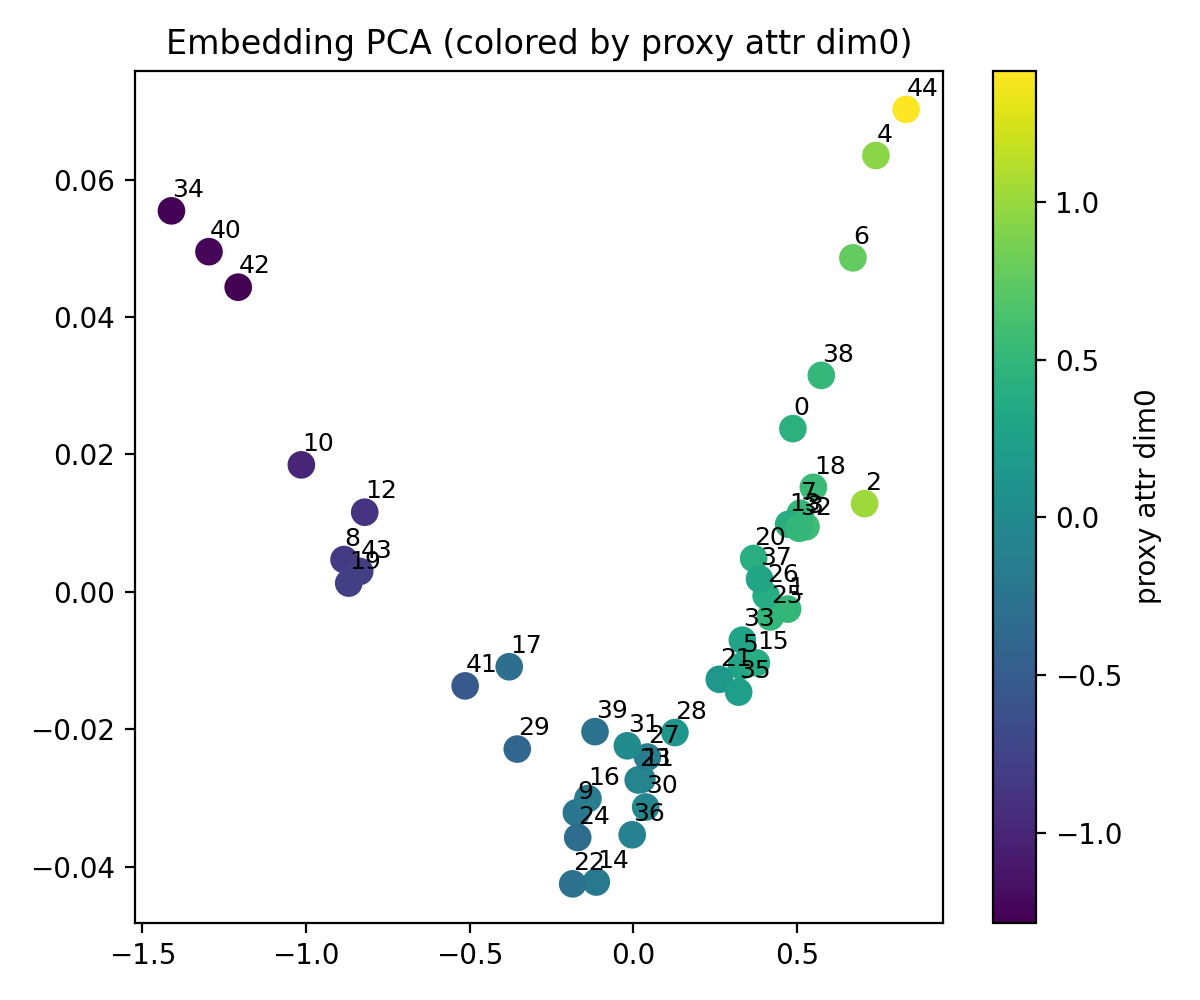}
        \includegraphics[width=\linewidth]{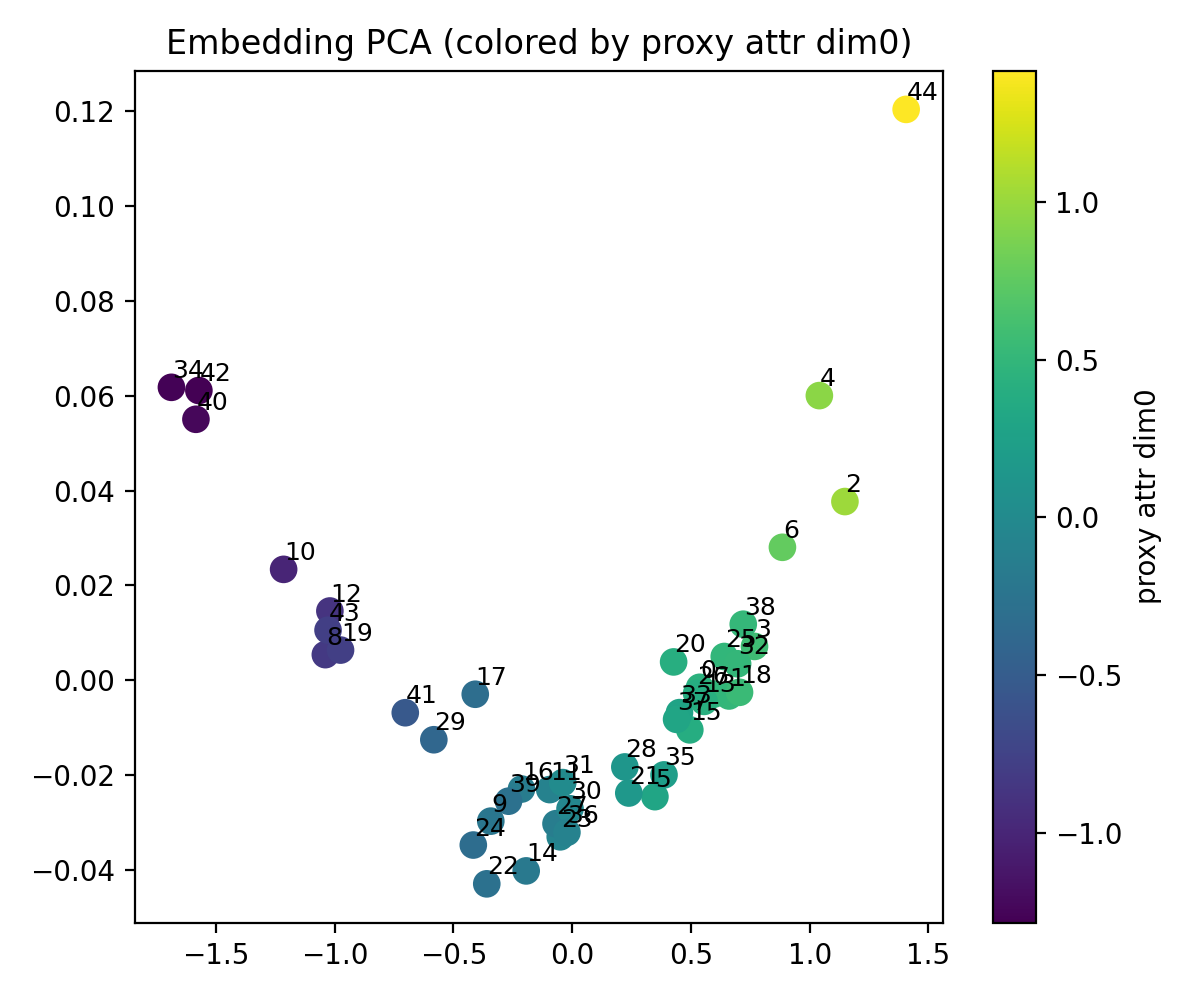}
        \subcaption{PCA embedding}
        \label{fig:emb_pca}
    \end{subfigure}
    \hfill
    \begin{subfigure}[t]{0.24\linewidth}
        \centering
        \includegraphics[width=\linewidth]{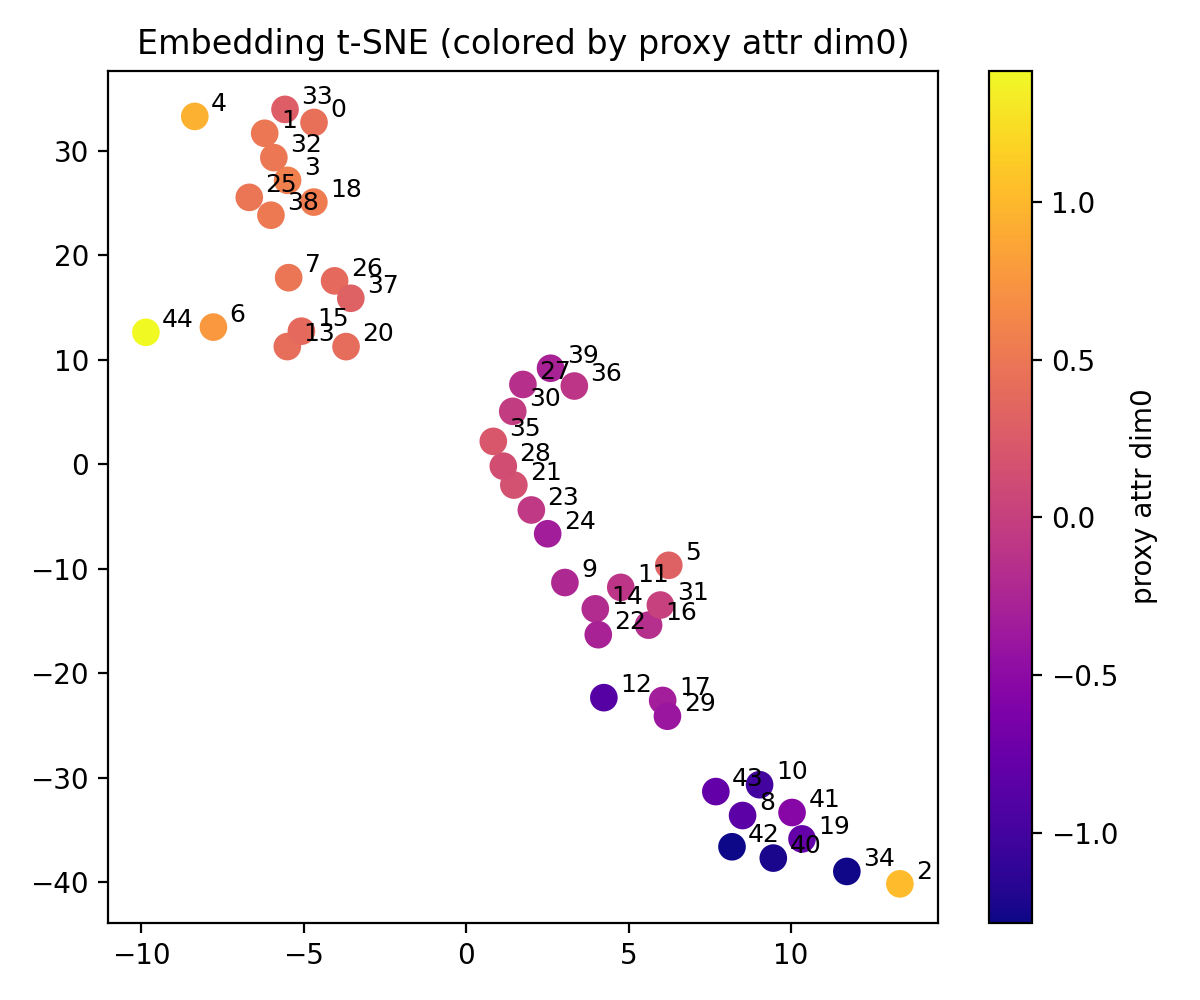}
        \includegraphics[width=\linewidth]{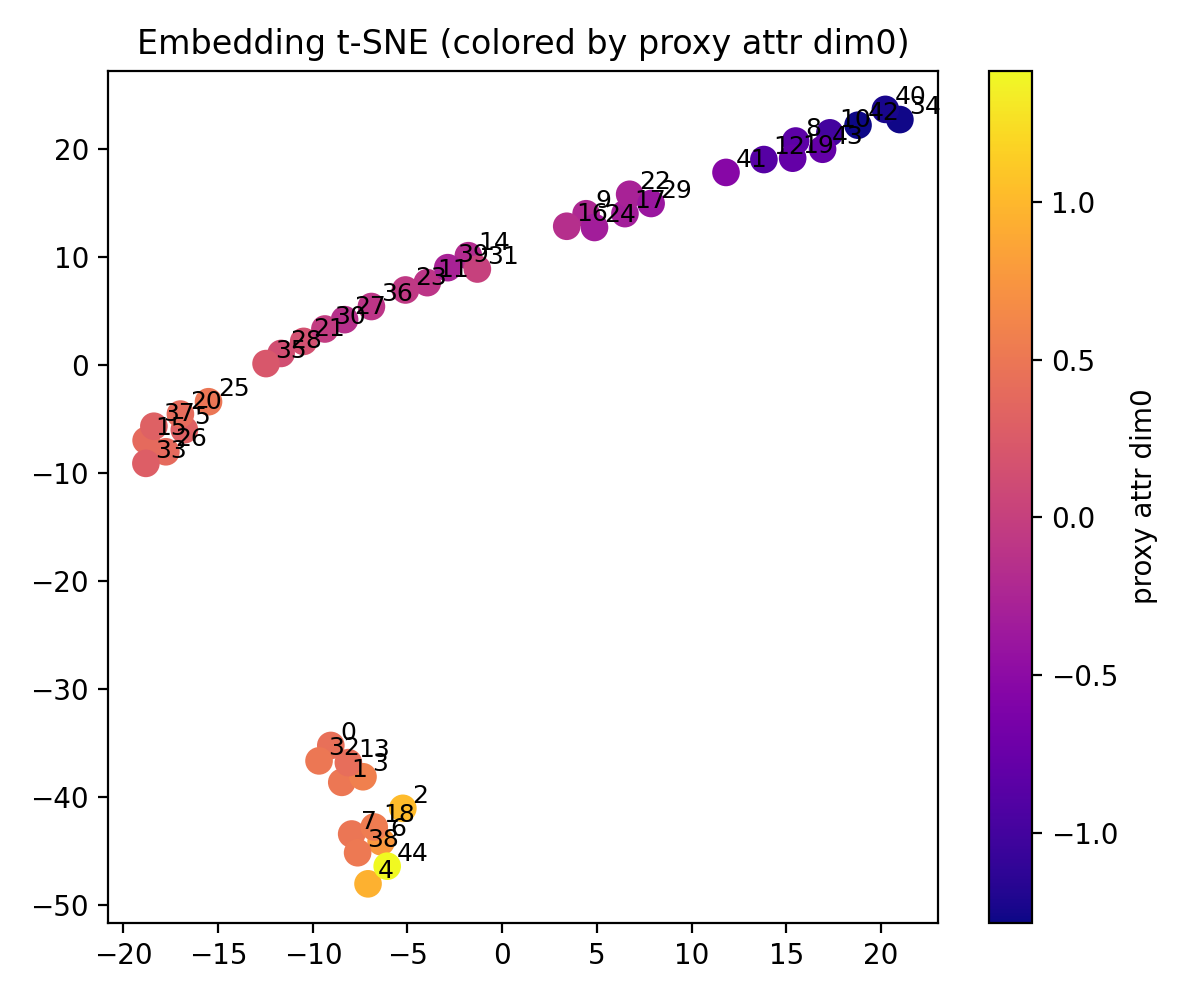}
        \includegraphics[width=\linewidth]{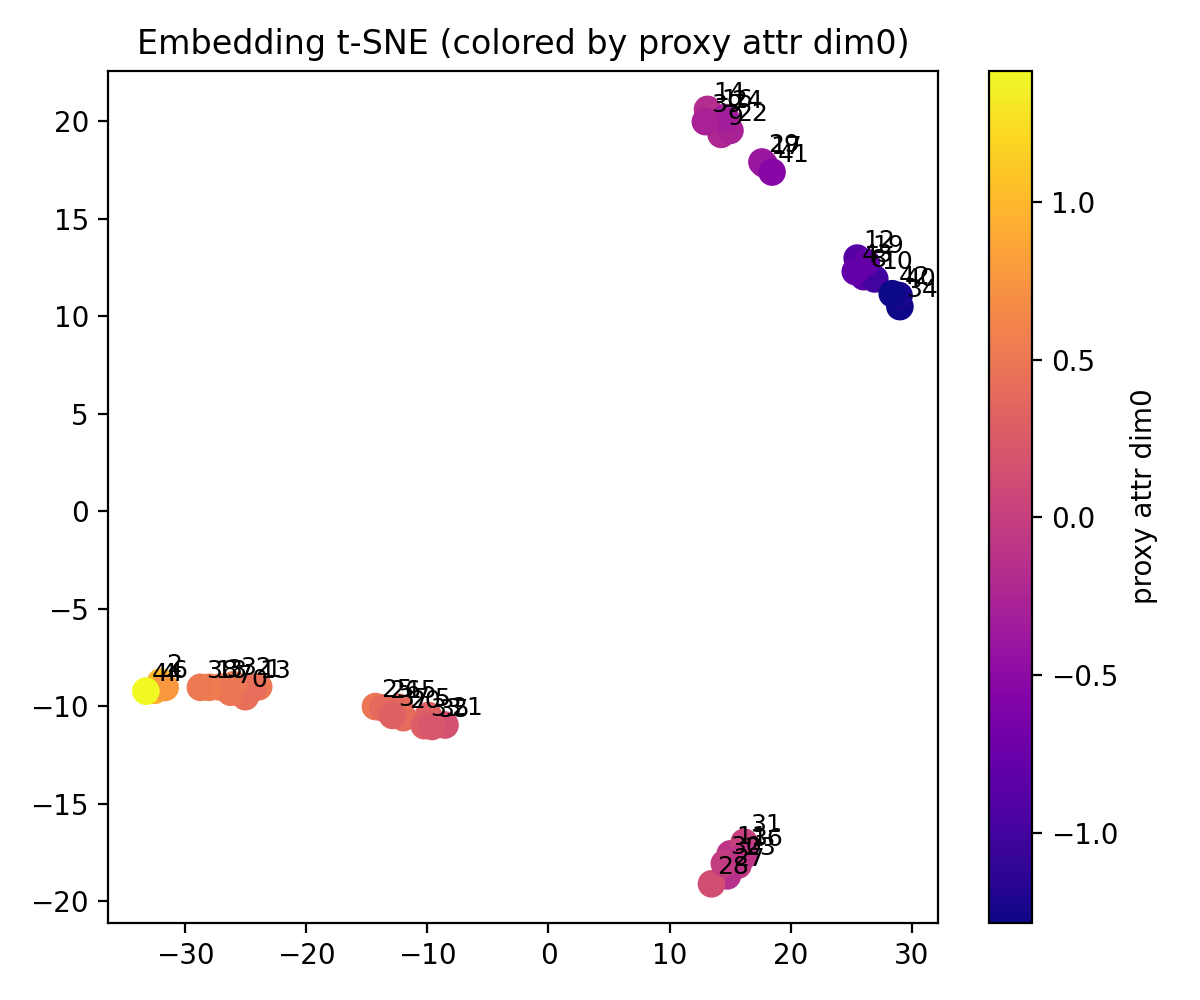}
        \includegraphics[width=\linewidth]{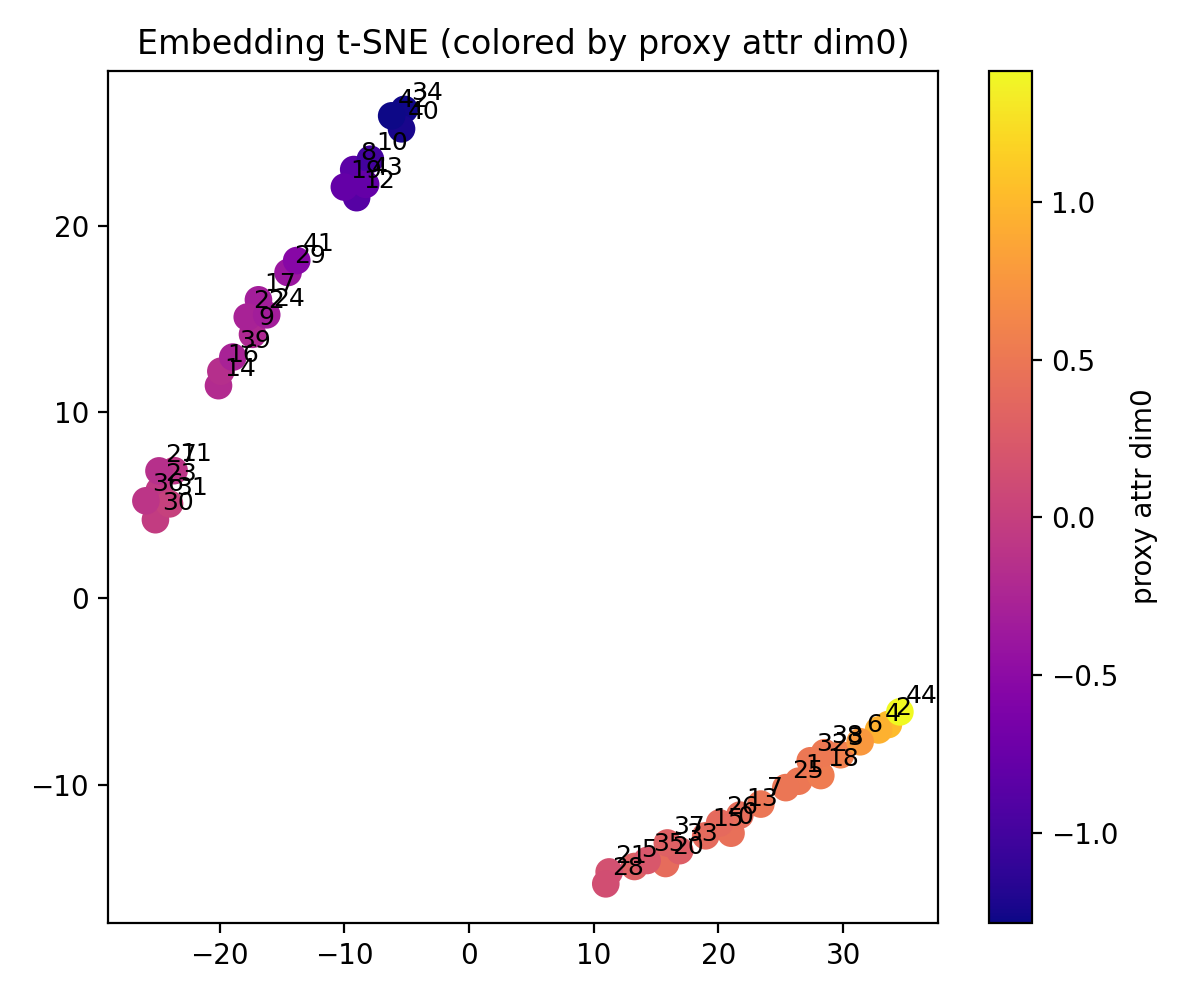}
        \subcaption{t-SNE embedding}
        \label{fig:emb_tsne}
    \end{subfigure}
    \hfill
    \begin{subfigure}[t]{0.2\linewidth}
        \centering
        \includegraphics[width=\linewidth]{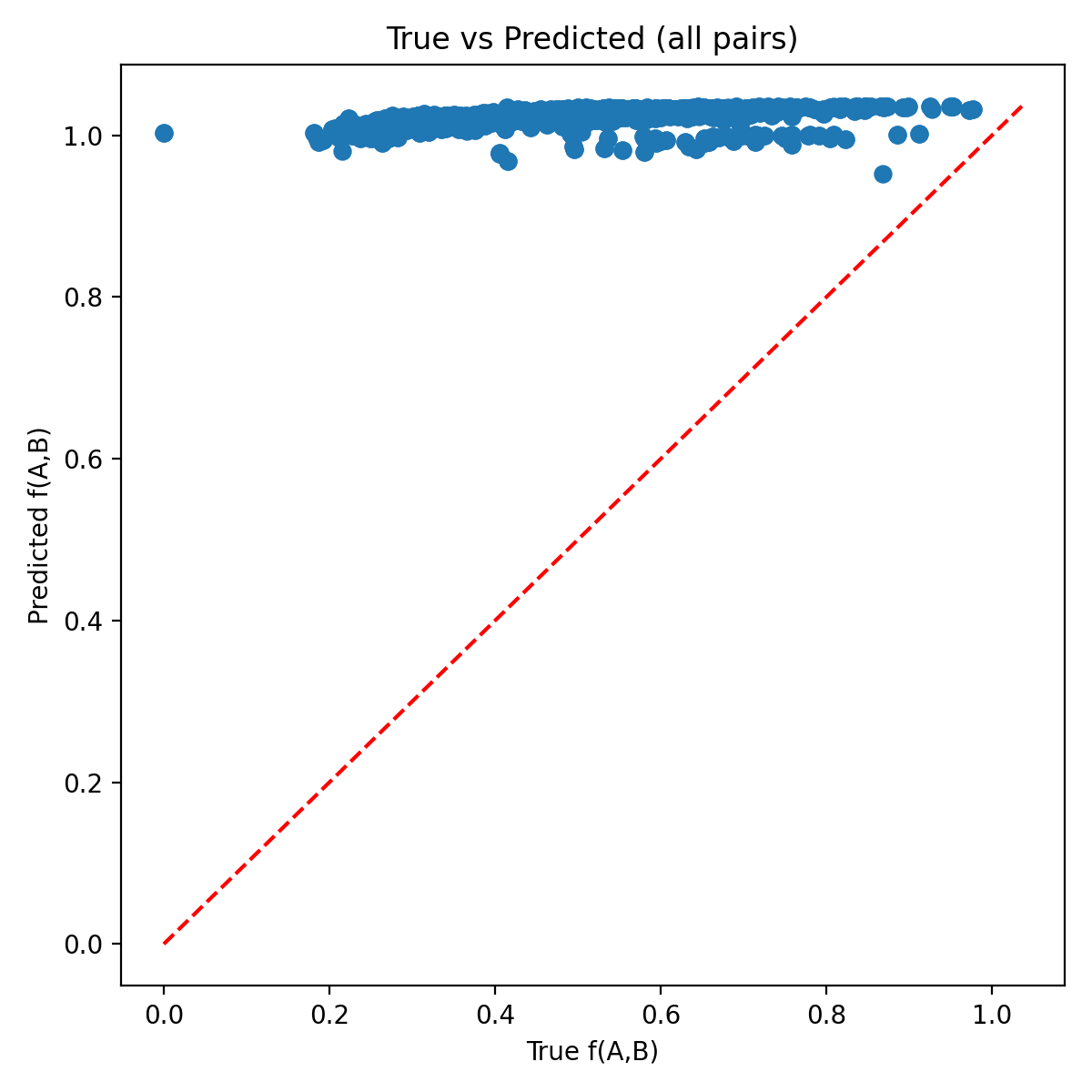}
        \includegraphics[width=\linewidth]{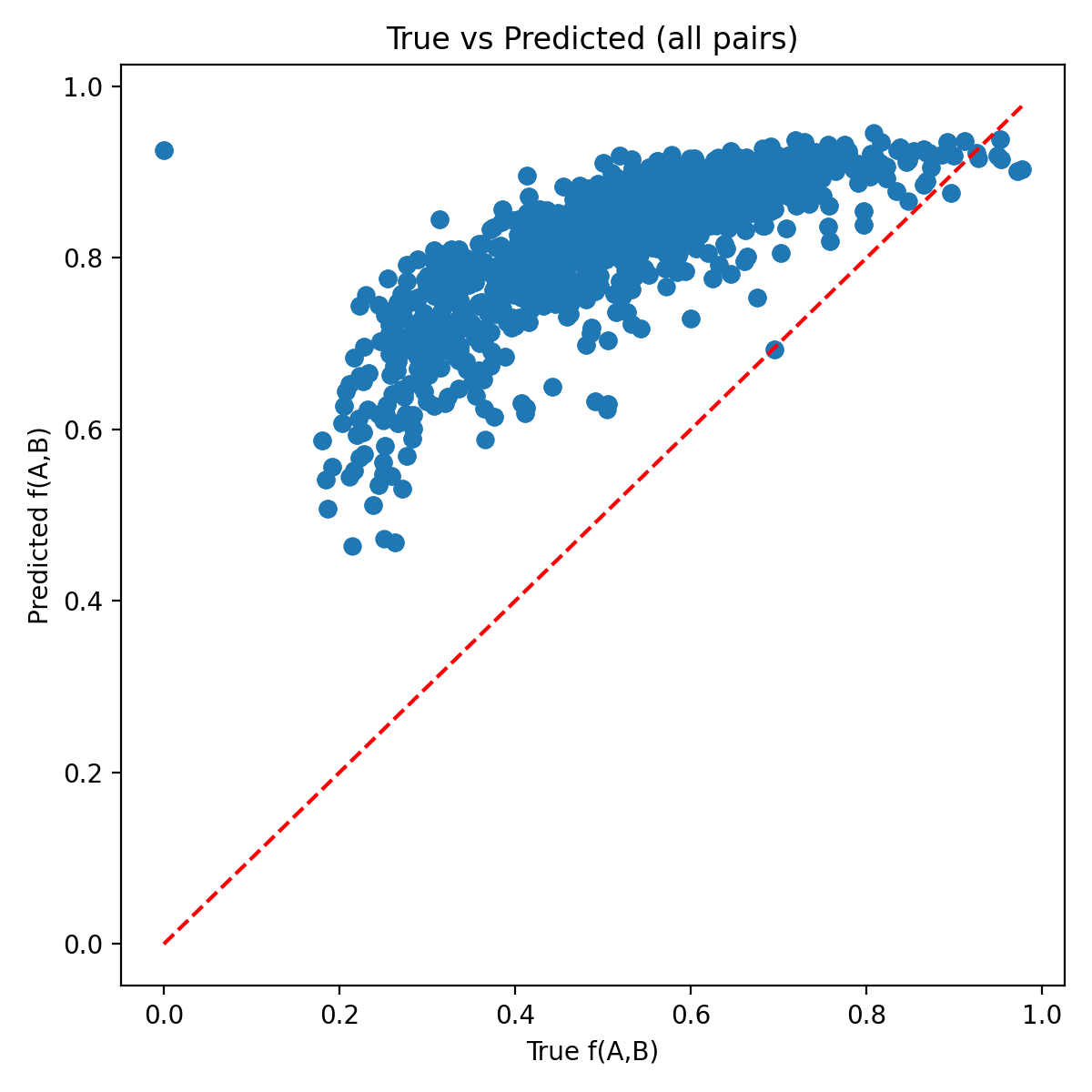}
        \includegraphics[width=\linewidth]{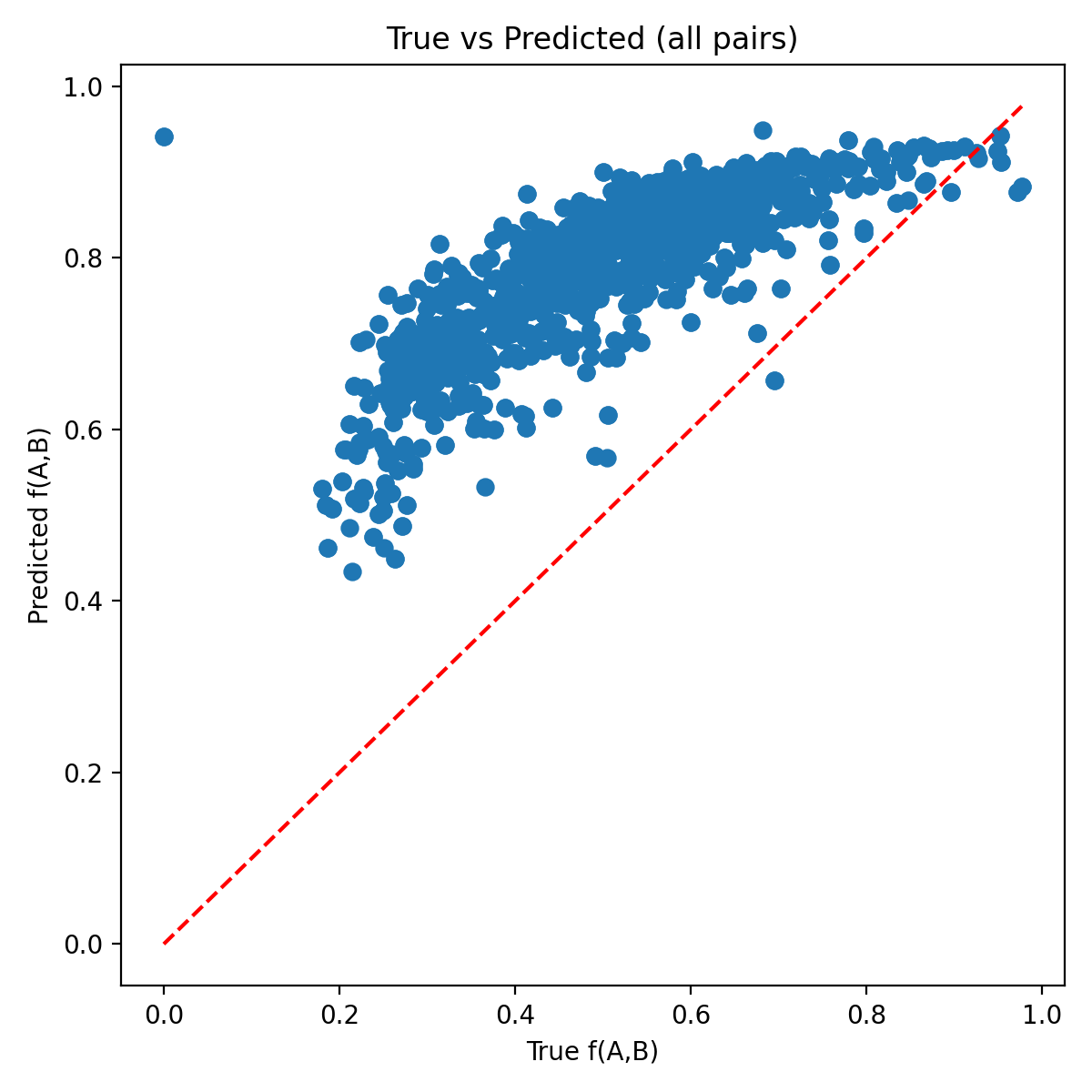}
        \includegraphics[width=\linewidth]{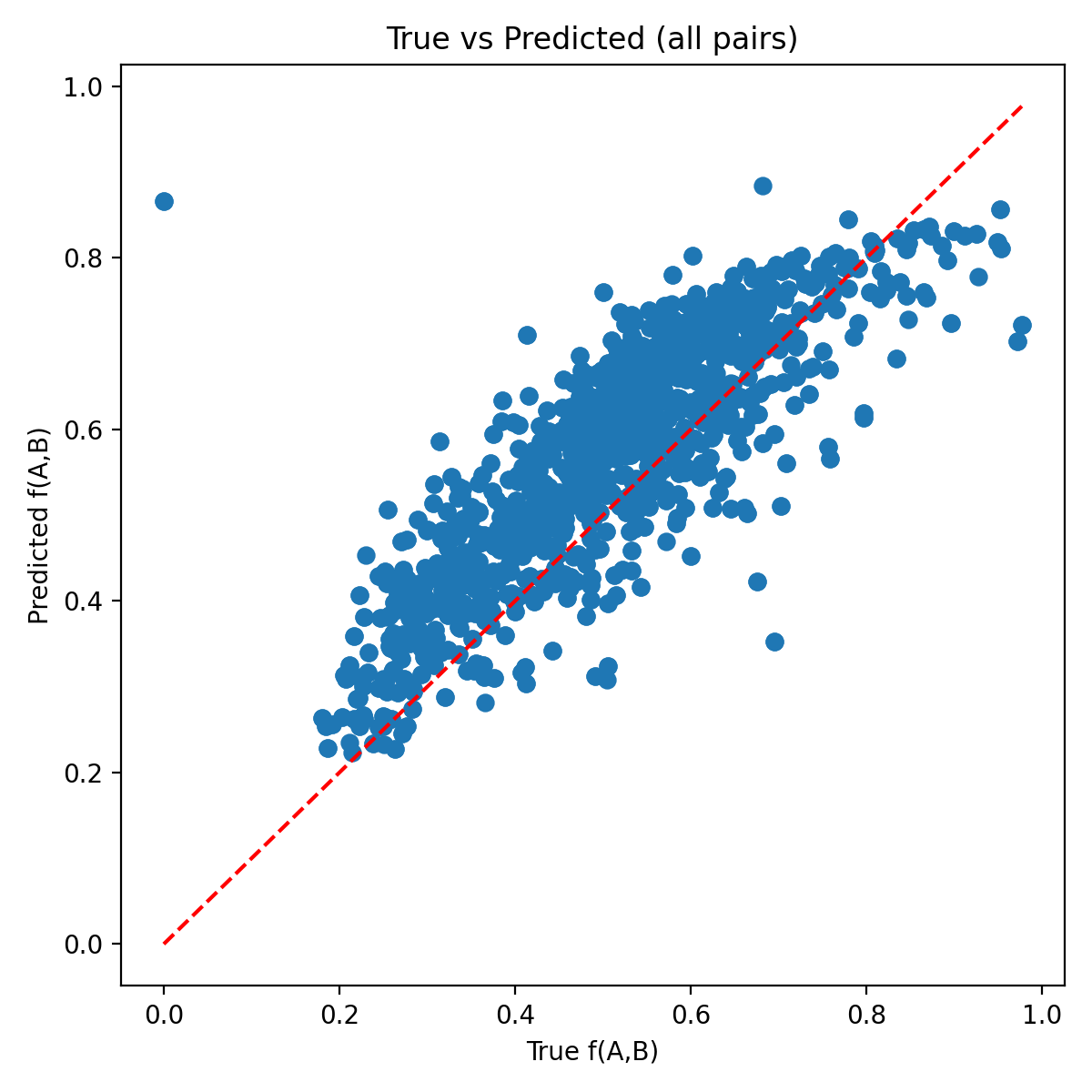}
        \subcaption{True vs. predicted}
        \label{fig:true_vs_pred}
    \end{subfigure}
    \begin{subfigure}[t]{0.3\linewidth}
        \centering
        \includegraphics[width=\linewidth]{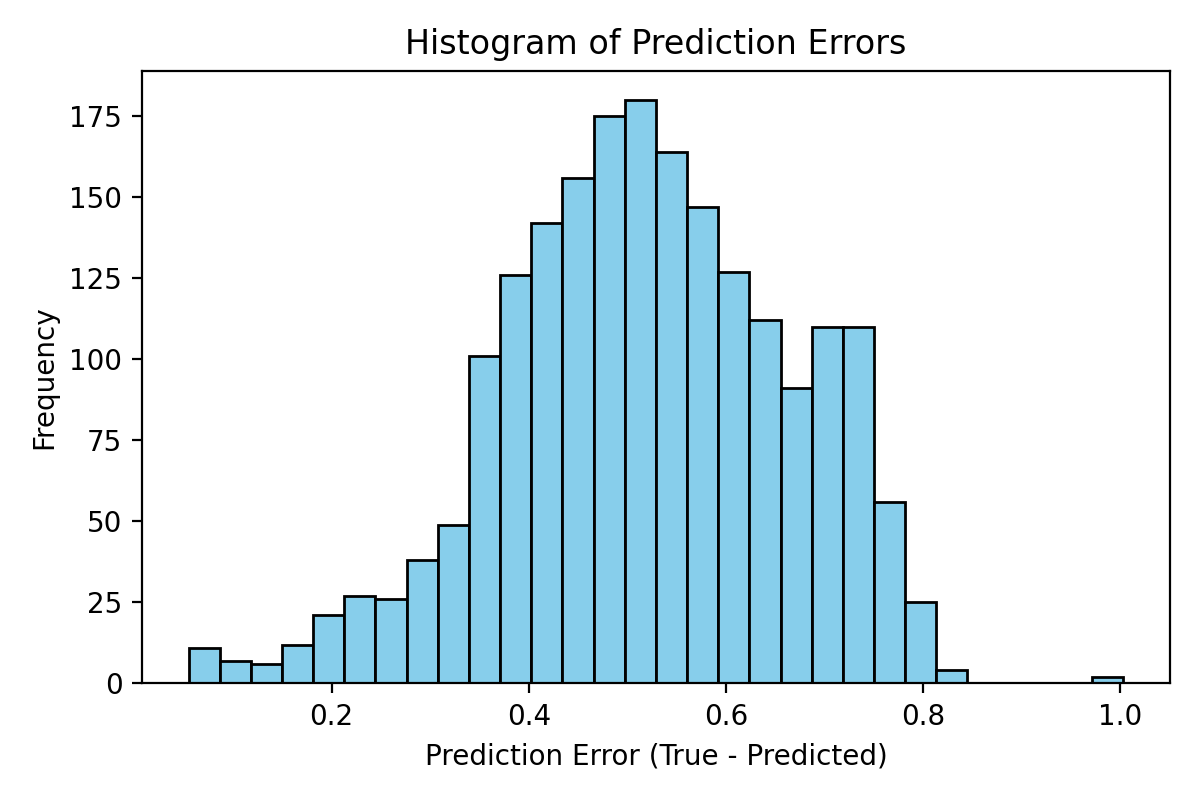}
        \includegraphics[width=\linewidth]{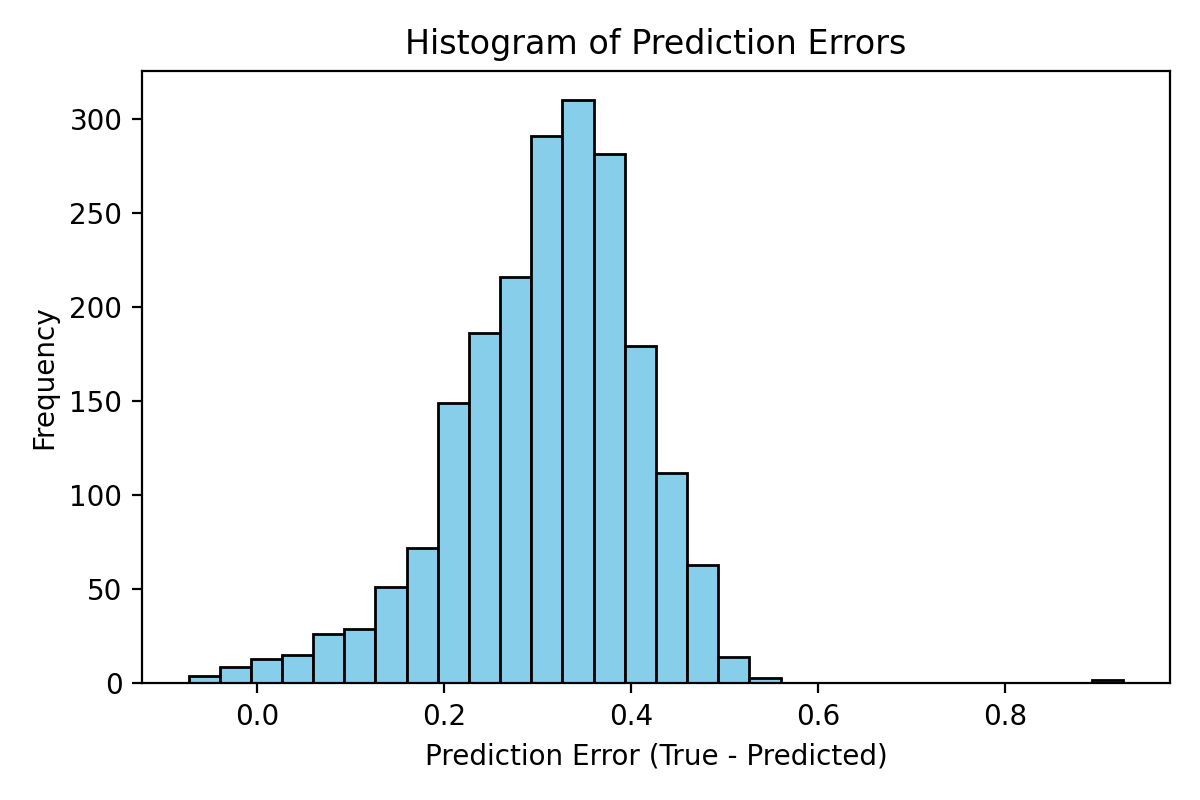}
        \includegraphics[width=\linewidth]{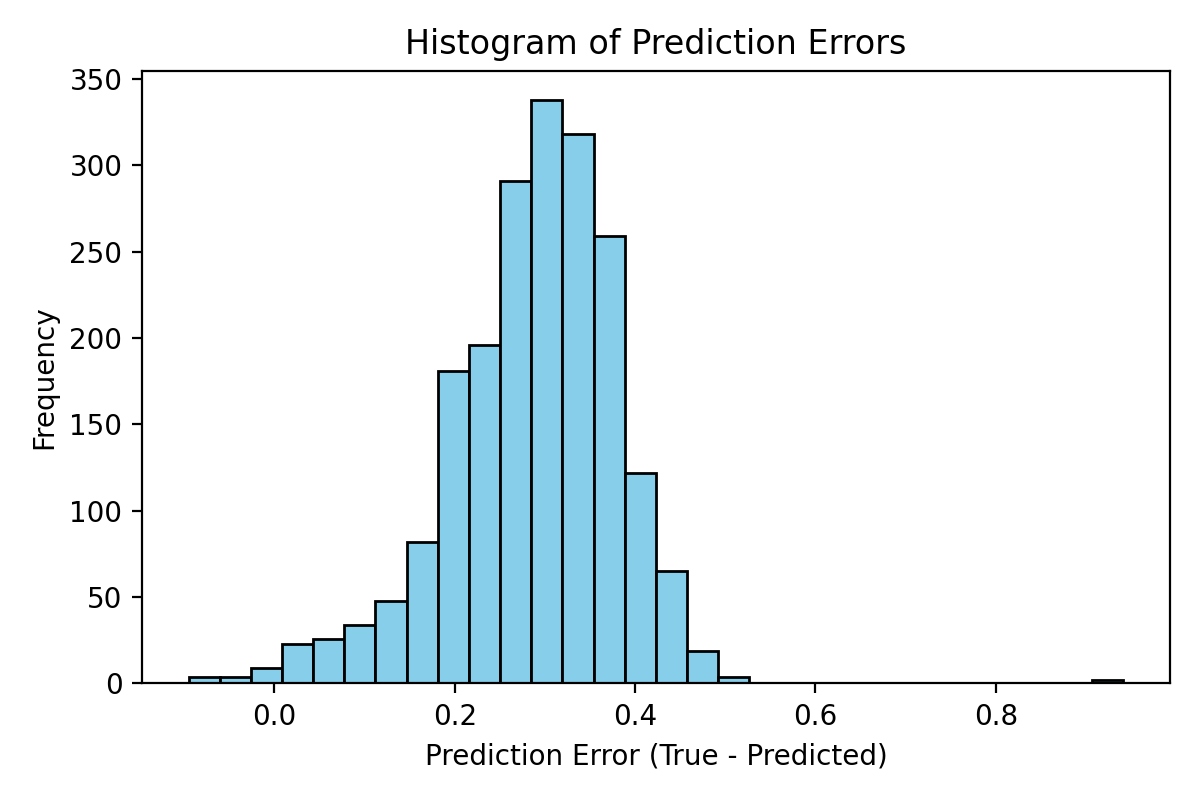}
        \includegraphics[width=\linewidth]{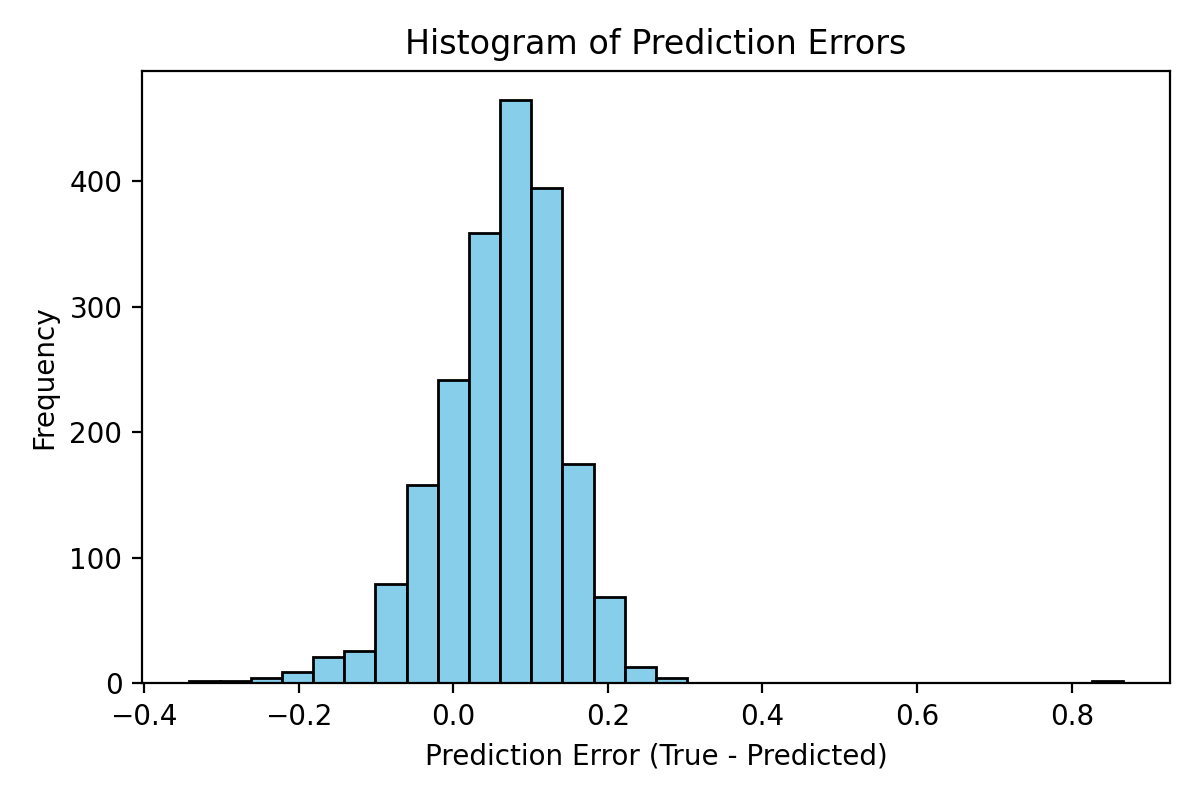}
        \subcaption{Error histogram}
        \label{fig:error_histogram}
    \end{subfigure}
    \caption{RRQR-based proxy selection under progressively stricter spectral retention thresholds (0.95, 0.99, 0.995, 0.999), producing proxy set cardinalities of 1, 7, 13, and 23, respectively. Larger thresholds (higher k) yield monotonic improvements in latent embedding cohesion and pairwise prediction fidelity, consistent with the baseline structure in Fig.~\ref{fig:embeddings_and_predictions}. Compared with the mask-optimization approach (Fig.~\ref{fig:mask_proxies}), RRQR achieves similar accuracy only at higher k, indicating reduced proxy efficiency with respect to embedding preservation and highlighting a trade-off between algebraic coverage and task-aligned selection.}
    \label{fig:rrqr_proxies}
\end{figure*}

\begin{figure*}
    \centering
    \begin{subfigure}[t]{0.24\linewidth}
        \centering
        \includegraphics[width=\linewidth]{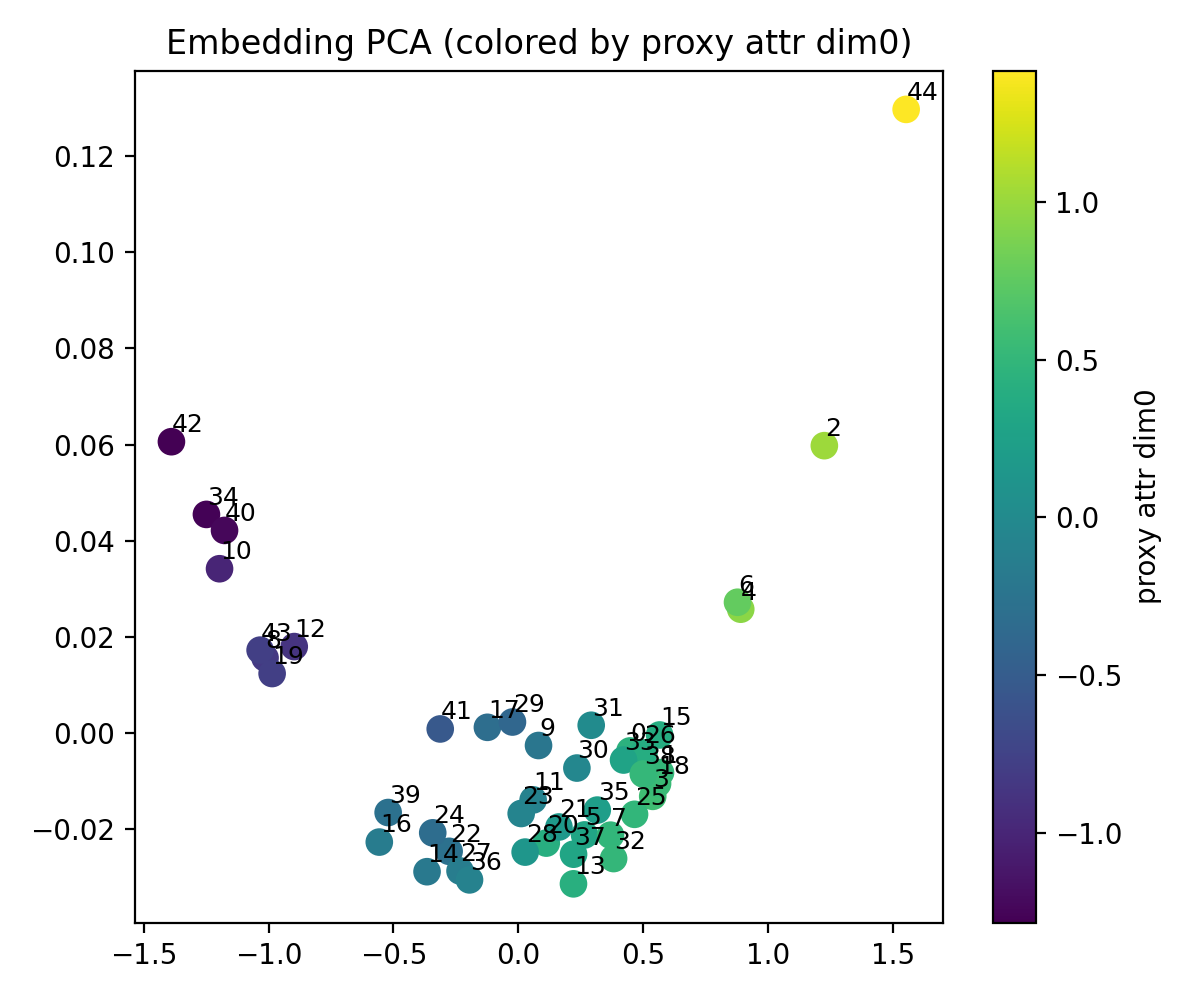}
        \subcaption{PCA embedding}
        \label{fig:emb_pca}
    \end{subfigure}
    \hfill
    \begin{subfigure}[t]{0.24\linewidth}
        \centering
        \includegraphics[width=\linewidth]{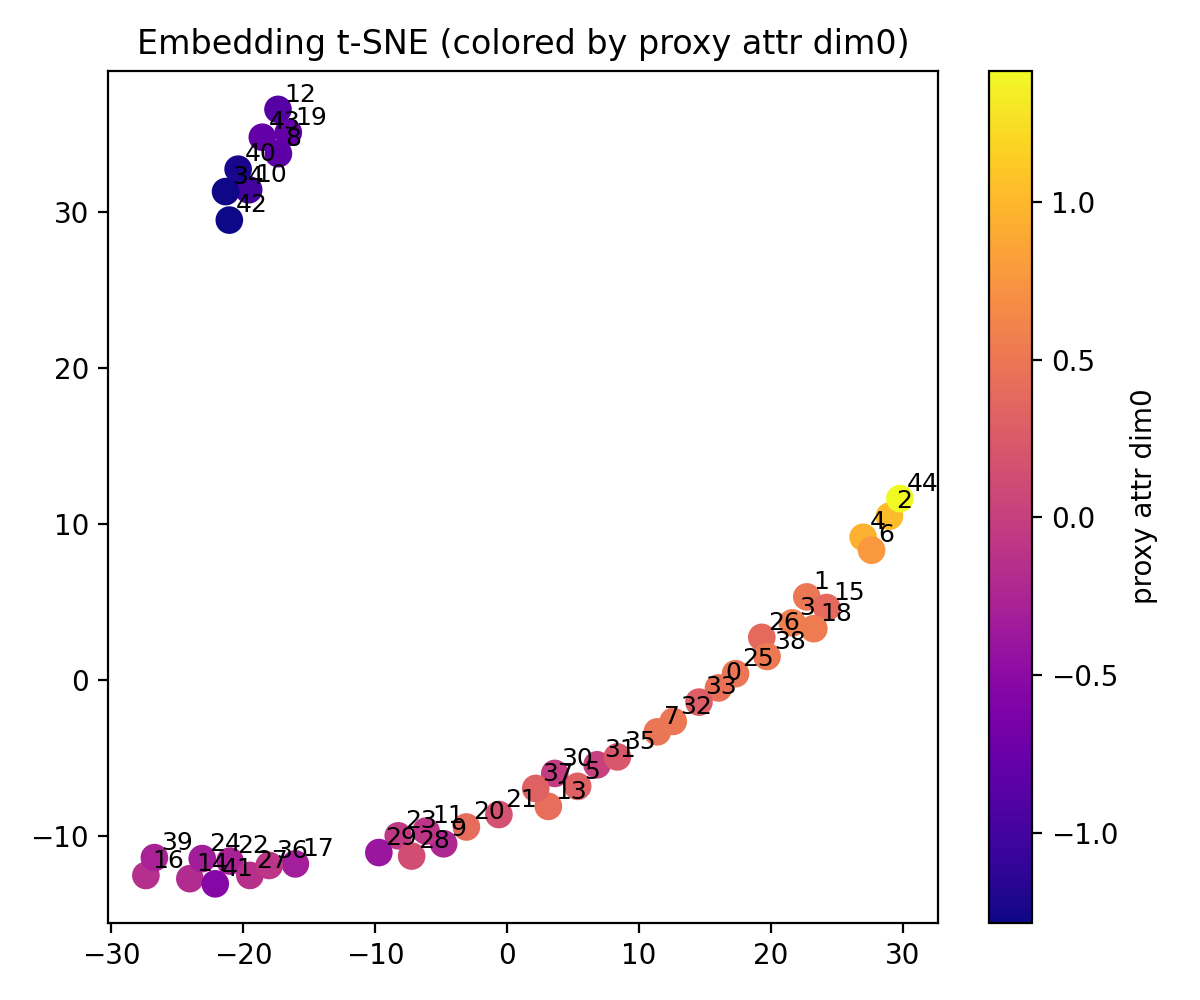}
        \subcaption{t-SNE embedding}
        \label{fig:emb_tsne}
    \end{subfigure}
    \hfill
    \begin{subfigure}[t]{0.2\linewidth}
        \centering
        \includegraphics[width=\linewidth]{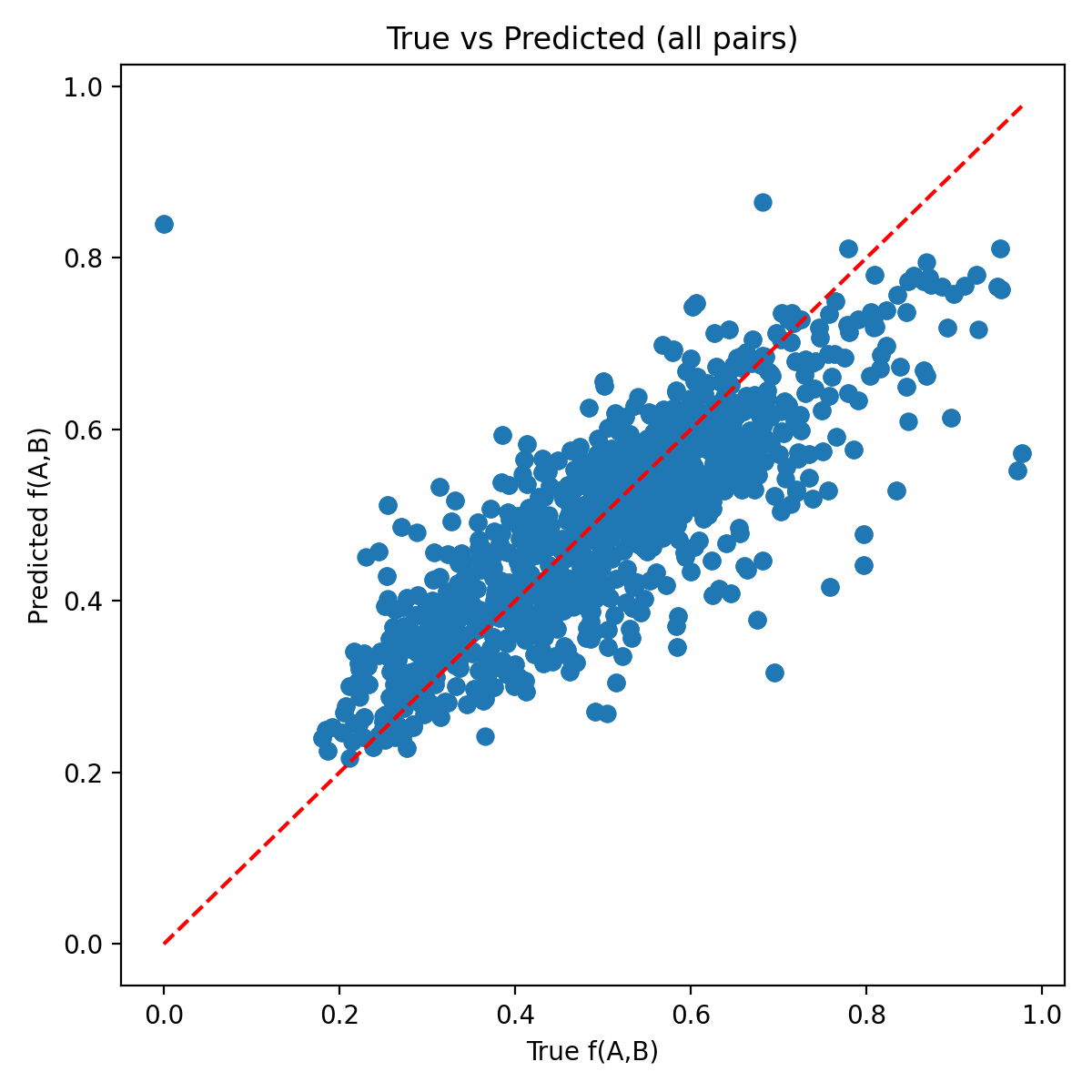}
        \subcaption{True vs. predicted}
        \label{fig:true_vs_pred}
    \end{subfigure}
    \begin{subfigure}[t]{0.3\linewidth}
        \centering
        \includegraphics[width=\linewidth]{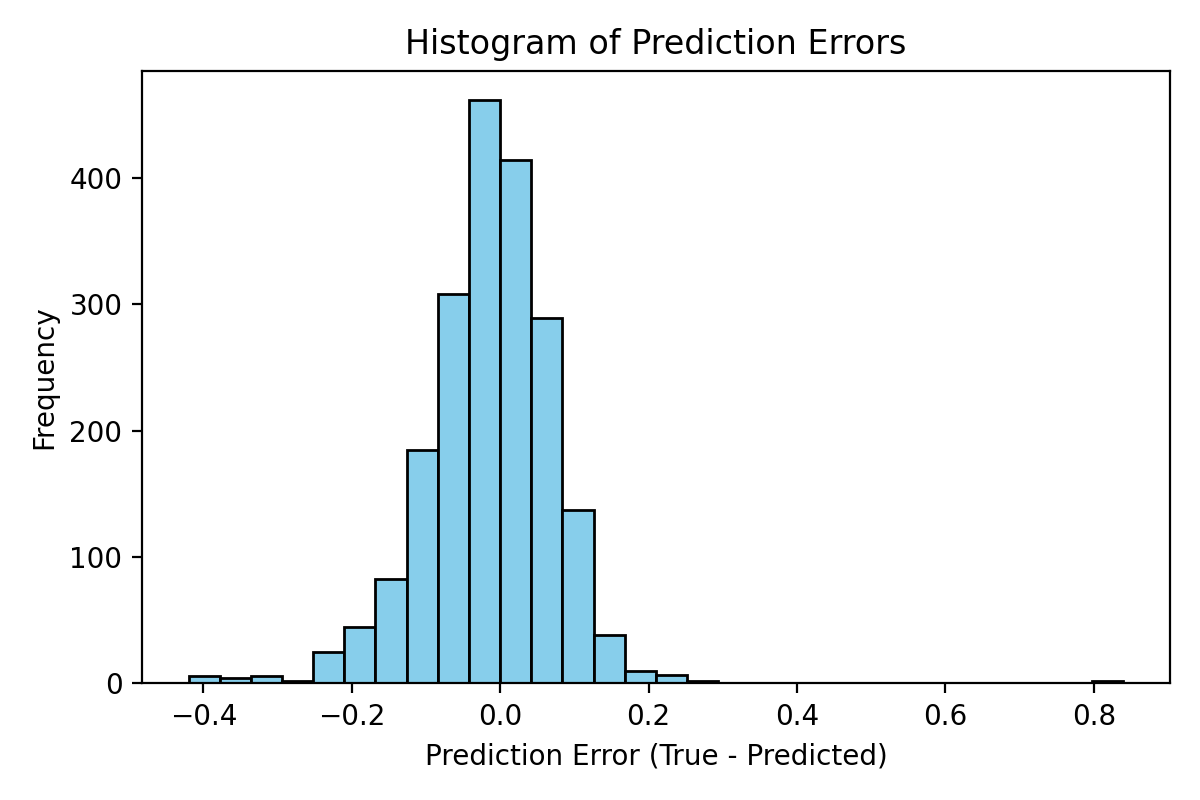}
        \subcaption{Error histogram}
        \label{fig:error_histogram}
    \end{subfigure}
    \caption{Leveraging the learned latent space, the mask-optimization-based proxy selection identifies an optimal set consisting of only two materials. These two proxies suffice to reconstruct the latent embedding structure shown in Figure~\ref{fig:embeddings_and_predictions} and to attain high predictive accuracy, whereas the RRQR-based method in Figure~\ref{fig:rrqr_proxies} requires a larger proxy set to achieve comparable performance.}
    \label{fig:mask_proxies}
\end{figure*}

\paragraph{RRQR Proxy Selection}
The proxy set \(C\) can be selected via rank-revealing QR decomposition on the friction matrix \(\mathbf{F}\). This method identifies a subset of columns (materials) that effectively span the interaction space, maximizing the volume of \(\mathbf{F}_{:,C}\) and enhancing the informativeness of proxy vectors \(v_A\). As demonstrated in Figure~\ref{fig:rrqr_proxies}, we apply RRQR-based proxy selection under progressively stricter spectral retention thresholds (0.95, 0.99, 0.995, 0.999), producing proxy set cardinalities of 1, 7, 13, and 23, respectively. Larger thresholds (higher k) yield monotonic improvements in latent embedding cohesion and pairwise prediction fidelity, consistent with the baseline structure in Fig.~\ref{fig:embeddings_and_predictions}. However, compared with the mask-optimization approach (Fig.~\ref{fig:mask_proxies}), RRQR achieves similar accuracy only at higher k, indicating reduced proxy efficiency with respect to embedding preservation and highlighting a trade-off between algebraic coverage and task-aligned selection.

\paragraph{Embedding-Preserving Proxy Refinement}
Therefore, the selection of the proxy set \(C\) should be based on optimizing the embedding space to ensure that the selected proxies \(v_A\) are representative of the underlying material interactions. This can be achieved through a two-step process: (i) alignment of proxy embeddings with the global latent space, and (ii) minimization of reconstruction error for unseen materials. Leveraging the learned latent space, the mask-optimization-based proxy selection identifies an optimal set consisting of only two materials. These two proxies suffice to reconstruct the latent embedding structure shown in Figure~\ref{fig:embeddings_and_predictions} and to attain high predictive accuracy, whereas the RRQR-based method in Figure~\ref{fig:rrqr_proxies} requires a larger proxy set to achieve comparable performance. This refinement process leads to improved predictive performance and robustness, particularly when the number of proxies is limited.

\subsection{More Evaluations}

\paragraph{Static vs. Kinetic Friction} We can train separate models for static and kinetic friction coefficients, or a joint model predicting both simultaneously. Joint training leverages shared structure and improves sample efficiency, while separate models allow tailored architectures and hyperparameters. We compare performance under both settings to assess trade-offs. Please refer to the supplementary materials for detailed results.

\paragraph{Anisotropy Analysis} For an anisotropic material, we can treat it as multiple distinct materials corresponding to different orientations (e.g., warp and weft directions for textiles). During training, the model learns separate embeddings for each orientation, capturing directional frictional properties. At inference, we can predict friction coefficients for any orientation pair by using the corresponding embeddings. Besides, we can also incorporate orientation information directly into the model, e.g., by augmenting the interaction vectors with directional features or using a conditional encoder that takes orientation as input. This allows the model to learn continuous orientation-dependent embeddings. Please refer to the supplementary materials for detailed results.

\paragraph{Unseen Material Prediction} To further evaluate generalization, we conduct leave-one-material-out cross-validation, where each material is held out in turn during training. The model predicts friction coefficients involving the unseen material using only its proxy interactions. This tests the model's ability to extrapolate to novel materials based on learned embeddings and fusion. Results demonstrate robust performance even for completely unseen materials, highlighting the framework's practical utility. Please refer to the supplementary materials for detailed results.


\section{Conclusions and Limitations}

We presented a proxy–interaction embedding framework for scalable, generalizable estimation of inter‑material friction coefficients. By substituting exhaustive pairwise measurement with a compact, information–optimized proxy set, experimental complexity is reduced from $O(n^2)$ to $O(nk)$ while preserving predictive fidelity for unseen pairs. A jointly trained, symmetry‑respecting encoder–fusion architecture yields latent material descriptors robust to missing proxy observations and supports calibrated uncertainty quantification. Empirical low‑rank structure of friction matrices justifies the proxy basis; rank‑revealing and embedding‑preserving selection strategies enhance efficiency, with mask optimization attaining high accuracy under minimal proxy budgets. The resulting embeddings are compact, task‑aligned, and interpretable, enabling data‑efficient experimental design and reliable integration into physics‑based simulation, thereby offering a practical pathway toward universal friction modeling across heterogeneous material libraries.

However, our method estimates friction coefficients based on measured data and does not explicitly account for underlying physical mechanisms. The learned embeddings may lack direct interpretability in terms of material properties or contact mechanics. Additionally, the approach assumes stationarity and symmetry in frictional interactions, which may not hold for all material pairs or conditions. The model's performance depends on the quality and representativeness of the proxy set; poorly chosen proxies could lead to degraded accuracy. Finally, while uncertainty quantification is incorporated, further validation is needed to ensure reliability under diverse experimental scenarios.

\section{Future Work}

Future work will scale the approach to larger material libraries by improving model scalability and task-aligned proxy selection while expanding the dataset. We will upgrade the measurement device to exploit the learned proxy set, reducing acquisition effort. We will integrate physics-based priors and multimodal data (e.g., surface roughness, hardness) to enhance interpretability and robustness, and investigate active learning for adaptive proxy acquisition to minimize experimental cost without compromising accuracy.

{
    \small
    \bibliographystyle{ieeenat_fullname}
    \bibliography{ref}
}


\input{sec/X_suppl}

\end{document}

%% file: sec/X_suppl.tex
\clearpage
\setcounter{page}{1}
\maketitlesupplementary

\setcounter{section}{0}
\setcounter{subsection}{0}
\setcounter{figure}{0}
\setcounter{table}{0}
\setcounter{equation}{0}
\renewcommand{\thesection}{S\arabic{section}}
\renewcommand{\thefigure}{S\arabic{figure}}
\renewcommand{\thetable}{S\arabic{table}}
\renewcommand{\theequation}{S\arabic{equation}}



\section{Dataset Details}
\label{sec:dataset_details}
In this section, we provide additional details about the dataset used in our study, including the types of materials and measurement procedures, and data preprocessing steps.

\subsection{Material Types}

We employed a diverse set of materials, with emphasis on textile substrates. The dataset comprises 45 fabrics, categorized as knit or woven. The non-fabric subset includes tiles, terracotta panels, glass, metals, wood, plastics, kraft paper, corrugated paper, cowhide leather, and pigskin leather. Each material is implicitly characterized by surface attributes (e.g., roughness, hardness) known to influence friction; however, these attributes were not directly measured. Our objective is to analyze frictional behavior using only friction measurements, without auxiliary surface metrology.

\begin{itemize}
    \item \textbf{Fabric Materials:} knit fabrics (e.g., jersey, rib knit, interlock) and woven fabrics (e.g., plain weave, twill weave, satin weave);
    \item \textbf{Non-fabric Materials:} tiles, terracotta panels, glass, metals, wood, plastics, kraft paper, corrugated paper, cowhide leather, and pigskin leather.
\end{itemize}

\subsection{Measurement Procedures}

Friction measurements were obtained using a custom-built tribometer (Figure \ref{fig:tribometer}) providing precise, continuous control of the slope inclination for a fabric-wrapped test block. A standardized experimental protocol was employed to ensure repeatability and to minimize environmental confounders. For fabric trials, the test block was tightly wrapped in the candidate fabric and placed on a motorized adjustable incline. The slope surface was clad in either (i) the same fabric (self pair), (ii) a different fabric (heterogeneous fabric pair), or (iii) a non-fabric substrate. Because non-fabric substrates are typically harder and smoother, all non-fabric measurements retained a fabric wrapper on the block; direct non-fabric/non-fabric pairings were not acquired. Each valid material pairing was replicated multiple times to capture variability, and reported friction coefficients correspond to the arithmetic mean over repeats.

\paragraph{Static Friction}
Initially, for every trial to measure static friction coefficients, the incline was set to a small angle and the fabric‑wrapped block positioned at its upper boundary. The inclination \(\theta\) was then increased quasi‑statically to induce the onset of motion. The critical angle at first slip, \(\theta_s\), was recorded and the static friction coefficient computed as \(\mu_s=\tan\theta_s\). However, because the slope is not a planar plane but a groove with a sunken angle of $\beta=15^\circ$ on one side, the actual static friction coefficient should be corrected as: $\mu_s = \tan\theta_s\cos\beta$.

\paragraph{Kinetic Friction}
After motion commenced, the incline angle was promptly set and held at a constant value \(\theta_k>\theta_s\) to maintain sustained sliding motion. As shown in Figure \ref{fig:tribometer}, two laser sensors measured the block’s position over time, enabling velocity estimation. For each kinetic‑friction trial, the block was released from rest at the upstream sensor and allowed to traverse a known separation \(d=0.3m\) to the downstream sensor; the transit time \(t\) was recorded. Approximating the motion as uniformly accelerated, \(d=0.5a t^2\). By Newton’s second law on a planar incline, the acceleration is \(a=g(\sin\theta_k-\mu_k\cos\theta_k)\), which yields $\mu_k=\tan\theta_k-2d(g\,t^2\cos\theta_k)^{-1}$, where \(g\) is the gravitational acceleration. On a groove, the corrected kinetic friction coefficient is: $\mu_k=[\tan\theta_k-2d(g\,t^2\cos\theta_k)^{-1}]\cos\beta$.

\subsection{Data Preprocessing}

Raw friction measurements were stored in tabular form; each row encodes a material pairing and friction regime (static or kinetic). The supplementary materials provide the complete data files together with a Python script for deterministic data loading and preprocessing.

We conducted two measurement rounds. The first round involved 45 woven fabrics only (no knit fabrics or non‑fabric substrates) and was performed on a planar incline distinct from the grooved slope described above. This campaign used our first‑generation tribometer; due to time constraints, trials were limited to fabric–fabric pairings. During testing, we observed pronounced anisotropy: friction coefficients depended on the sliding direction relative to the weave or knit. Several fabrics exhibited different behavior along warp versus weft, including lateral drift when misaligned. To ensure consistency in this round, we selected, for each pair, a sliding orientation that produced straight, rectilinear motion down the incline (no side slip), and reported the corresponding coefficients as isotropic. The isotropic fabric friction coefficients from this round form the basis of the isotropic friction matrix \(\mathbf{F}_\text{iso}\in\mathbb{R}^{45\times 45}\). The isotropic friction matrix \(\mathbf{F}_\text{iso}\) is a symmetric \(45\times 45\) matrix where each entry \(f_{i,j}\) represents the friction coefficient between material \(i\) and material \(j\). The diagonal entries \(f_{i,i}\) correspond to self-pair friction coefficients. The matrix is constructed such that
\begin{equation}
\mathbf{F}_\text{iso}=
\begin{bmatrix}
f_{1,1}  & f_{1,2}  & \cdots & f_{1,44}  & f_{1,45}  \\
f_{2,1}  & f_{2,2}  & \cdots & f_{2,44}  & f_{2,45}  \\
\vdots   & \vdots   & \ddots & \vdots    & \vdots    \\
f_{44,1} & f_{44,2} & \cdots & f_{44,44} & f_{44,45} \\
f_{45,1} & f_{45,2} & \cdots & f_{45,44} & f_{45,45}
\end{bmatrix}
\end{equation}

To account for fabric anisotropy, we performed measurements on the grooved tribometer under the two principal, orthogonal directions (warp and weft) for each fabric pair. The groove on the slope was aligned with either the warp or the weft direction of the surface fabric; independently, the fabric on the block was oriented along either warp or weft. This configuration suppresses lateral drift and enforces a rectilinear trajectory. Each orientation combination yielded a distinct pair of friction coefficients \((\mu_s,\mu_k)\), resulting in four orientation conditions---eight coefficients in total---per fabric pair. The second measurement campaign encompassed 30 fabrics (12 knits, 18 wovens) and 10 non-fabric substrates, with all trials conducted on the grooved slope. Owing to scheduling constraints, the second measurement campaign sampled only a subset of the full Cartesian set of admissible fabric–fabric and fabric–non‑fabric pairings. The resulting anisotropic friction data populate an incomplete block structure: complete coverage was obtained for intra-class knit pairings \(\mathbf{F}_\text{knit}\in\mathbb{R}^{12\times 12}\) and intra-class woven pairings \(\mathbf{F}_\text{woven}\in\mathbb{R}^{18\times 18}\); only a partial subset of inter-class knit–woven pairings \(\mathbf{F}_\text{knit-woven}\in\mathbb{R}^{3\times 18}\) and of fabric–non‑fabric pairings \(\mathbf{F}_\text{non-fabric}\in\mathbb{R}^{10\times 24}\) was acquired. Only 12 woven fabrics were included in the non-fabric–woven subset. Each populated fabric–fabric cell stores the orientation‑resolved coefficient quadruple (warp/warp, warp/weft, weft/warp, weft/weft). Unmeasured pairings are recorded as missing (N/A). For fabric–non‑fabric pairings, each populated cell contains the orientation‑resolved pair (warp, weft), expanded to an ordered four‑tuple (warp, weft, warp, weft) to maintain schema consistency, since only the block’s fabric orientation varies whereas the non‑fabric substrate is treated as orientation‑invariant; unmeasured pairings remain unpopulated. The anisotropic friction data can be conceptually organized into the following block matrix structure:
\begin{equation}
\mathbf{F}_\text{aniso}=\begin{bmatrix}
    \mathbf{F}_\text{knit} & \mathbf{F}_\text{knit-woven} & \times\\
    \times & \mathbf{F}_\text{woven} & \times\\
    \mathbf{F}_\text{non-fabric-knit} & \mathbf{F}_\text{non-fabric-woven} & \mathbf{Nan}\\
\end{bmatrix}
\end{equation}
where \(\times\) denotes transposition for matrix symmetry and \(\mathbf{Nan}\) indicates that non-fabric–non-fabric pairings were not measured. The supplementary materials include the complete anisotropic friction dataset in tabular form, along with a Python script for deterministic data loading and preprocessing.

\section{Experimental Details}
\label{sec:experimental_details}

The principal results in the main text were derived from the isotropic static friction matrix \(\mathbf{F}_\text{iso-s}\in\mathbb{R}^{45\times 45}\), measured across 45 woven fabrics under a single rectilinear orientation. Figure~\ref{fig:rrqr_proxies} illustrates representative proxy sets obtained via rank‑revealing QR (RRQR) applied to \(\mathbf{F}_\text{iso-s}\). The selected material indices (zero‑based) are: size 1: [44]; size 7: [2, 4, 11, 16, 17, 20, 44]; size 13: [0, 2, 4, 9, 11, 16, 17, 20, 25, 29, 38, 42, 44]; size 23: [0, 2, 4, 9, 11, 13, 16, 17, 18, 19, 20, 25, 26, 28, 29, 30, 31, 36, 38, 39, 42, 43, 44]. Figure~\ref{fig:mask_proxies} reports results from the masking‑based optimization applied to the same matrix. Under increasing spectral retention thresholds, material 26 is admitted at retention ratio 0.999 with proxy set sizes 23, while material 24 is never selected. Although materials 24 and 26 are close in number, their distinct fiber compositions (24: 50\% cotton / 50\% ramie; 26: 67.4\% cotton / 25.8\% polyester / 6.8\% chinlon) imply different interfacial mechanics; the exclusion of 24 suggests its contribution is redundant relative to the span already captured by earlier proxies. While the precise mechanistic basis for this selection outcome remains undetermined, the observation underscores the influence of compositional heterogeneity on interfacial friction. Empirically, it supports the working hypothesis that the ensemble of material friction behaviors admits a low-dimensional representation spanned by a finite set of proxy materials, thereby reinforcing the theoretical premise of the proposed proxy selection methods.


\begin{figure*}
    \centering
    \begin{subfigure}[t]{0.24\linewidth}
        \centering
        \includegraphics[width=\linewidth]{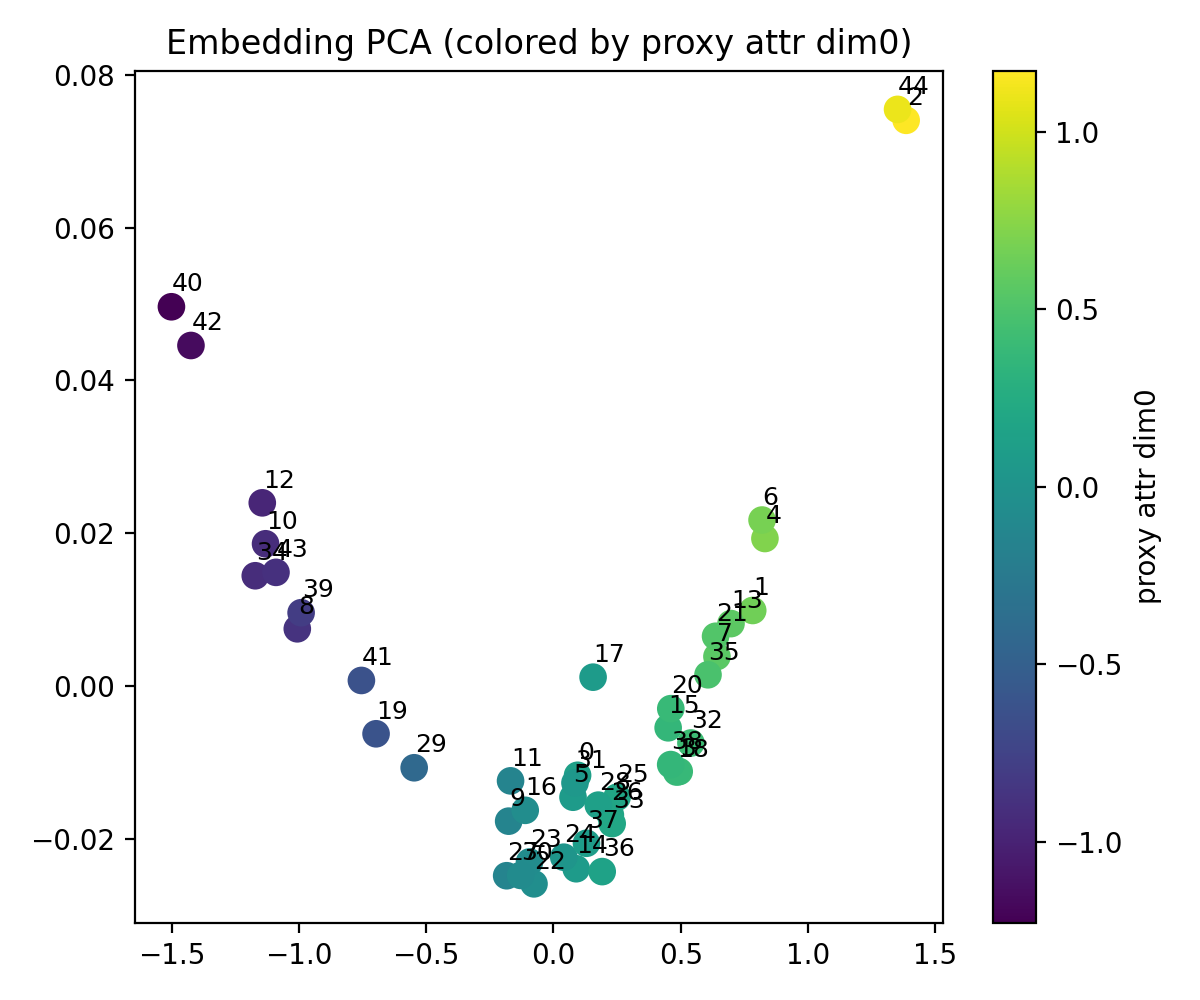}
        \subcaption{PCA embedding}
        \label{fig:emb_pca}
    \end{subfigure}
    \hfill
    \begin{subfigure}[t]{0.24\linewidth}
        \centering
        \includegraphics[width=\linewidth]{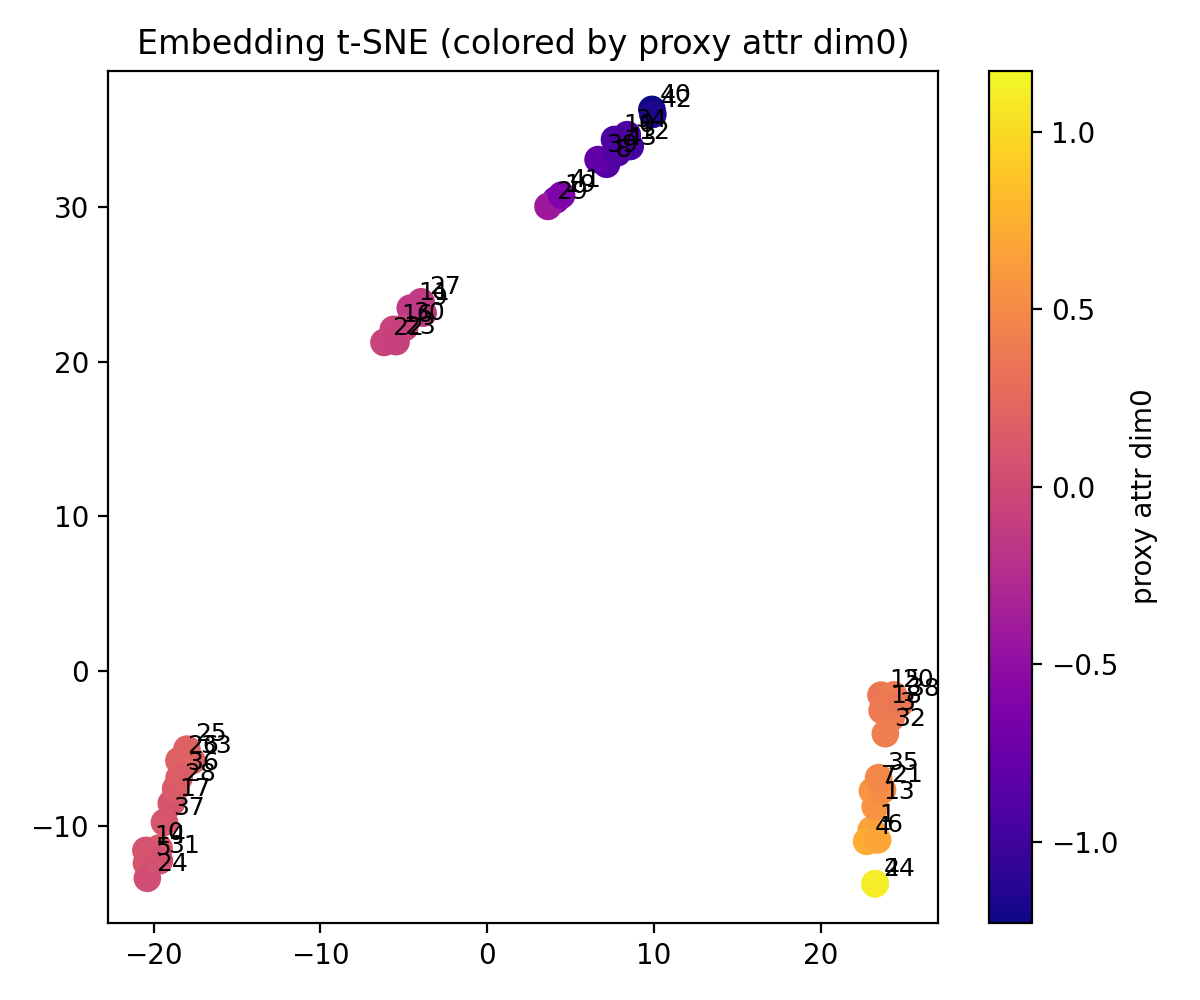}
        \subcaption{t-SNE embedding}
        \label{fig:emb_tsne}
    \end{subfigure}
    \hfill
    \begin{subfigure}[t]{0.2\linewidth}
        \centering
        \includegraphics[width=\linewidth]{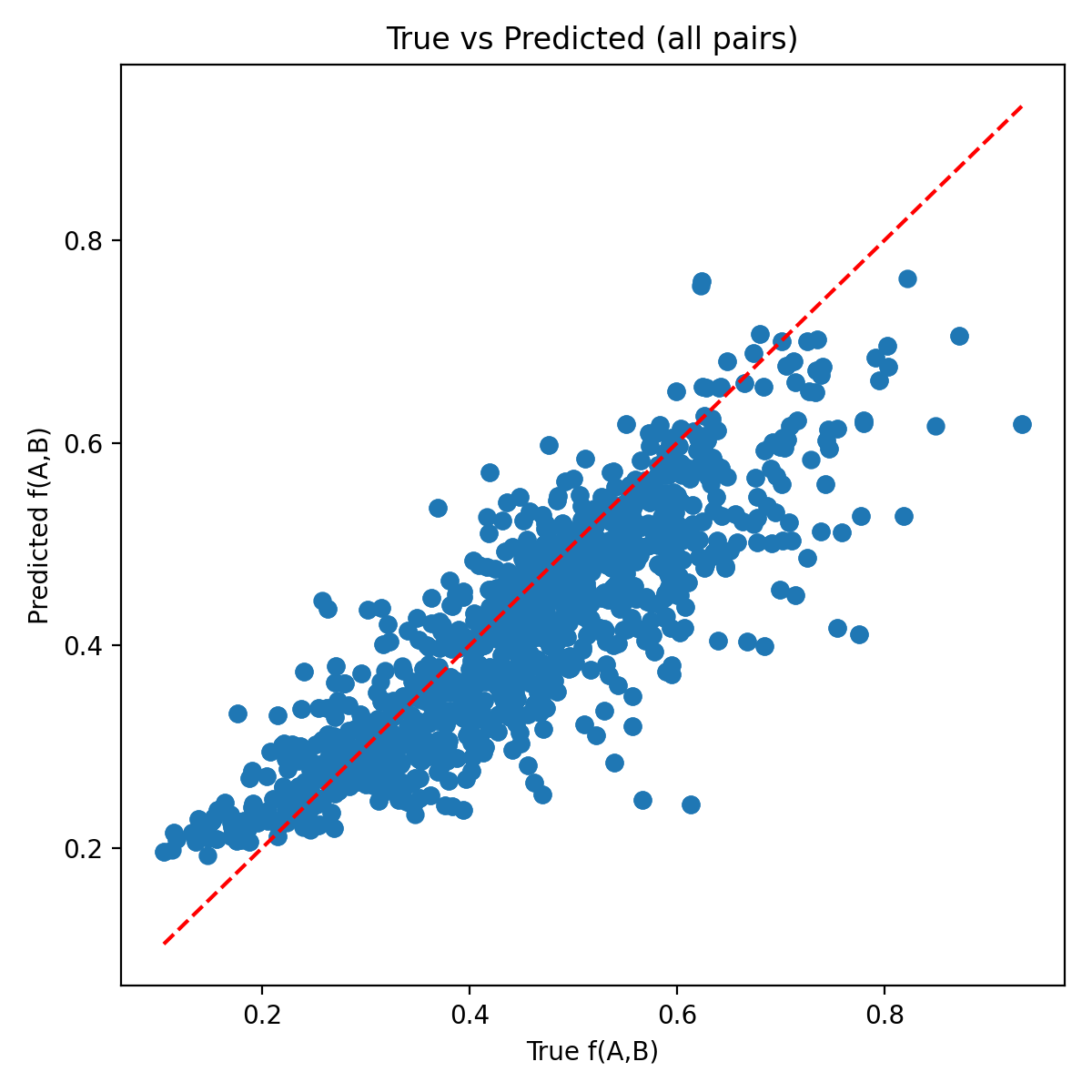}
        \subcaption{True vs. predicted}
        \label{fig:true_vs_pred}
    \end{subfigure}
    \begin{subfigure}[t]{0.3\linewidth}
        \centering
        \includegraphics[width=\linewidth]{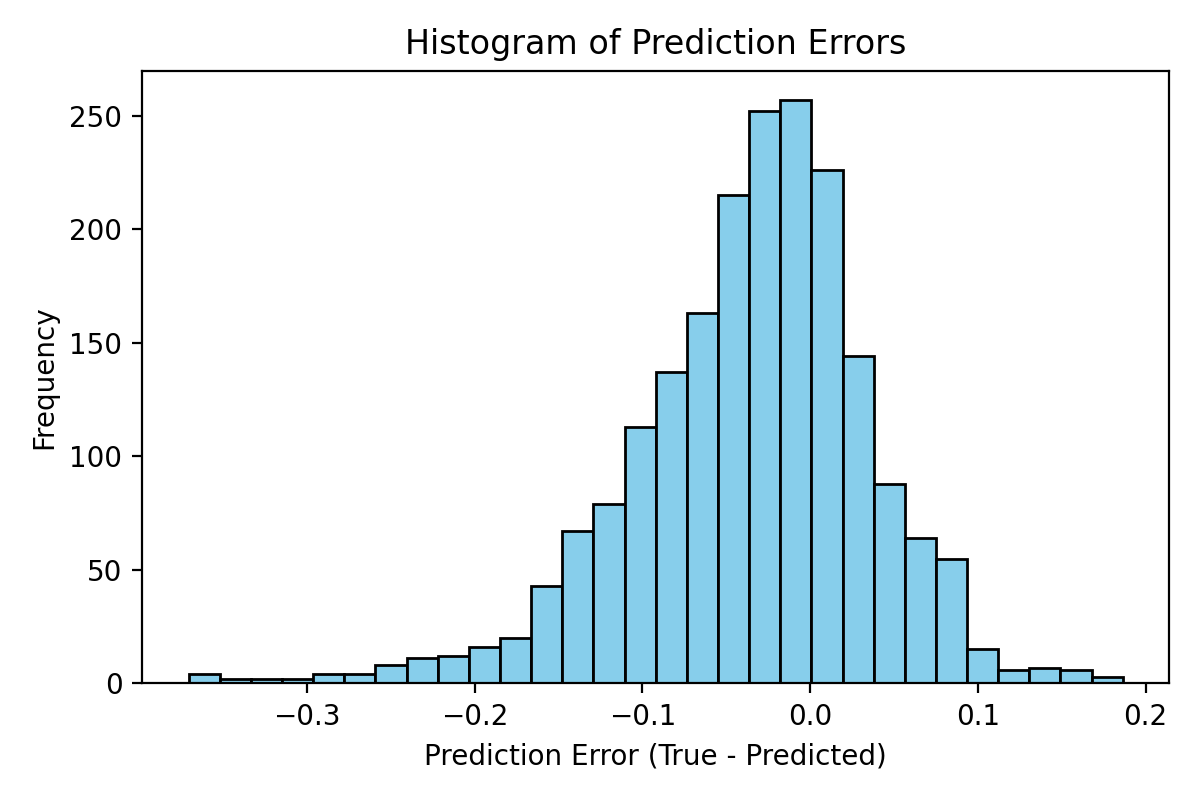}
        \subcaption{Error histogram}
        \label{fig:error_histogram}
    \end{subfigure}
    \caption{When applying the model trained on isotropic static friction data to predict isotropic kinetic friction coefficients, we visualize the learned embeddings and prediction performance. (a) PCA projection of the learned embeddings shows clustering of materials with similar kinetic friction behavior. (b) t-SNE projection further reveals non-linear relationships among material embeddings. (c) The scatter plot of true versus predicted kinetic friction coefficients demonstrates strong correlation, indicating accurate predictions. (d) The error histogram illustrates the distribution of prediction errors, with most errors concentrated around zero, confirming the model's effectiveness in capturing kinetic friction characteristics.}
    \label{fig:kinect-on-static}
\end{figure*}

\begin{figure*}
    \centering
    \begin{subfigure}[t]{0.24\linewidth}
        \centering
        \includegraphics[width=\linewidth]{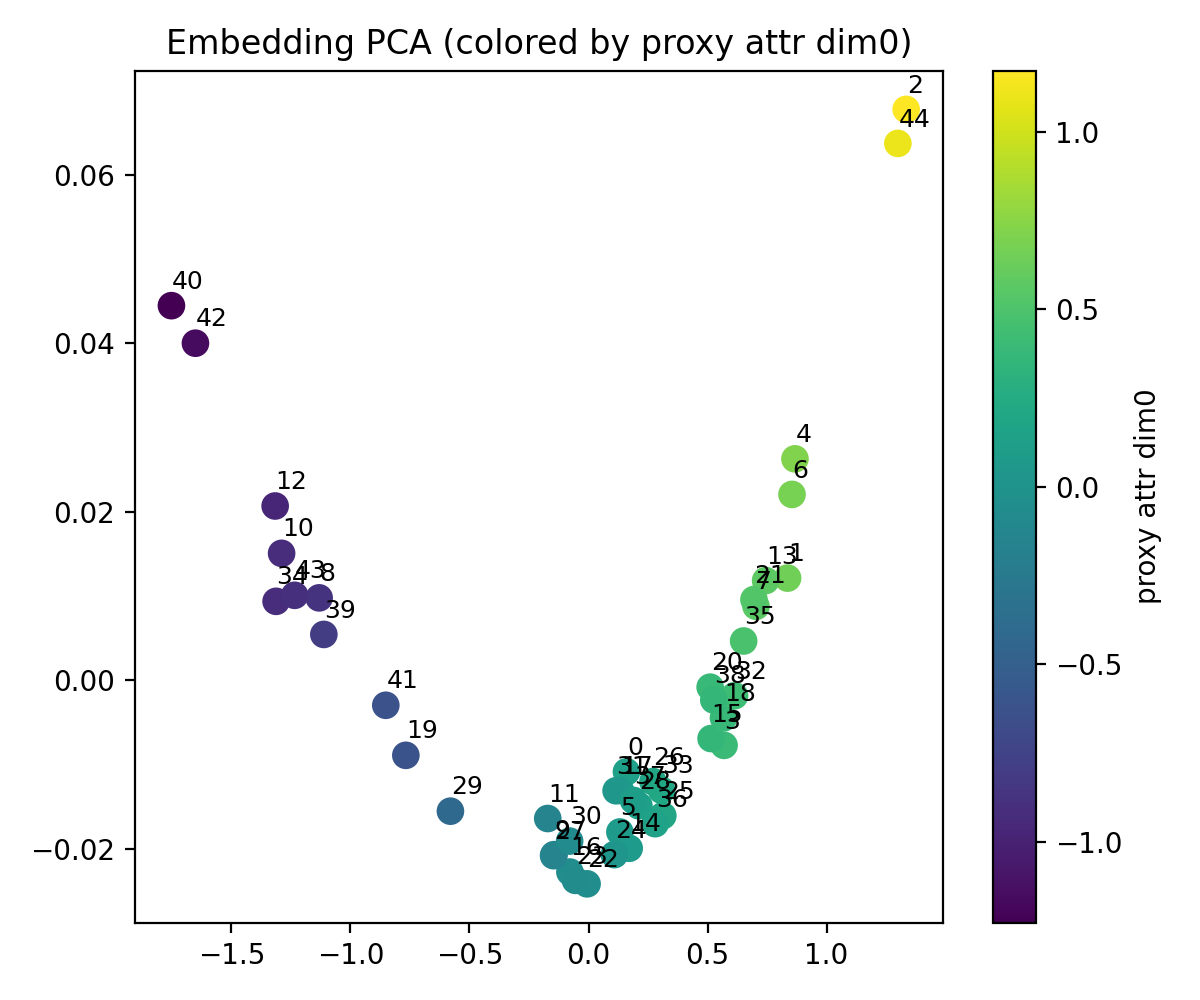}
        \subcaption{PCA embedding}
        \label{fig:emb_pca}
    \end{subfigure}
    \hfill
    \begin{subfigure}[t]{0.24\linewidth}
        \centering
        \includegraphics[width=\linewidth]{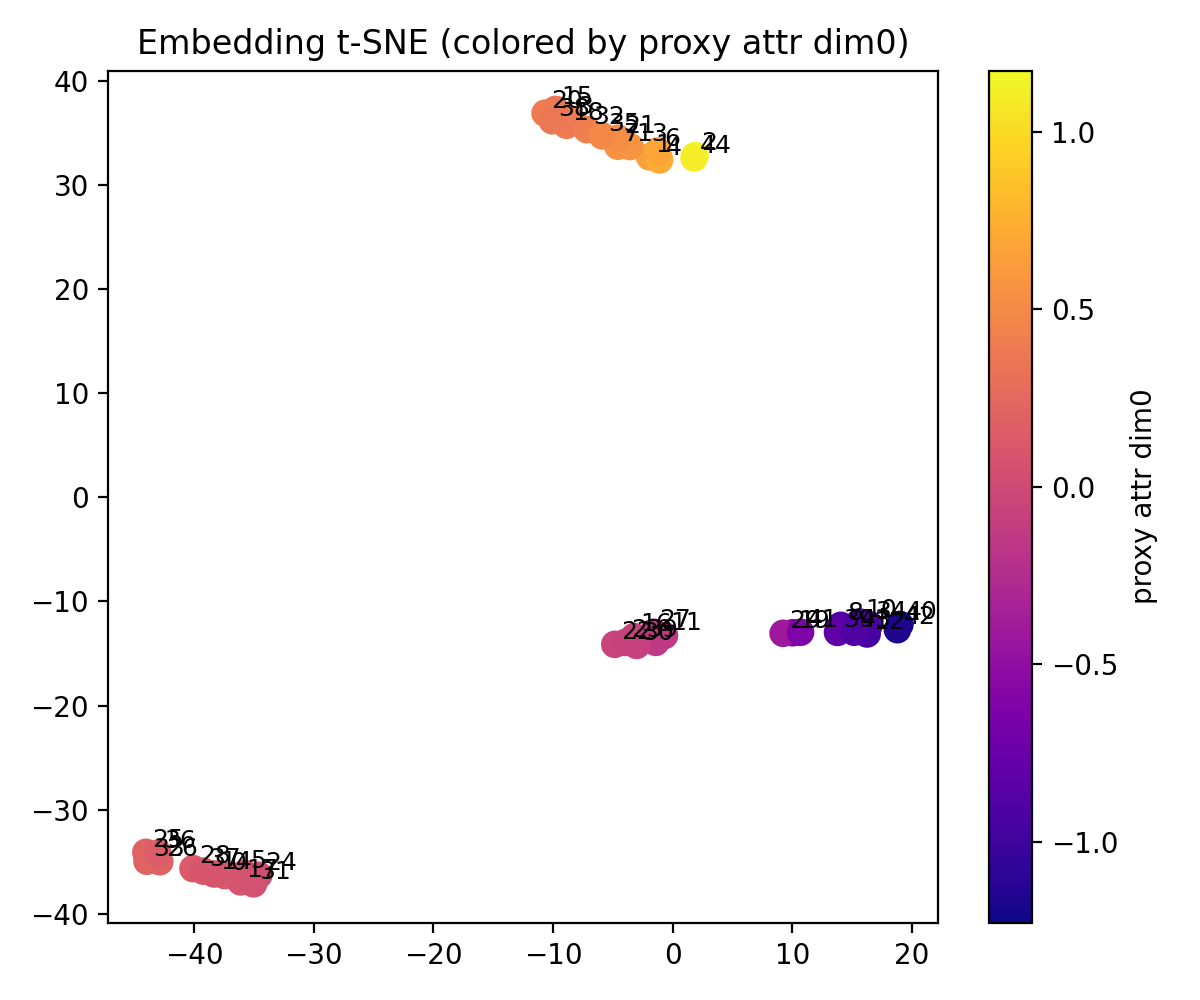}
        \subcaption{t-SNE embedding}
        \label{fig:emb_tsne}
    \end{subfigure}
    \hfill
    \begin{subfigure}[t]{0.2\linewidth}
        \centering
        \includegraphics[width=\linewidth]{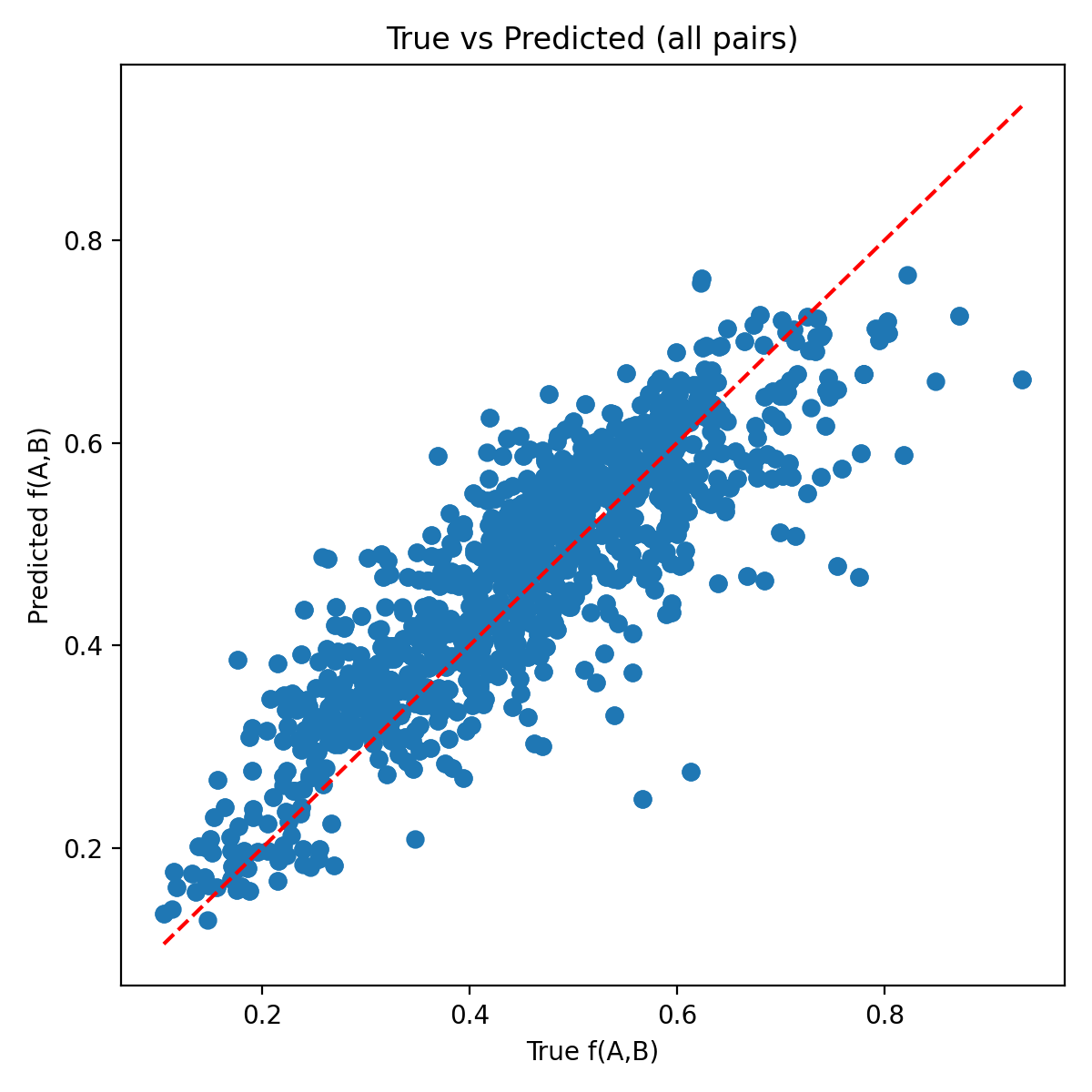}
        \subcaption{True vs. predicted}
        \label{fig:true_vs_pred}
    \end{subfigure}
    \begin{subfigure}[t]{0.3\linewidth}
        \centering
        \includegraphics[width=\linewidth]{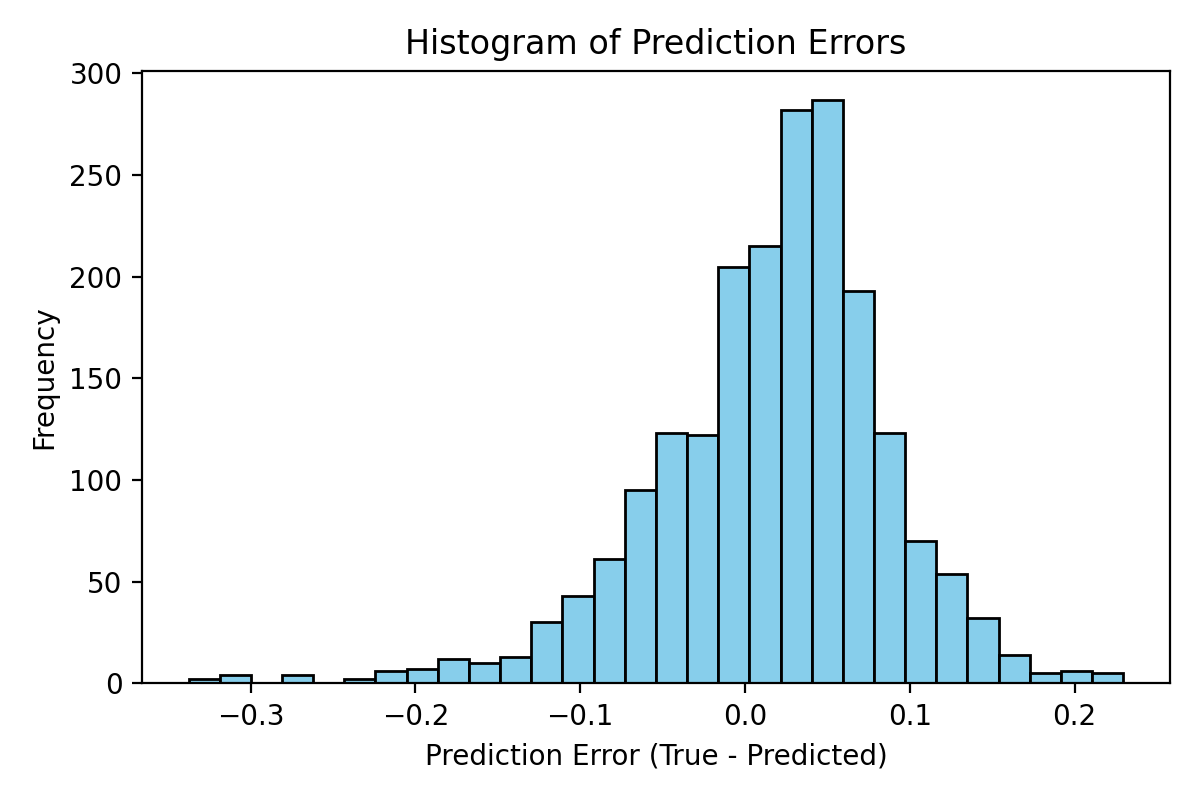}
        \subcaption{Error histogram}
        \label{fig:error_histogram}
    \end{subfigure}
    \caption{When training a separate model on isotropic kinetic friction data, we visualize the learned embeddings and prediction performance. (a) PCA projection of the learned embeddings shows clustering of materials with similar kinetic friction behavior. (b) t-SNE projection further reveals non-linear relationships among material embeddings. (c) The scatter plot of true versus predicted kinetic friction coefficients demonstrates strong correlation, indicating accurate predictions. (d) The error histogram illustrates the distribution of prediction errors, with most errors concentrated around zero, confirming the model's effectiveness in capturing kinetic friction characteristics.}
    \label{fig:embeddings_and_predictions-kinect}
\end{figure*}

\section{Additional Experimental Results}
\label{sec:additional_results}
In this section, we present additional experimental results that complement the findings reported in the main text.

\subsection{Isotropic Friction}

We conducted a comparative analysis of static and kinetic friction across all isotropically measured pairings. Consistent with tribological theory, the static coefficient \(\mu_s\) generally exceeded the kinetic coefficient \(\mu_k\) by a modest margin, with rare deviations within experimental uncertainty. The magnitude of the \(\mu_s-\mu_k\) gap varied with interfacial descriptors (e.g., surface roughness, compliance, and fiber composition). Nevertheless, \(\mu_s\) and \(\mu_k\) exhibited strong concordance in rank ordering and trends across materials, indicating shared governing mechanisms. This alignment facilitates modeling and simulation by enabling a shared latent representation with regime-specific calibration.


\paragraph{Transferability Between Friction Regimes}
We extended the analysis to the isotropic kinetic friction matrix $\mathbf{F}_{\text{iso-k}}$ compiled from the same material set. To evaluate transferability, models trained on $\mathbf{F}_{\text{iso-s}}$ were used to predict kinetic coefficients. While transfer learning provides competitive accuracy, and the results reproduce the qualitative structure observed for the static regime, supporting the cross‑regime generalizability of the methodology, regime‑dependent biases remain; thus, modest recalibration and rigorous validation are required to ensure reliable performance. As demonstrated in Figure~\ref{fig:kinect-on-static}, the model trained on isotropic static friction data effectively captures the underlying relationships in isotropic kinetic friction data, as evidenced by the coherent embeddings and not bad accurate predictions. This suggests that the learned representations are robust and can generalize across different friction regimes. However, to achieve optimal performance, it may be necessary to fine-tune the model specifically for kinetic friction data, as the distinct characteristics of each friction regime can influence the accuracy of predictions.


\begin{figure*}
    \centering
    \begin{subfigure}[t]{0.24\linewidth}
        \centering
        \includegraphics[width=\linewidth]{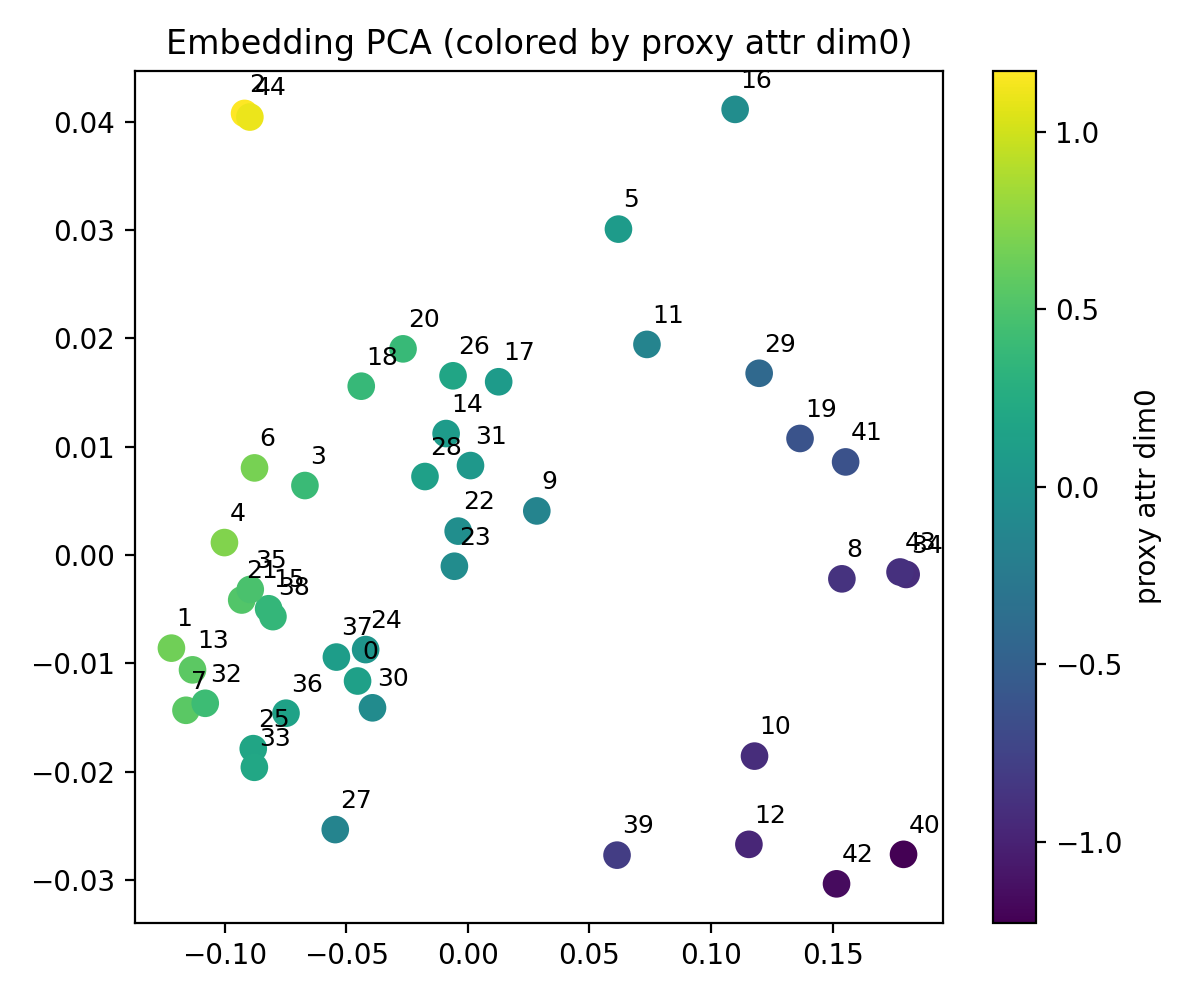}
        \includegraphics[width=\linewidth]{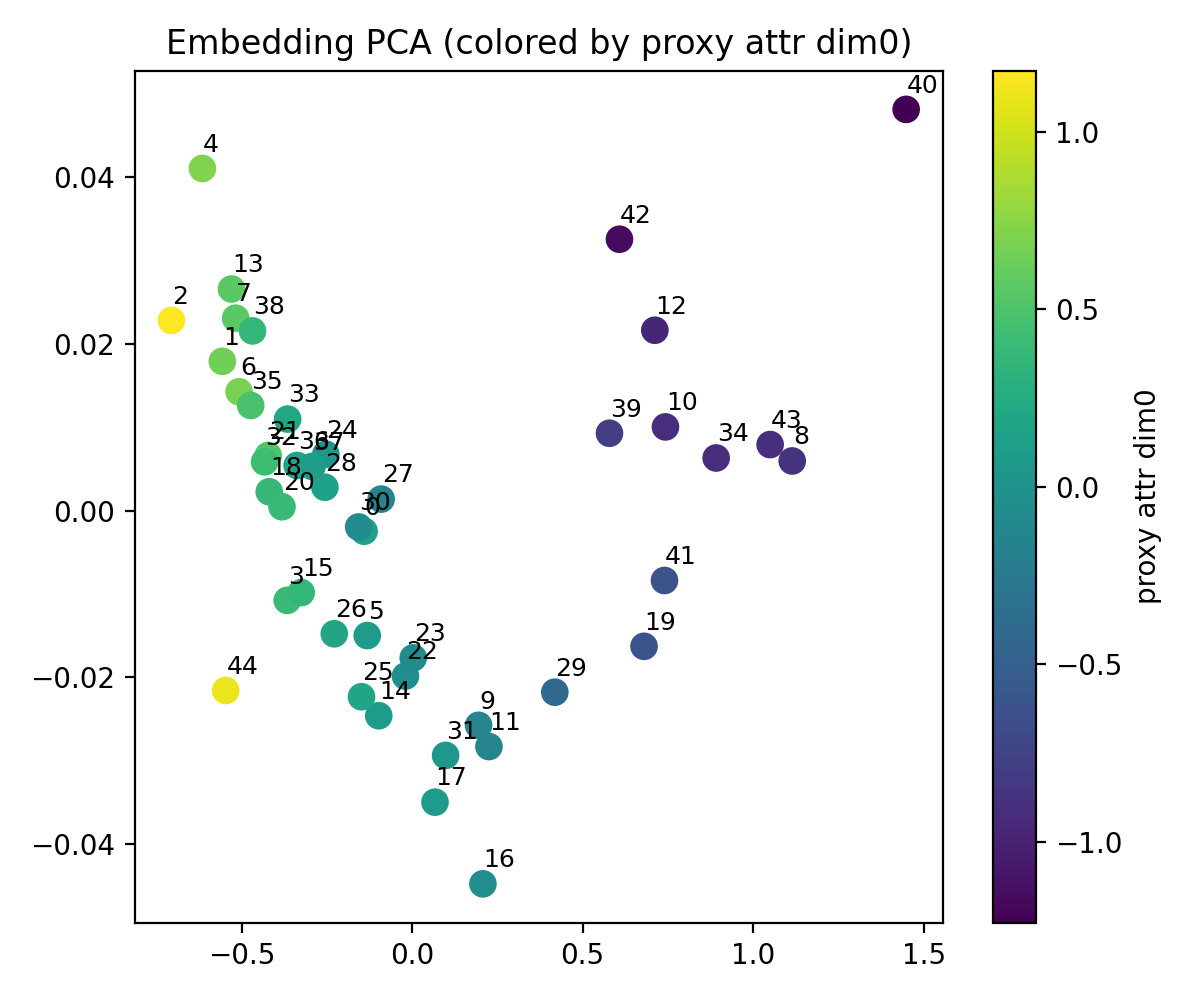}
        \includegraphics[width=\linewidth]{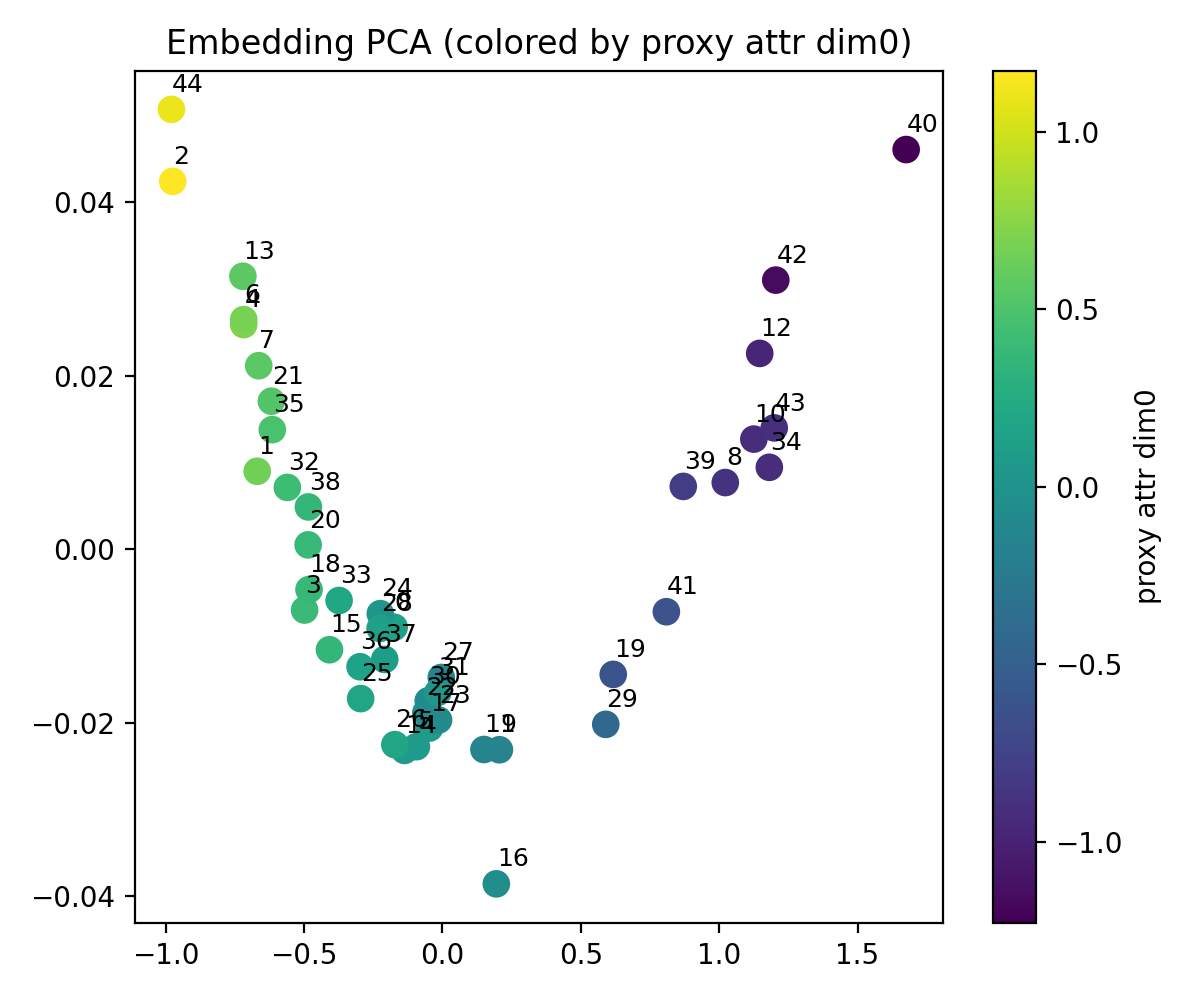}
        \includegraphics[width=\linewidth]{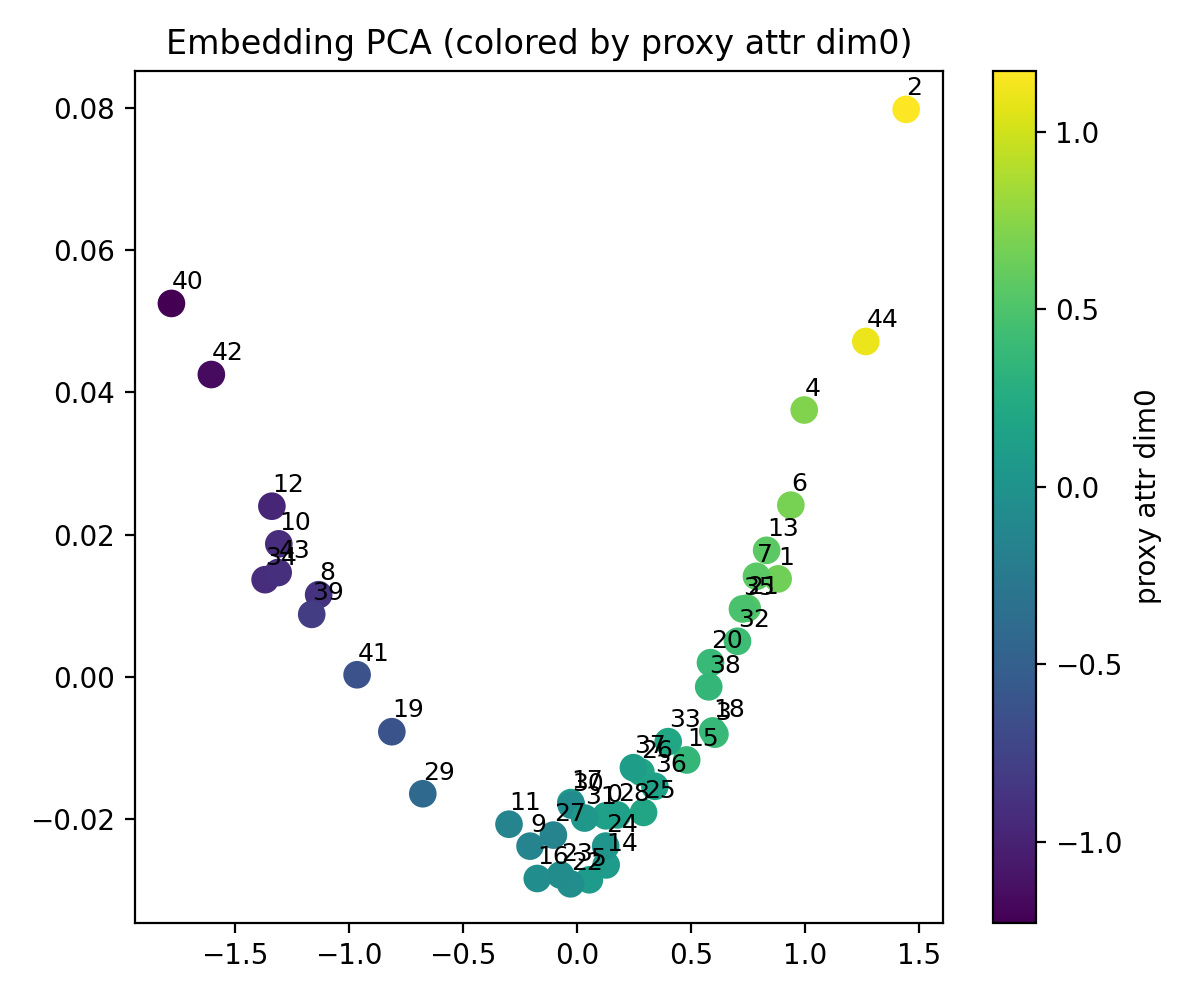}
        \subcaption{PCA embedding}
        \label{fig:emb_pca}
    \end{subfigure}
    \hfill
    \begin{subfigure}[t]{0.24\linewidth}
        \centering
        \includegraphics[width=\linewidth]{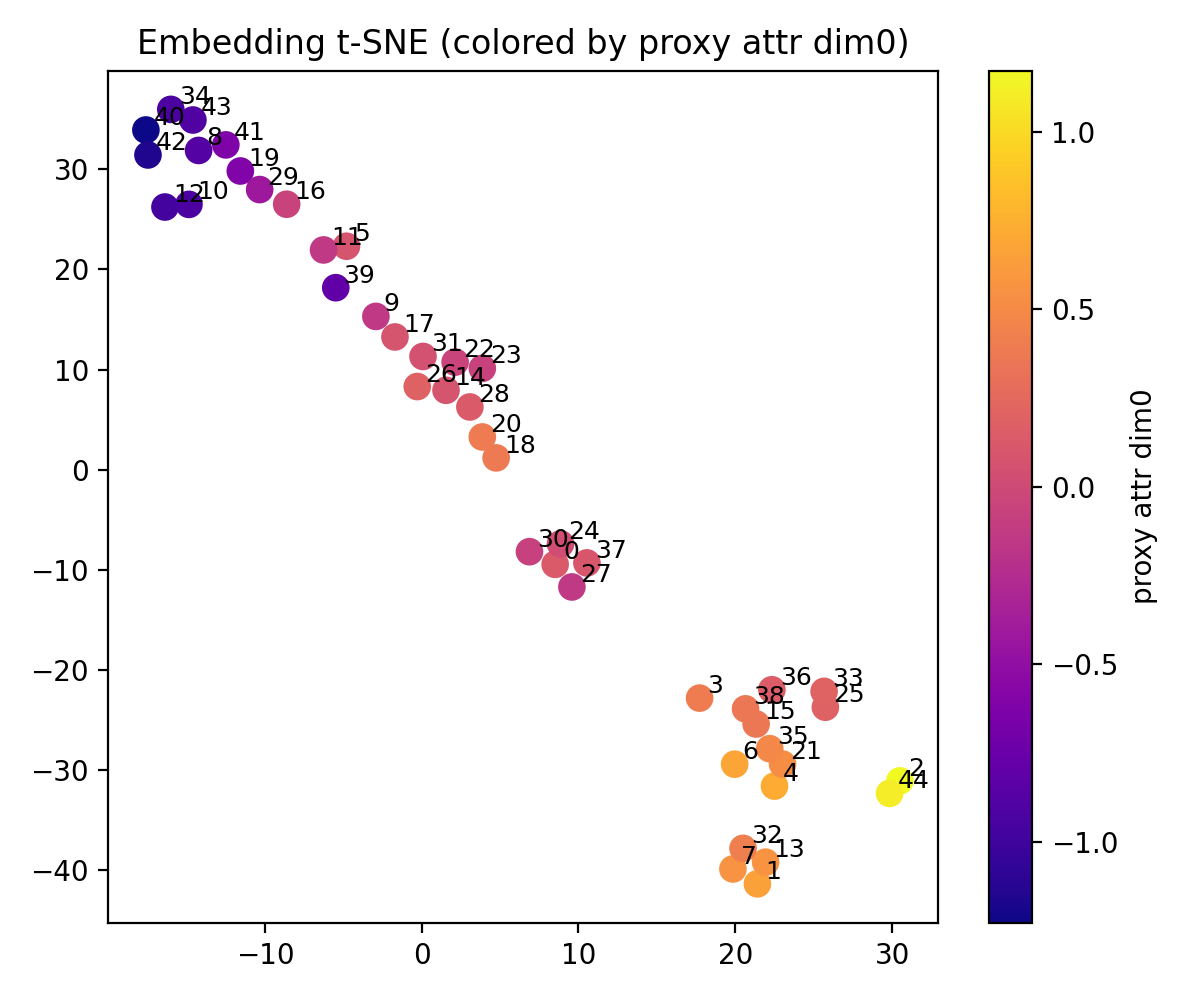}
        \includegraphics[width=\linewidth]{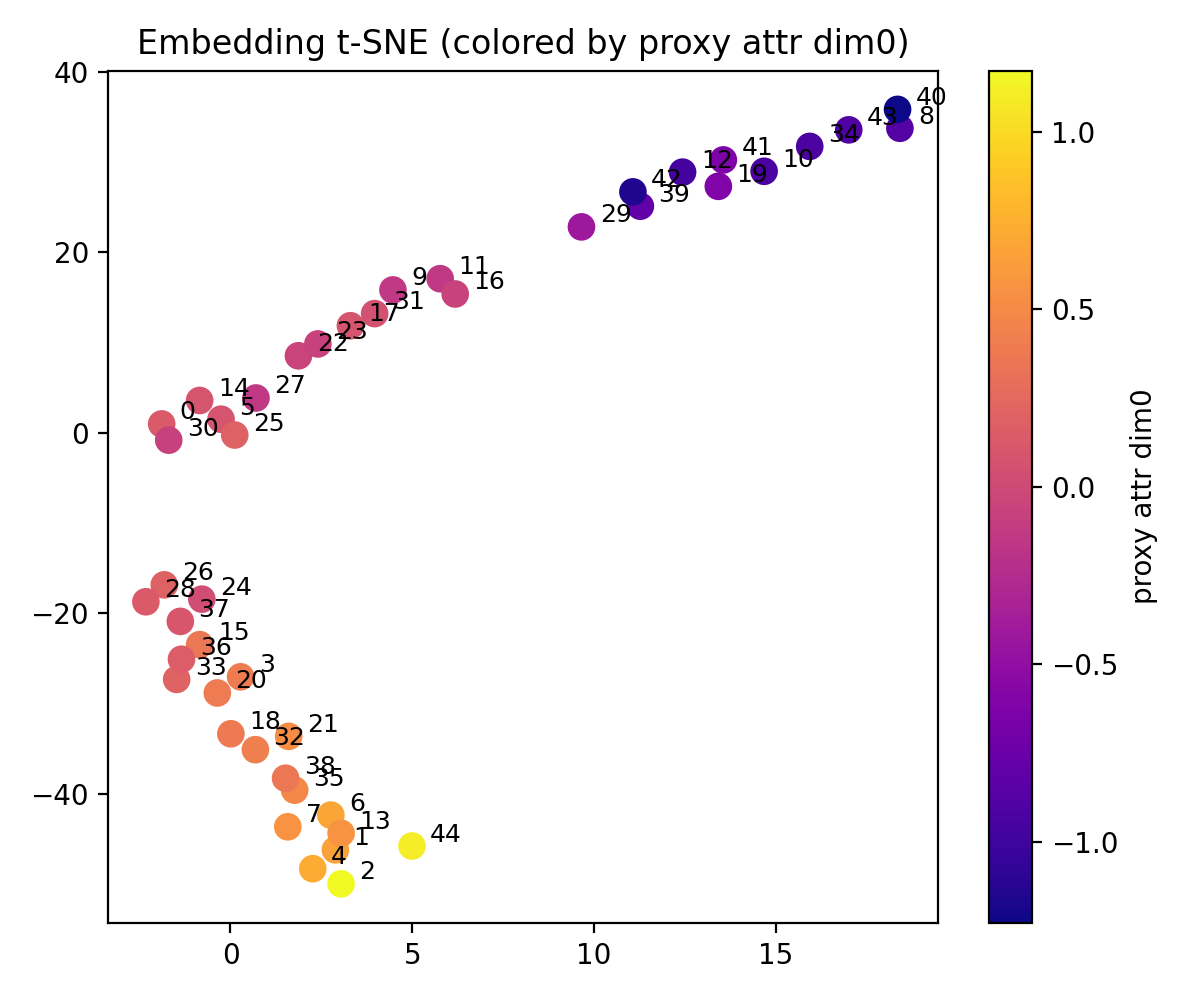}
        \includegraphics[width=\linewidth]{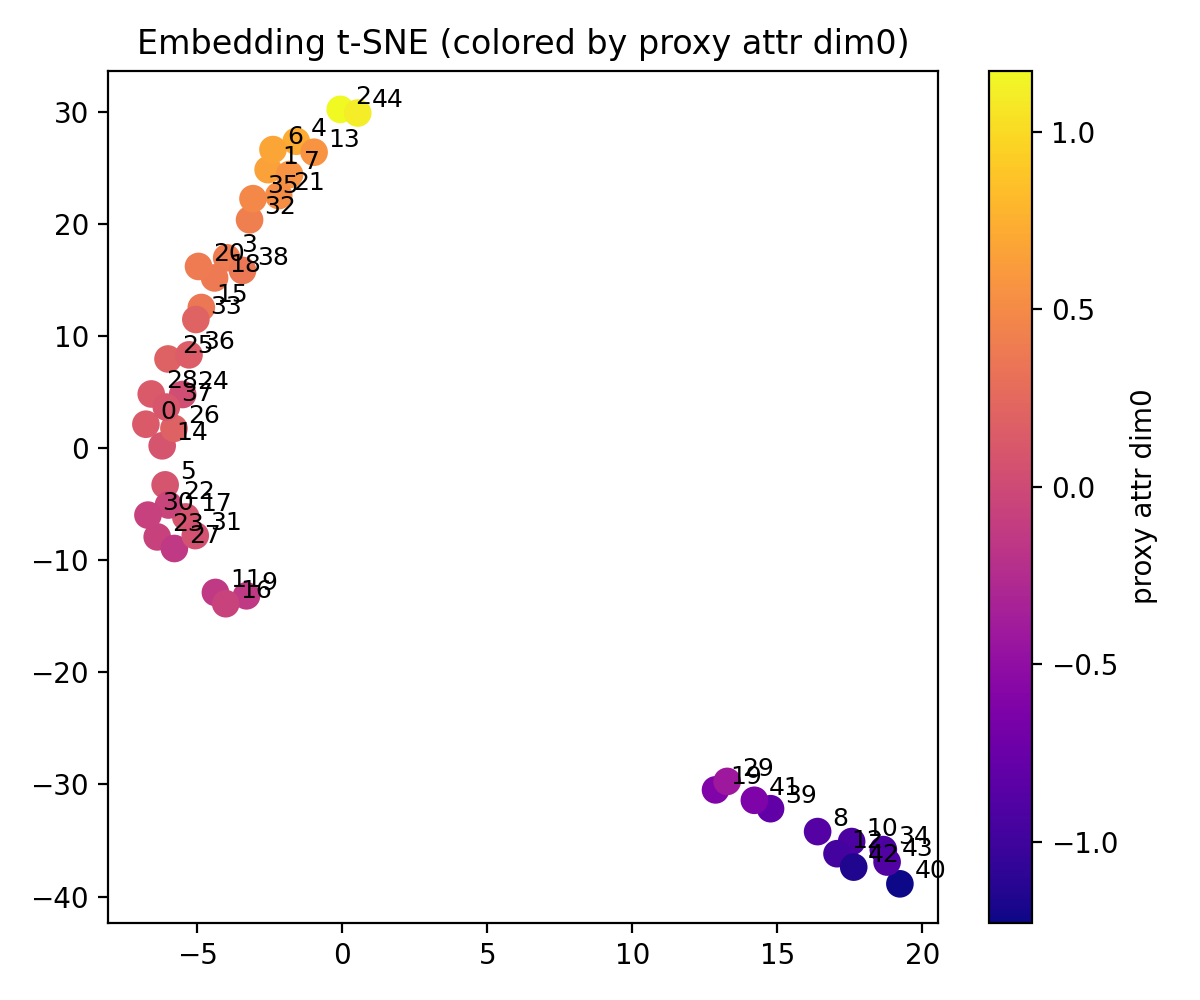}
        \includegraphics[width=\linewidth]{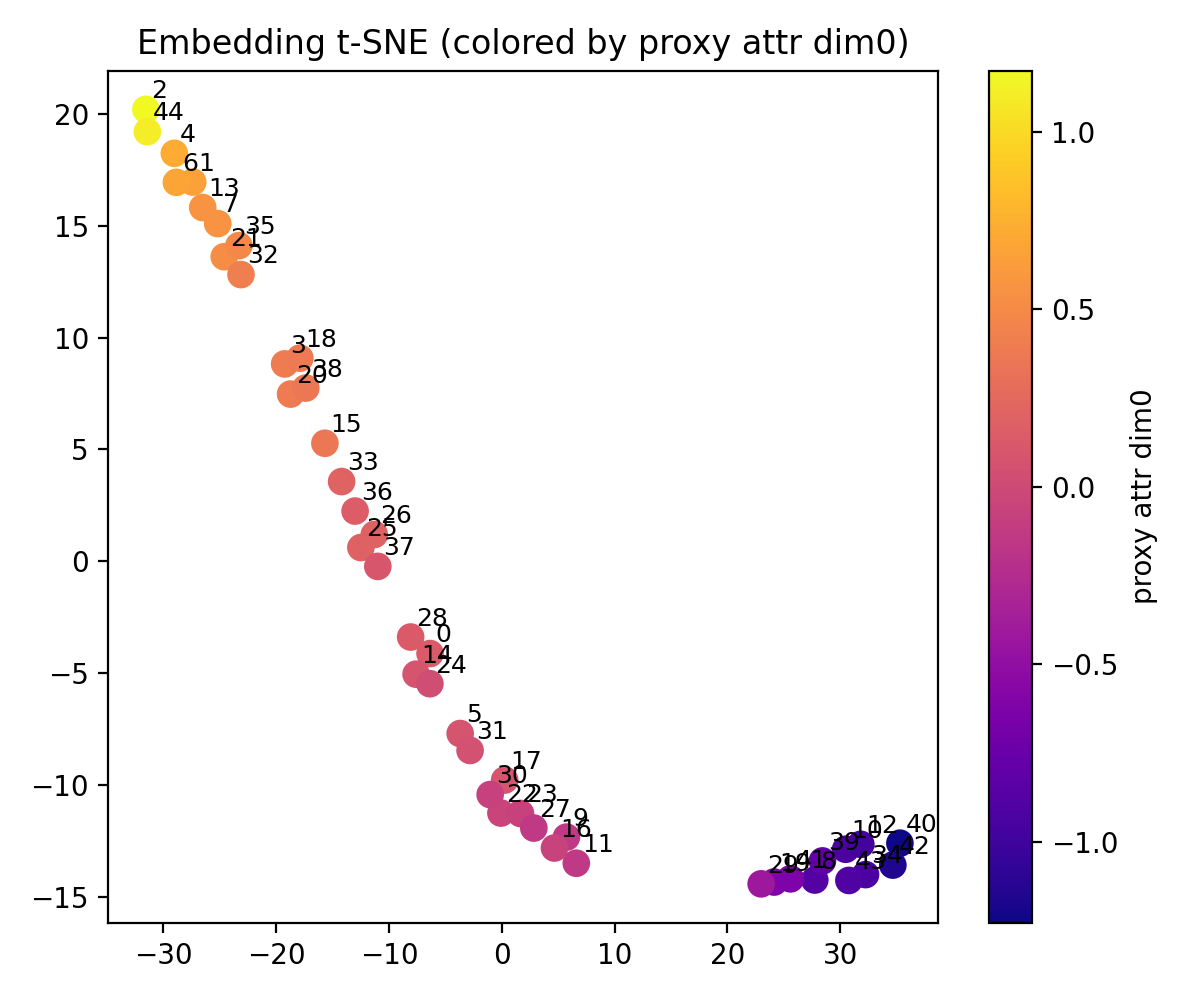}
        \subcaption{t-SNE embedding}
        \label{fig:emb_tsne}
    \end{subfigure}
    \hfill
    \begin{subfigure}[t]{0.2\linewidth}
        \centering
        \includegraphics[width=\linewidth]{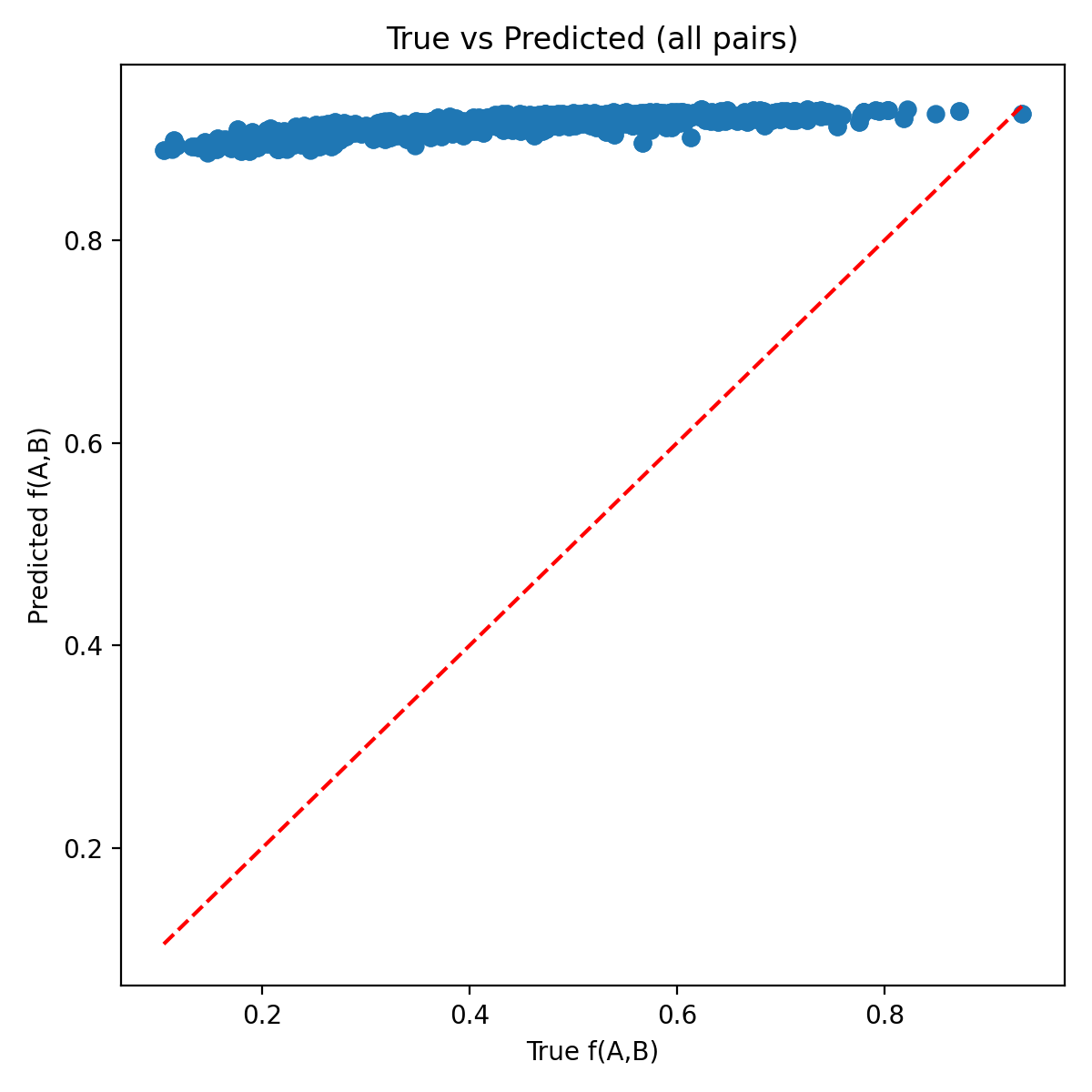}
        \includegraphics[width=\linewidth]{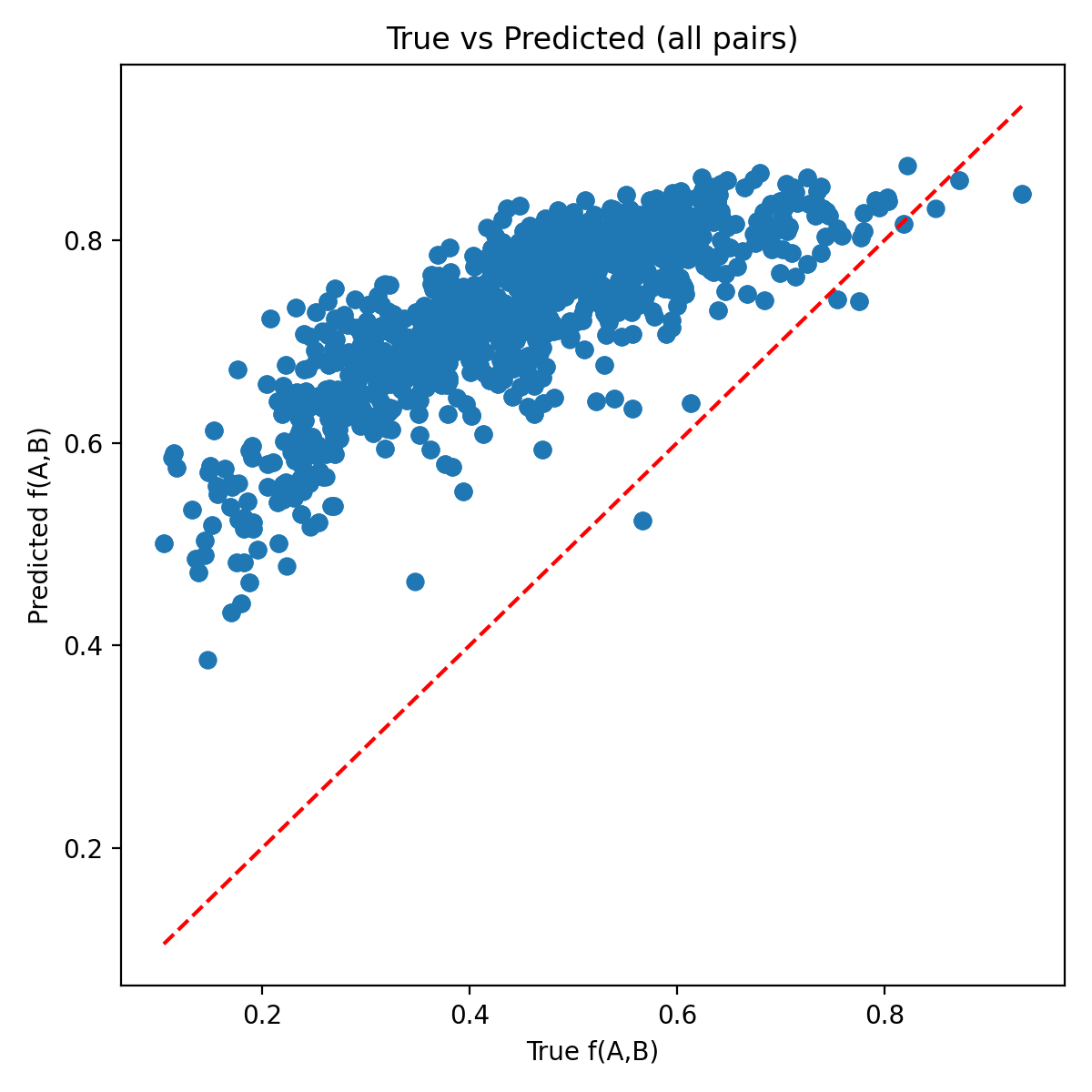}
        \includegraphics[width=\linewidth]{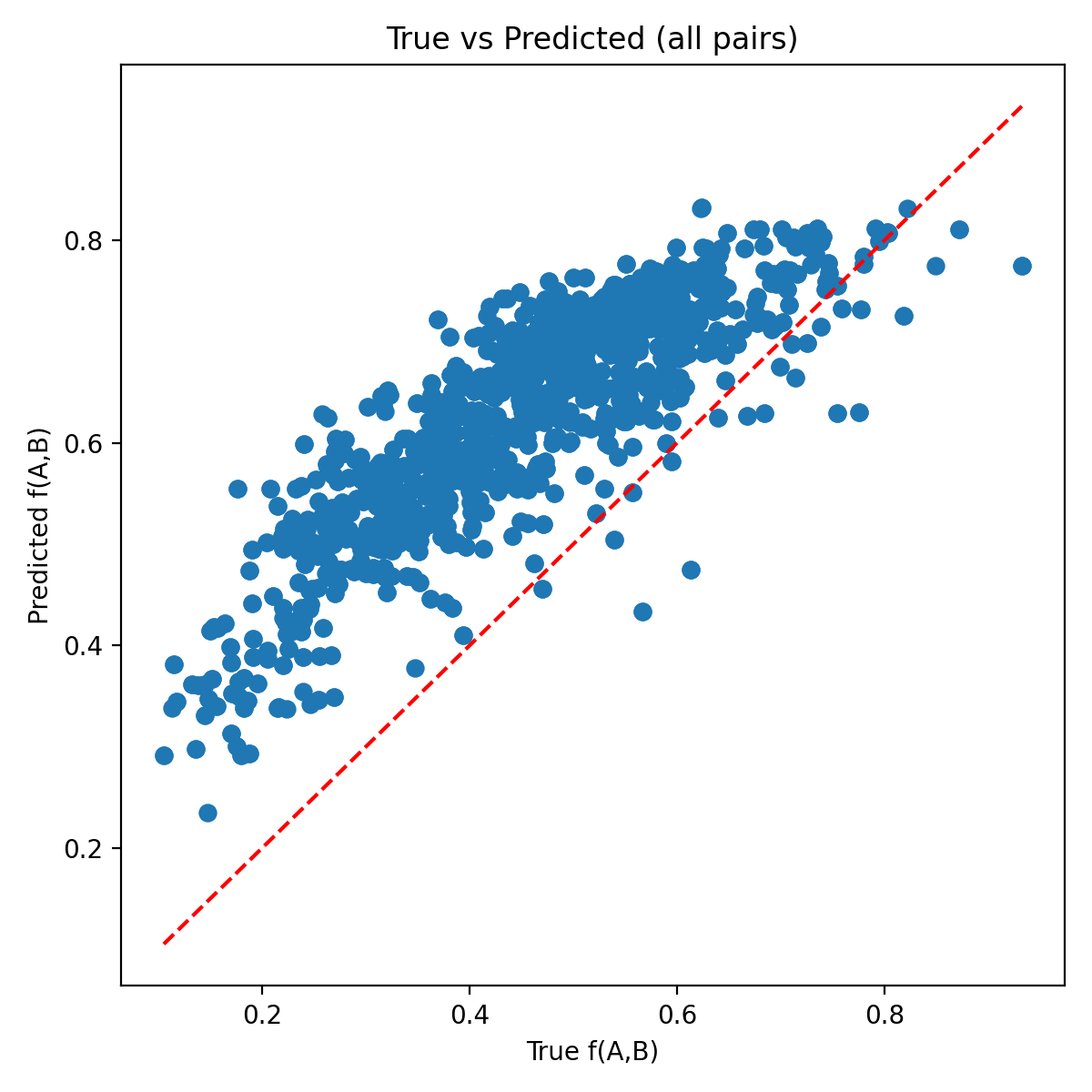}
        \includegraphics[width=\linewidth]{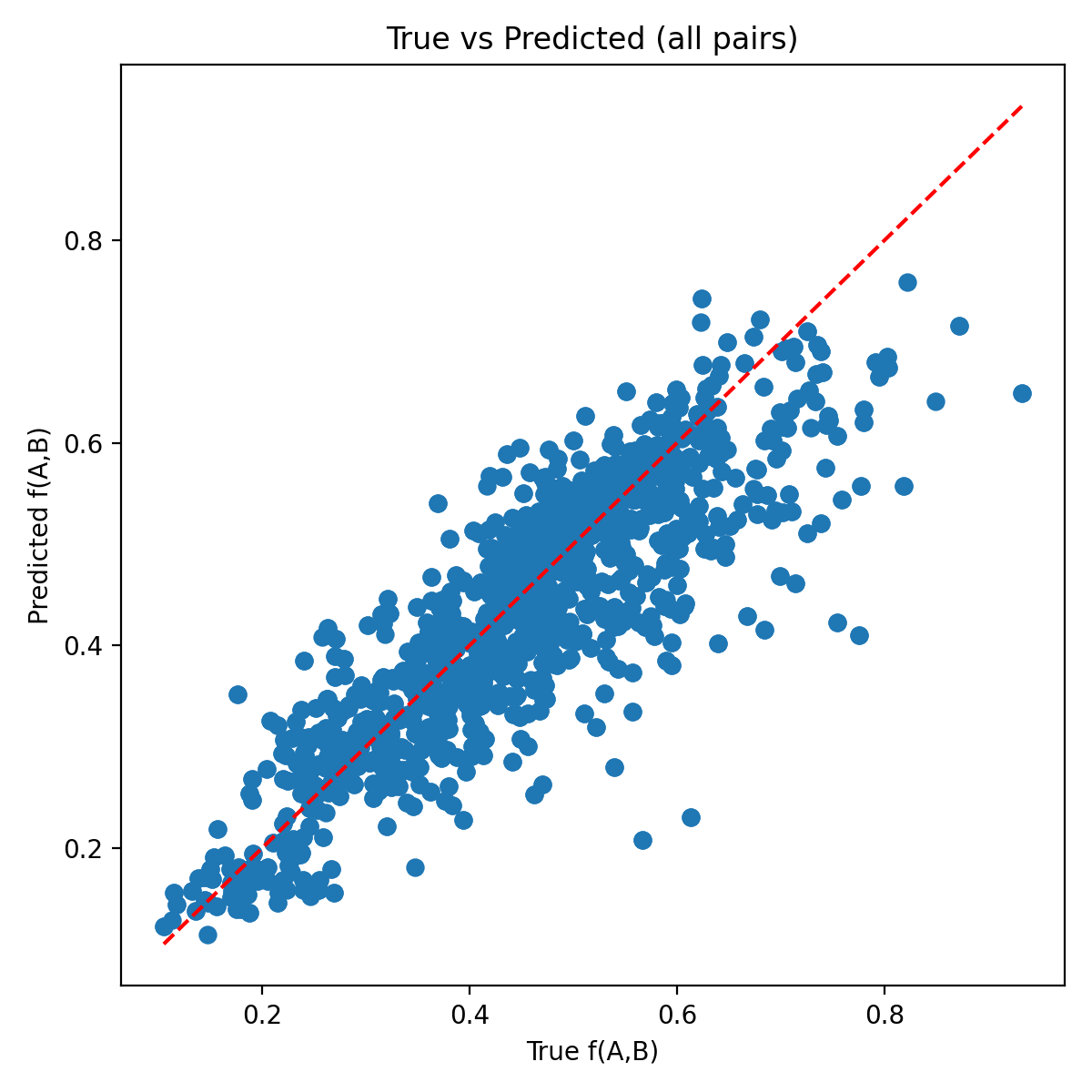}
        \subcaption{True vs. predicted}
        \label{fig:true_vs_pred}
    \end{subfigure}
    \begin{subfigure}[t]{0.3\linewidth}
        \centering
        \includegraphics[width=\linewidth]{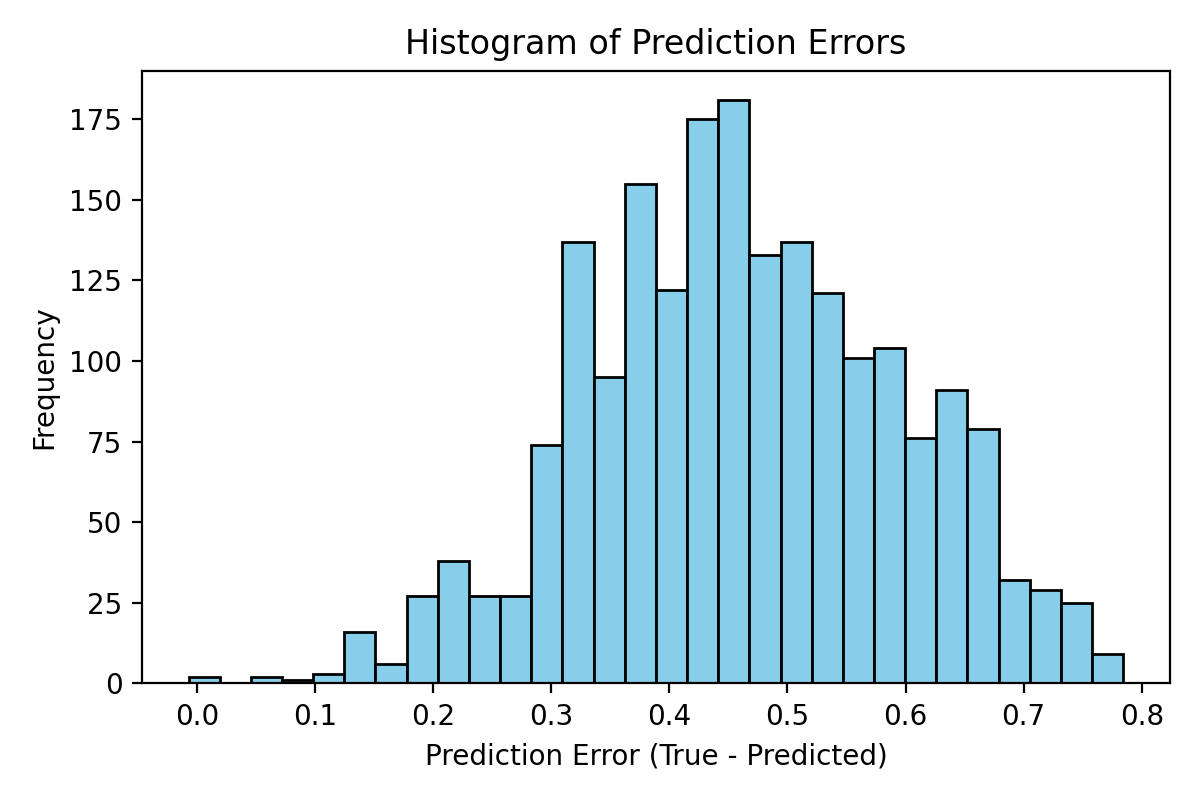}
        \includegraphics[width=\linewidth]{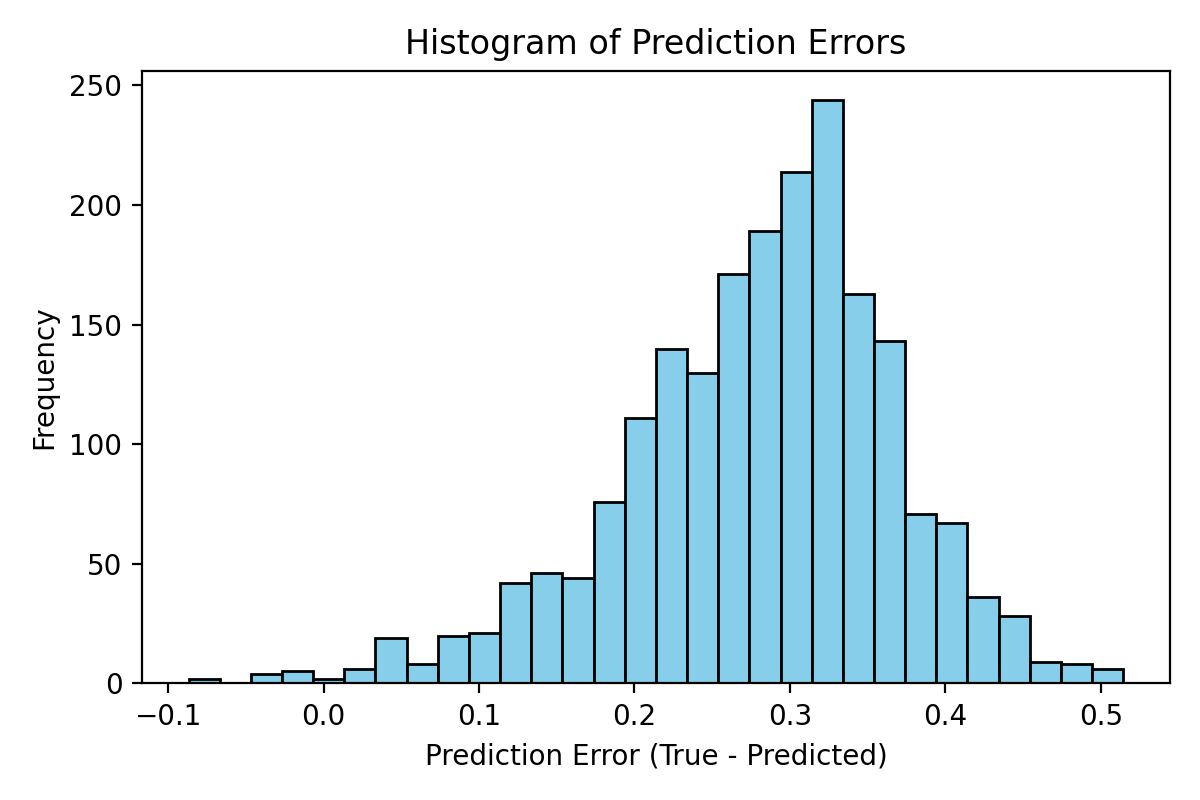}
        \includegraphics[width=\linewidth]{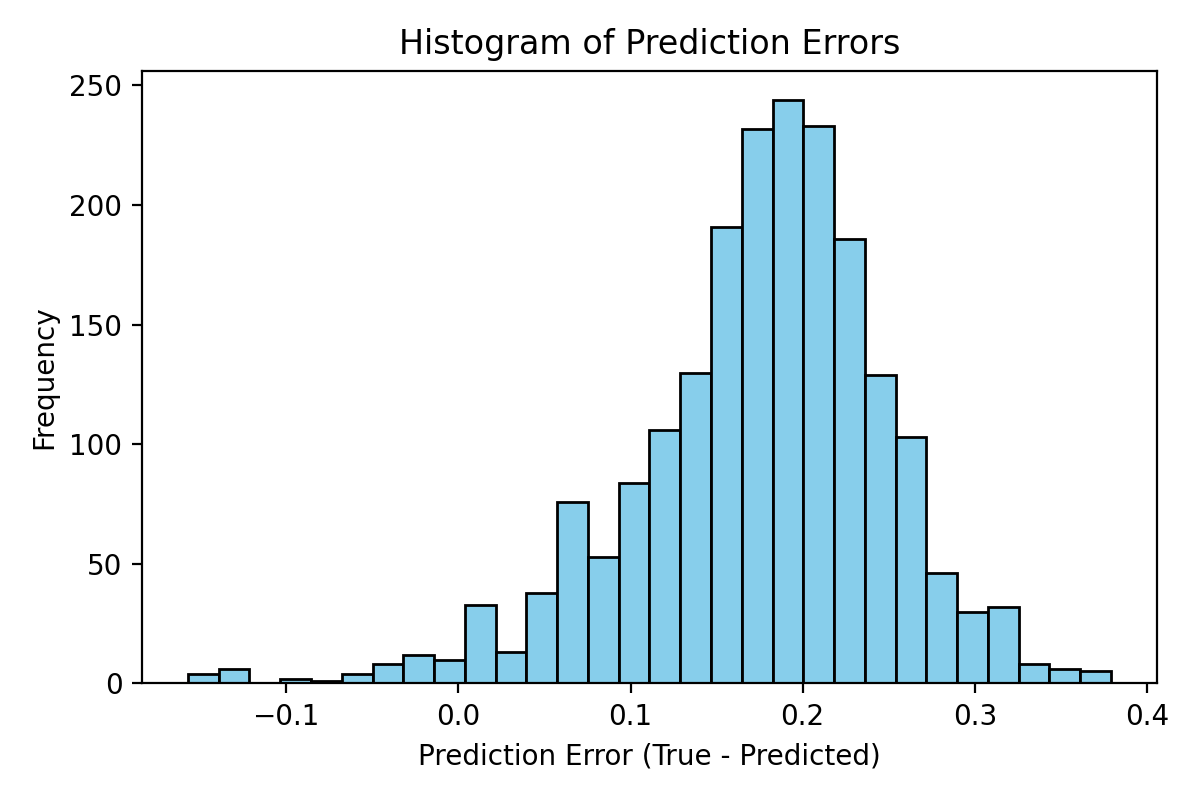}
        \includegraphics[width=\linewidth]{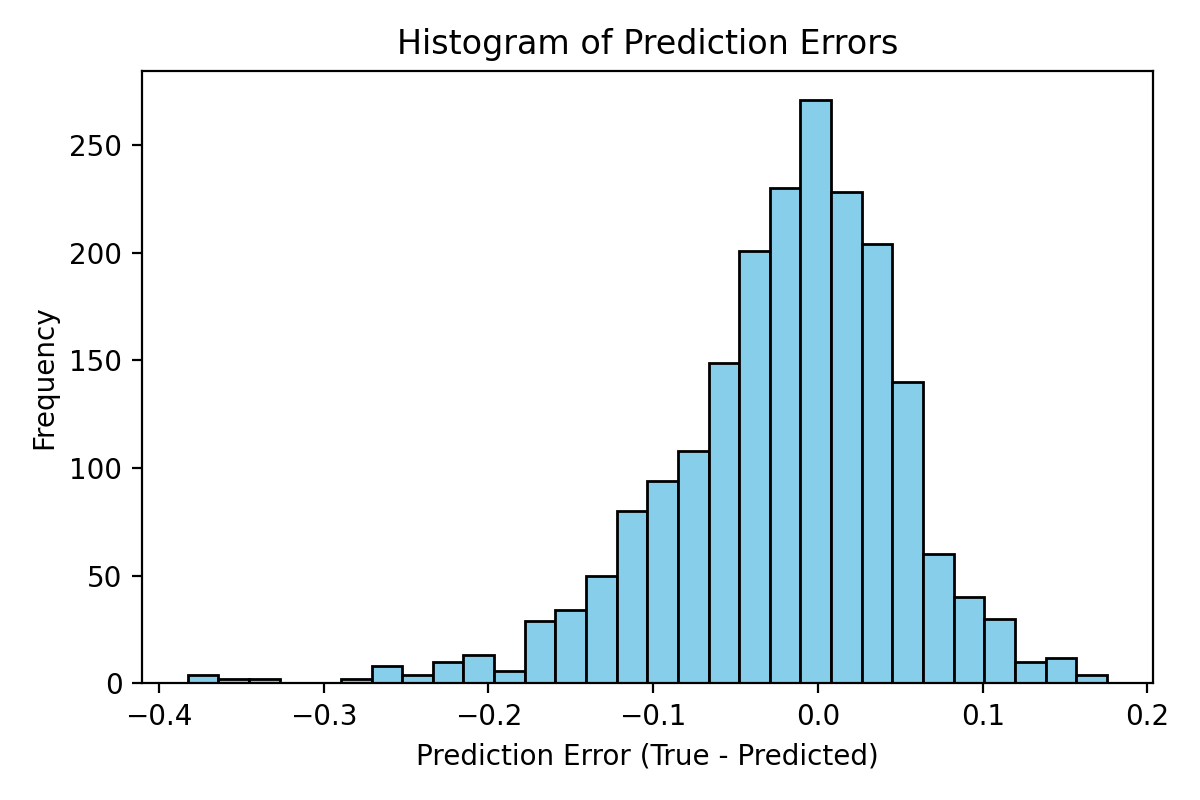}
        \subcaption{Error histogram}
        \label{fig:error_histogram}
    \end{subfigure}
    \caption{For kinetic friction, we also apply RRQR-based proxy selection under progressively stricter spectral retention thresholds (0.95, 0.99, 0.995, 0.999), producing proxy set cardinalities of 1, 5, 11, and 23, respectively. Larger thresholds (higher k) yield monotonic improvements in latent embedding cohesion and pairwise prediction fidelity, consistent with the baseline structure in Fig.~\ref{fig:embeddings_and_predictions-kinect}.}
    \label{fig:rrqr_proxies_kinect}
\end{figure*}

\begin{figure*}
    \centering
    \begin{subfigure}[t]{0.24\linewidth}
        \centering
        \includegraphics[width=\linewidth]{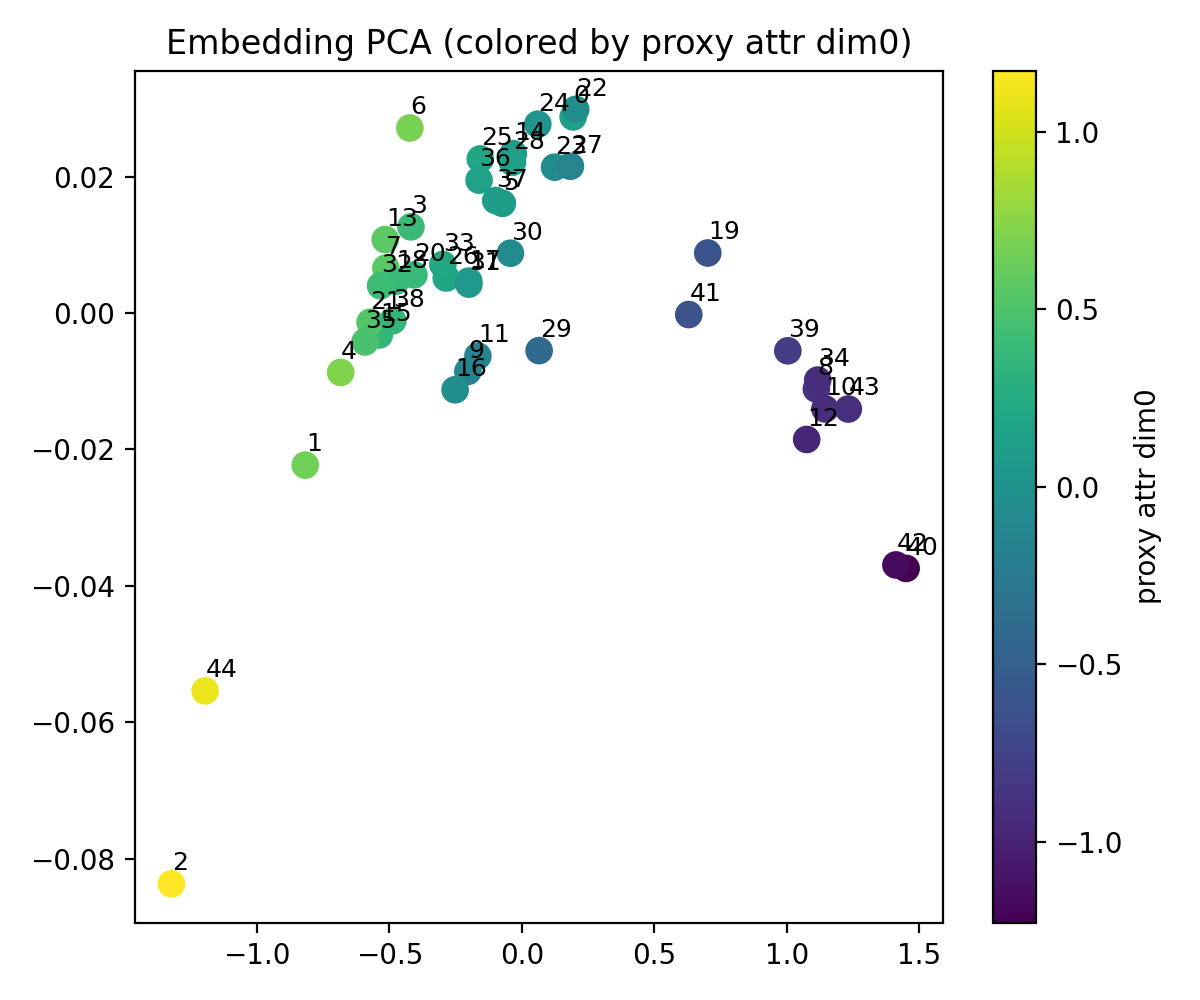}
        \subcaption{PCA embedding}
        \label{fig:emb_pca}
    \end{subfigure}
    \hfill
    \begin{subfigure}[t]{0.24\linewidth}
        \centering
        \includegraphics[width=\linewidth]{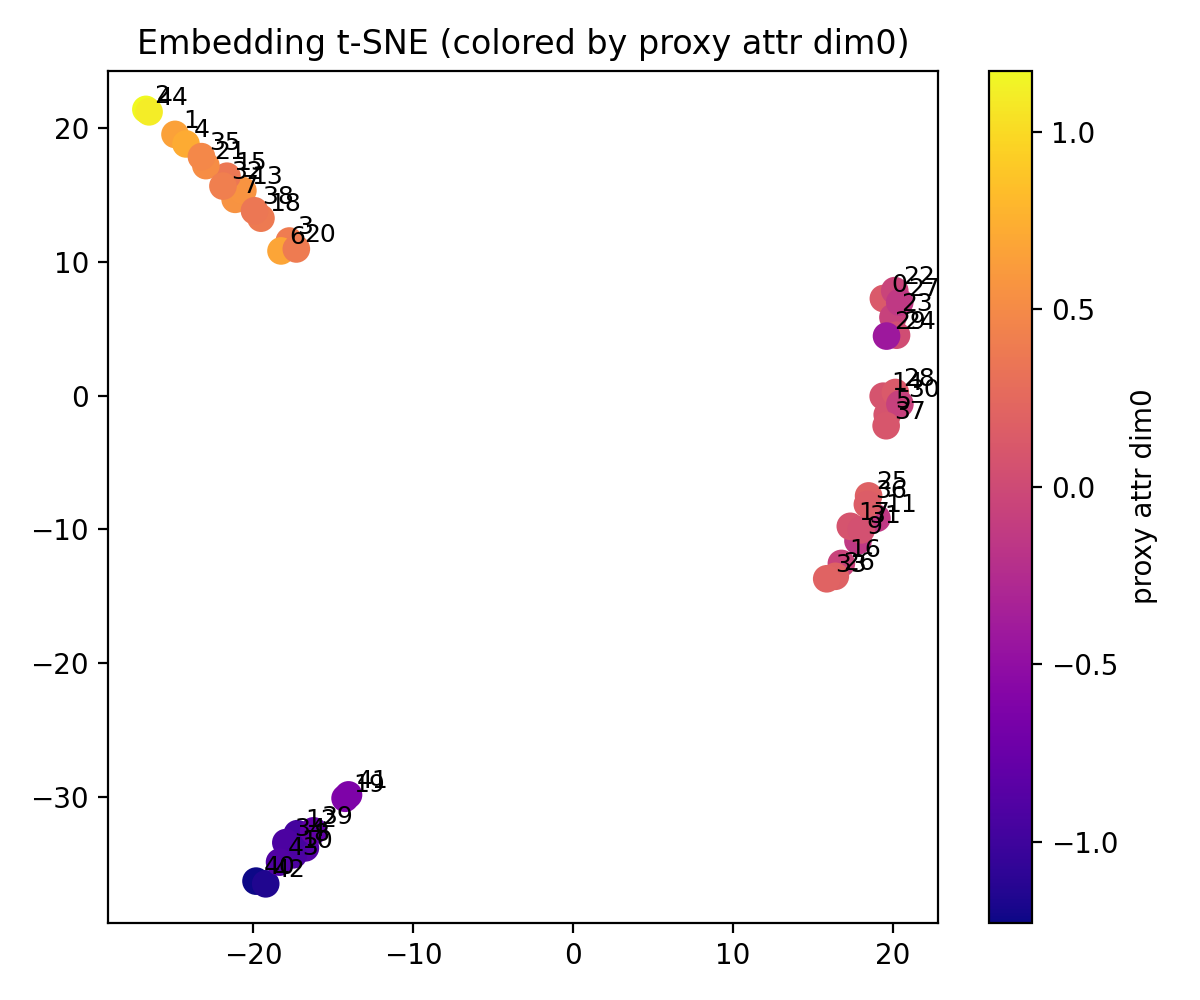}
        \subcaption{t-SNE embedding}
        \label{fig:emb_tsne}
    \end{subfigure}
    \hfill
    \begin{subfigure}[t]{0.2\linewidth}
        \centering
        \includegraphics[width=\linewidth]{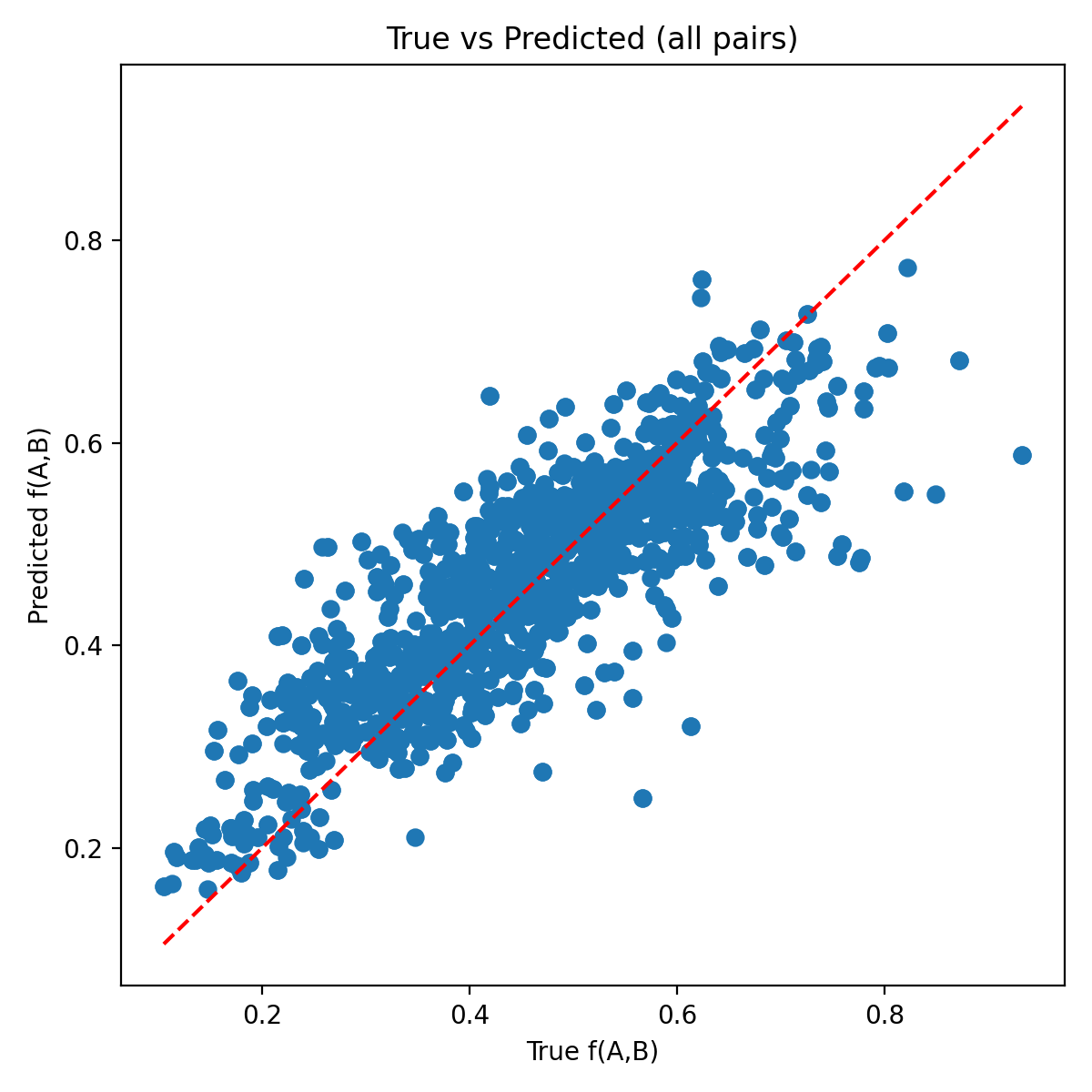}
        \subcaption{True vs. predicted}
        \label{fig:true_vs_pred}
    \end{subfigure}
    \begin{subfigure}[t]{0.3\linewidth}
        \centering
        \includegraphics[width=\linewidth]{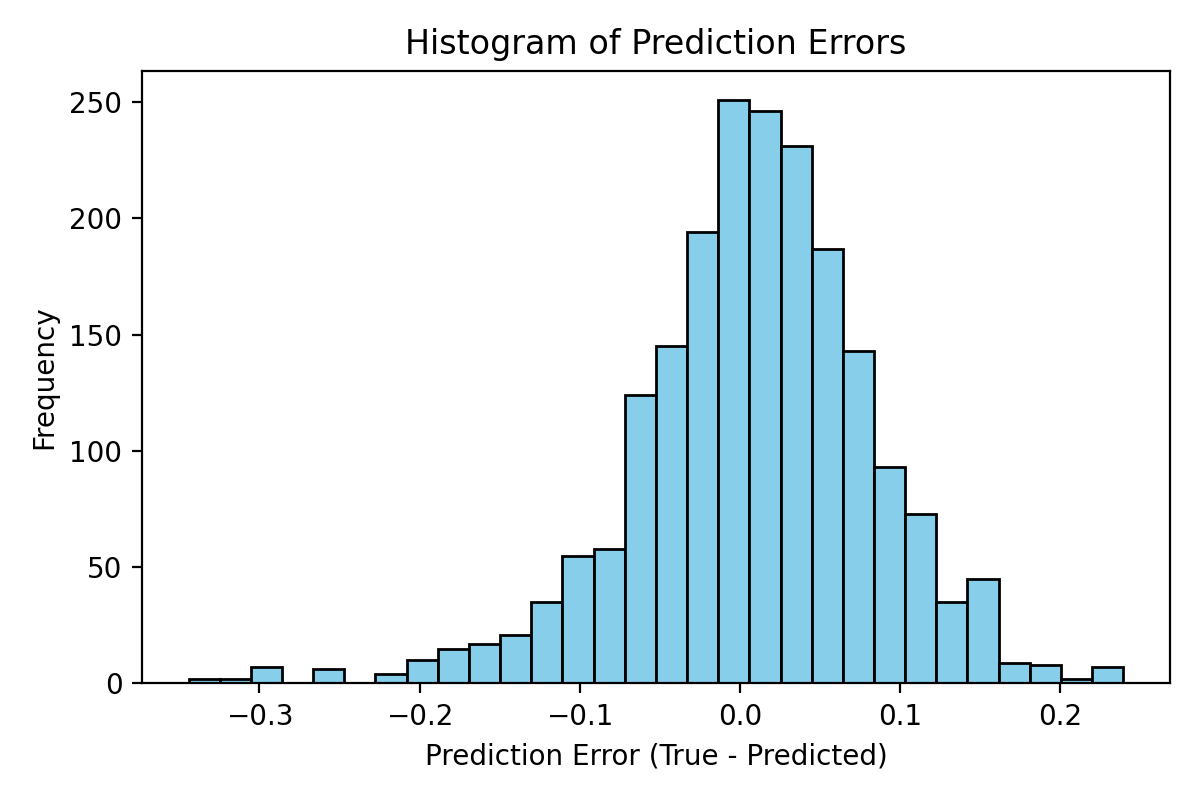}
        \subcaption{Error histogram}
        \label{fig:error_histogram}
    \end{subfigure}
    \caption{Leveraging the latent space learned from kinetic-friction data, the mask-optimization method selects an optimal proxy set comprising two materials (indices 24 and 26, zero-based), identical to the set selected for the static-friction regime.}
    \label{fig:mask_proxies_kinect}
\end{figure*}

\begin{figure*}
    \centering
    \begin{subfigure}[t]{0.395\linewidth}
        \centering
        \includegraphics[width=\linewidth]{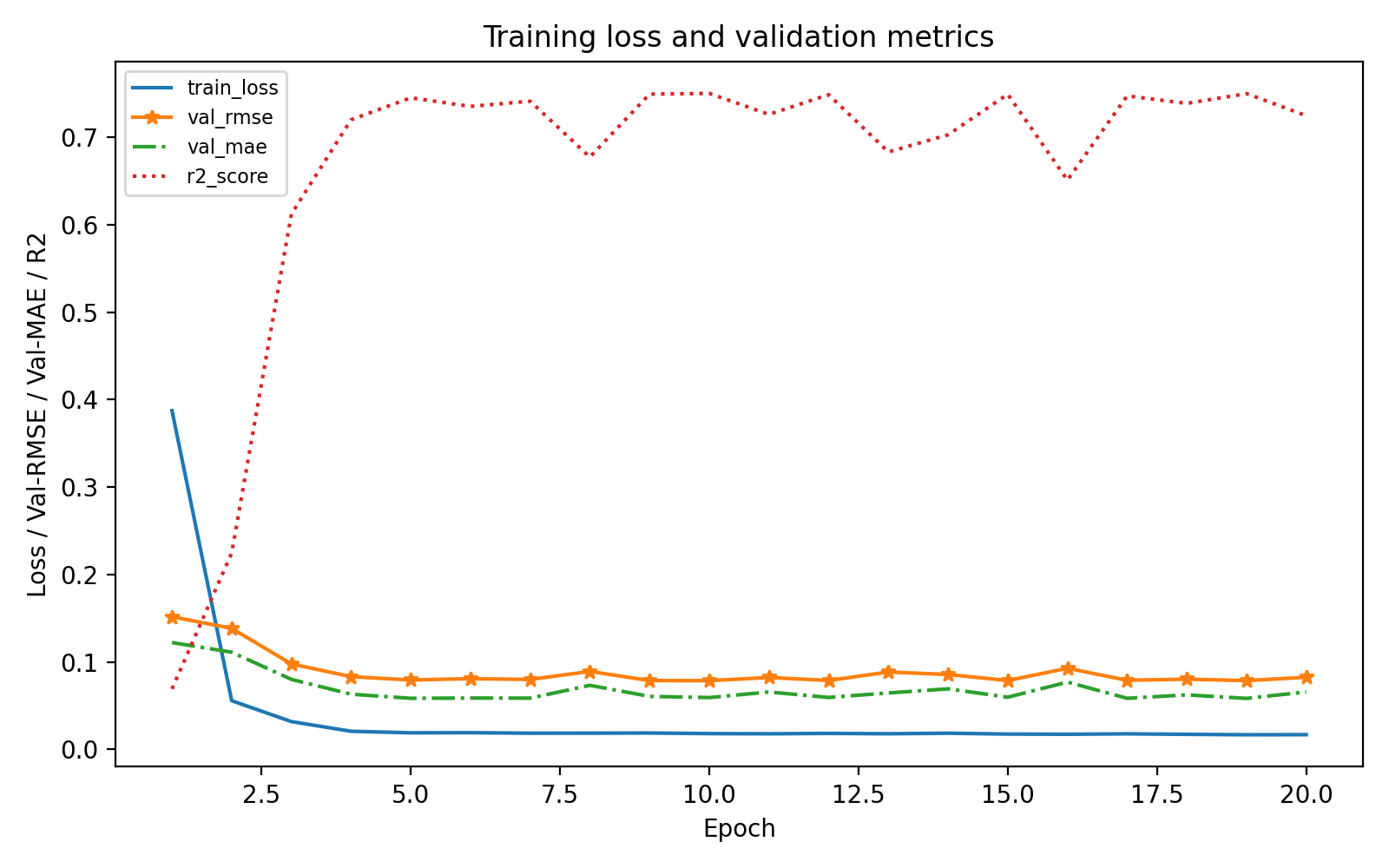}
        \subcaption{Training and validation loss curves}   
        \label{fig:true_vs_pred}
    \end{subfigure}
    \hfill
    \begin{subfigure}[t]{0.295\linewidth}
        \centering
        \includegraphics[width=\linewidth]{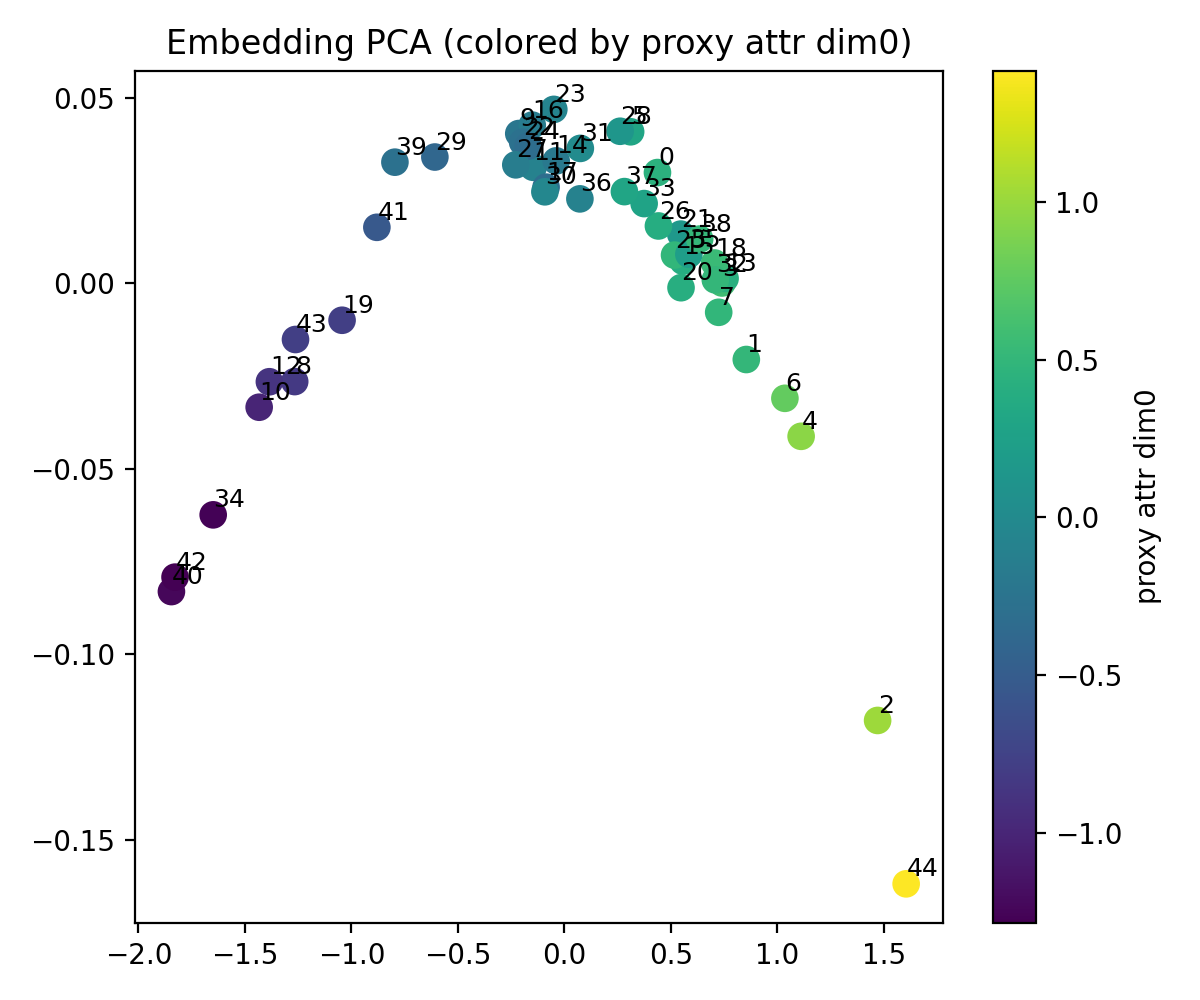}
        \subcaption{PCA embedding}
        \label{fig:emb_pca}
    \end{subfigure}
    \hfill
    \begin{subfigure}[t]{0.295\linewidth}
        \centering
        \includegraphics[width=\linewidth]{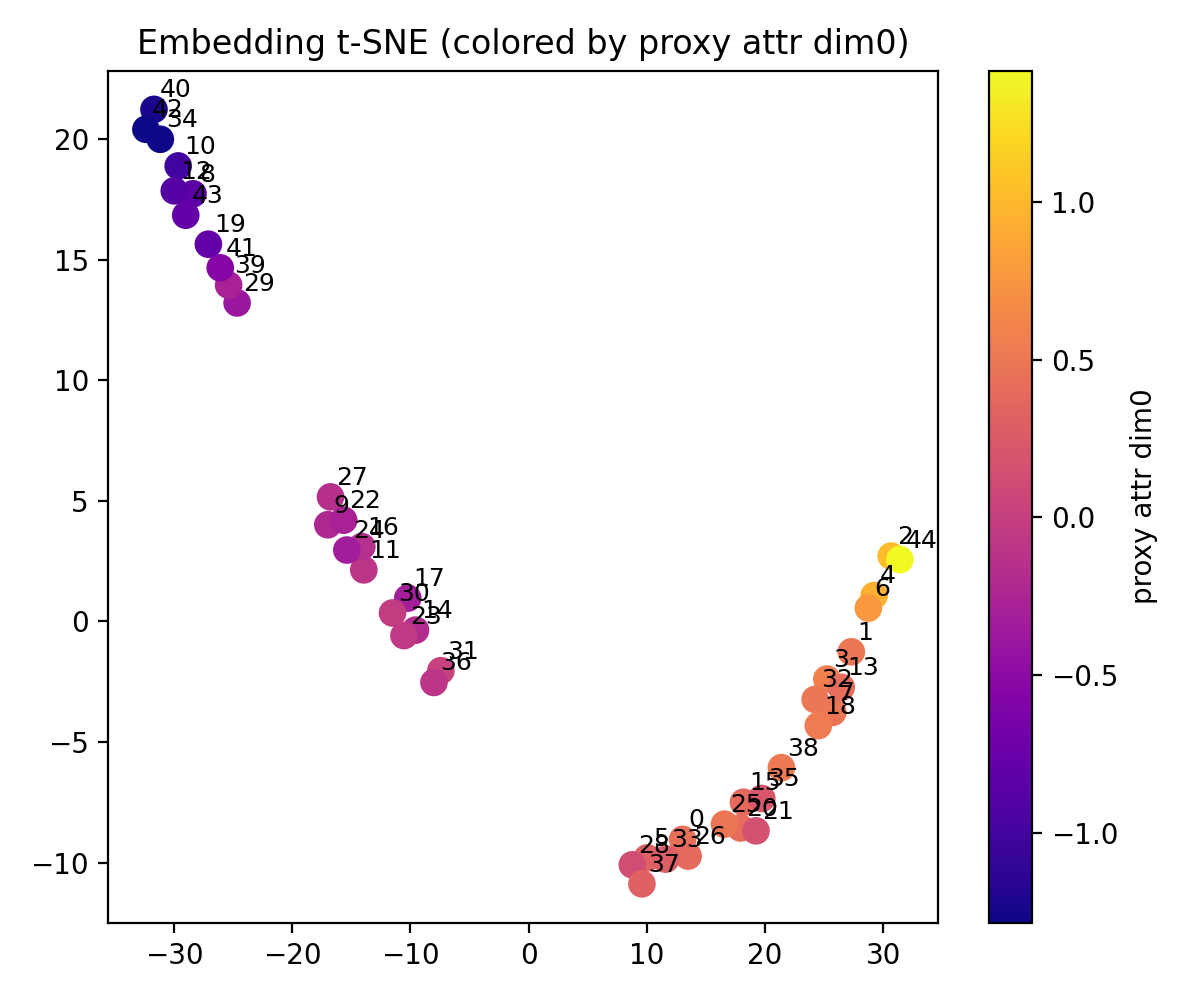}
        \subcaption{t-SNE embedding}
        \label{fig:emb_tsne}
    \end{subfigure}
    \hfill
    \begin{subfigure}[t]{0.195\linewidth}
        \centering
        \includegraphics[width=\linewidth]{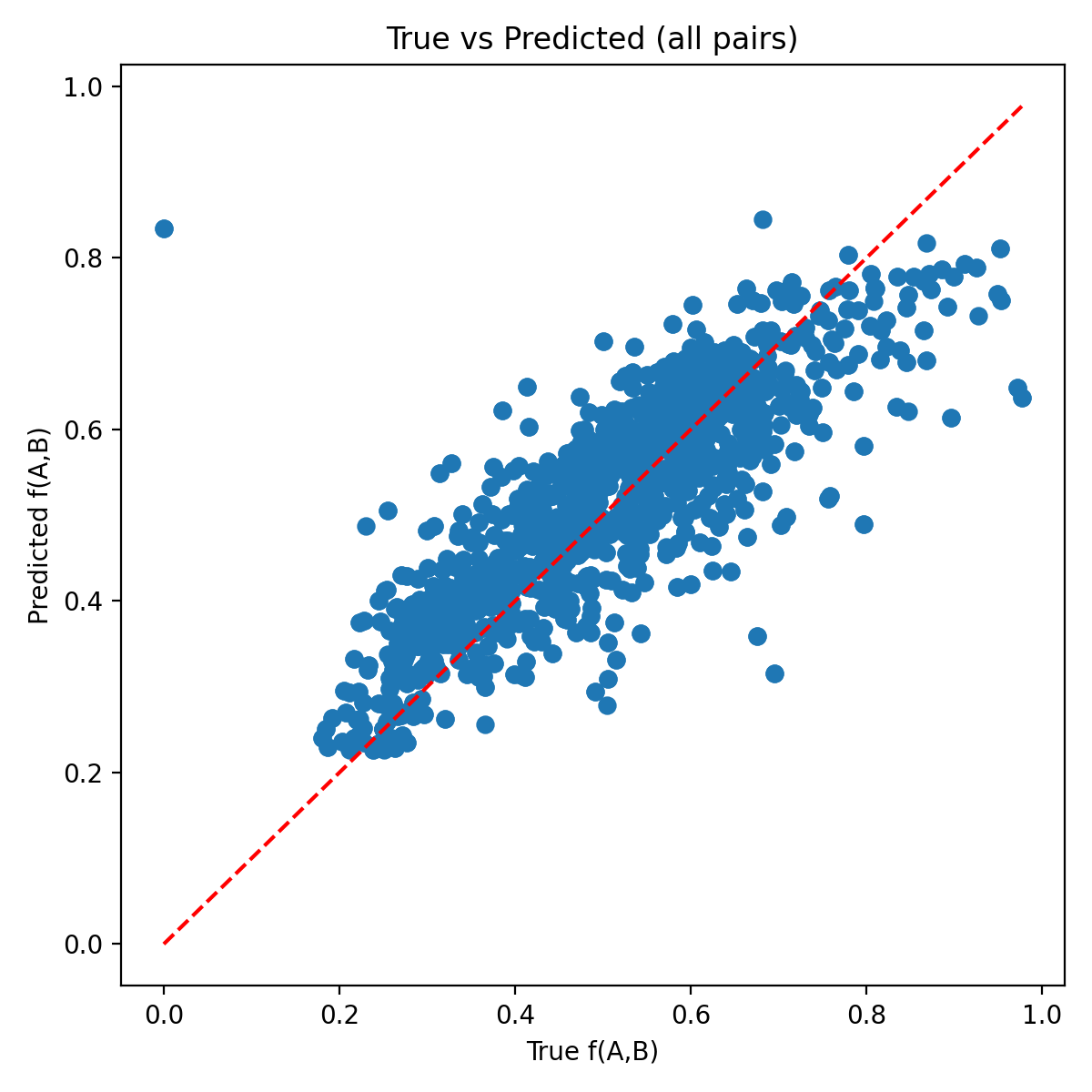}
        \subcaption{True vs. predicted: static}
        \label{fig:emb_pca}
    \end{subfigure}
    \hfill
    \begin{subfigure}[t]{0.295\linewidth}
        \centering
        \includegraphics[width=\linewidth]{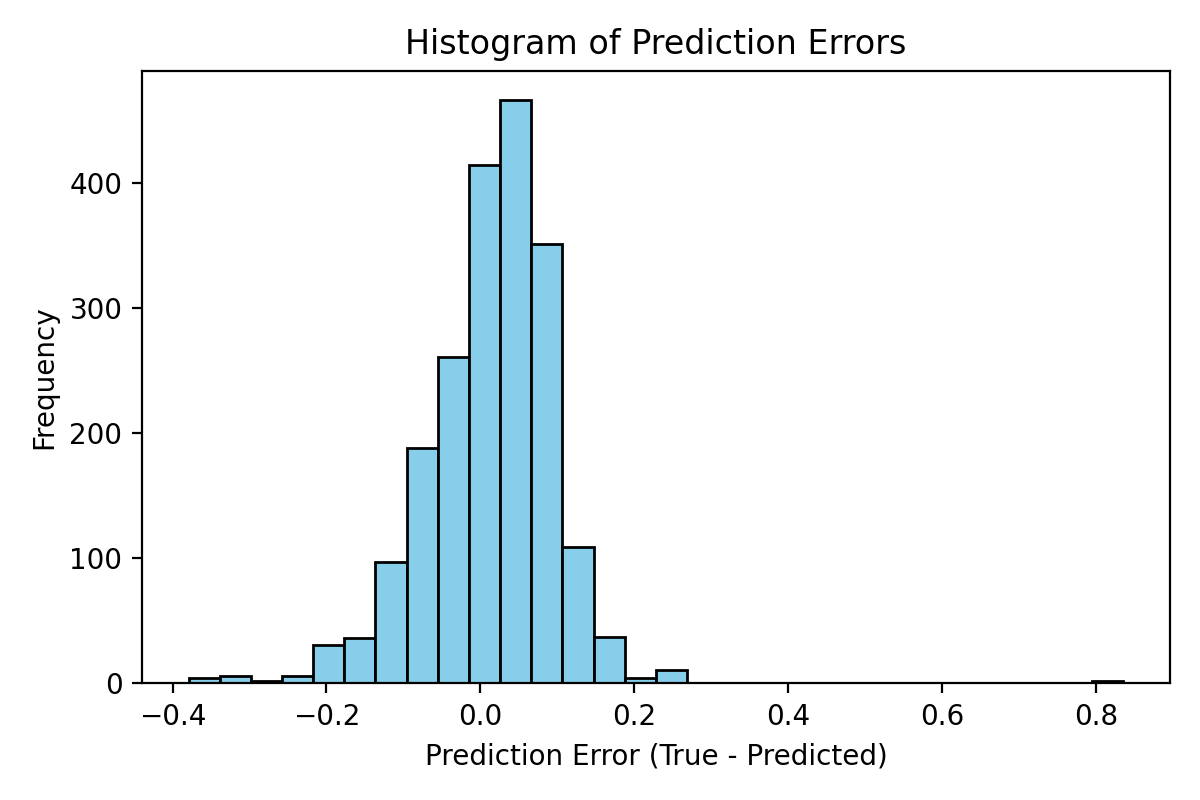}
        \subcaption{Error histogram: static}
        \label{fig:emb_tsne}
    \end{subfigure}
    \hfill
    \begin{subfigure}[t]{0.195\linewidth}
        \centering
        \includegraphics[width=\linewidth]{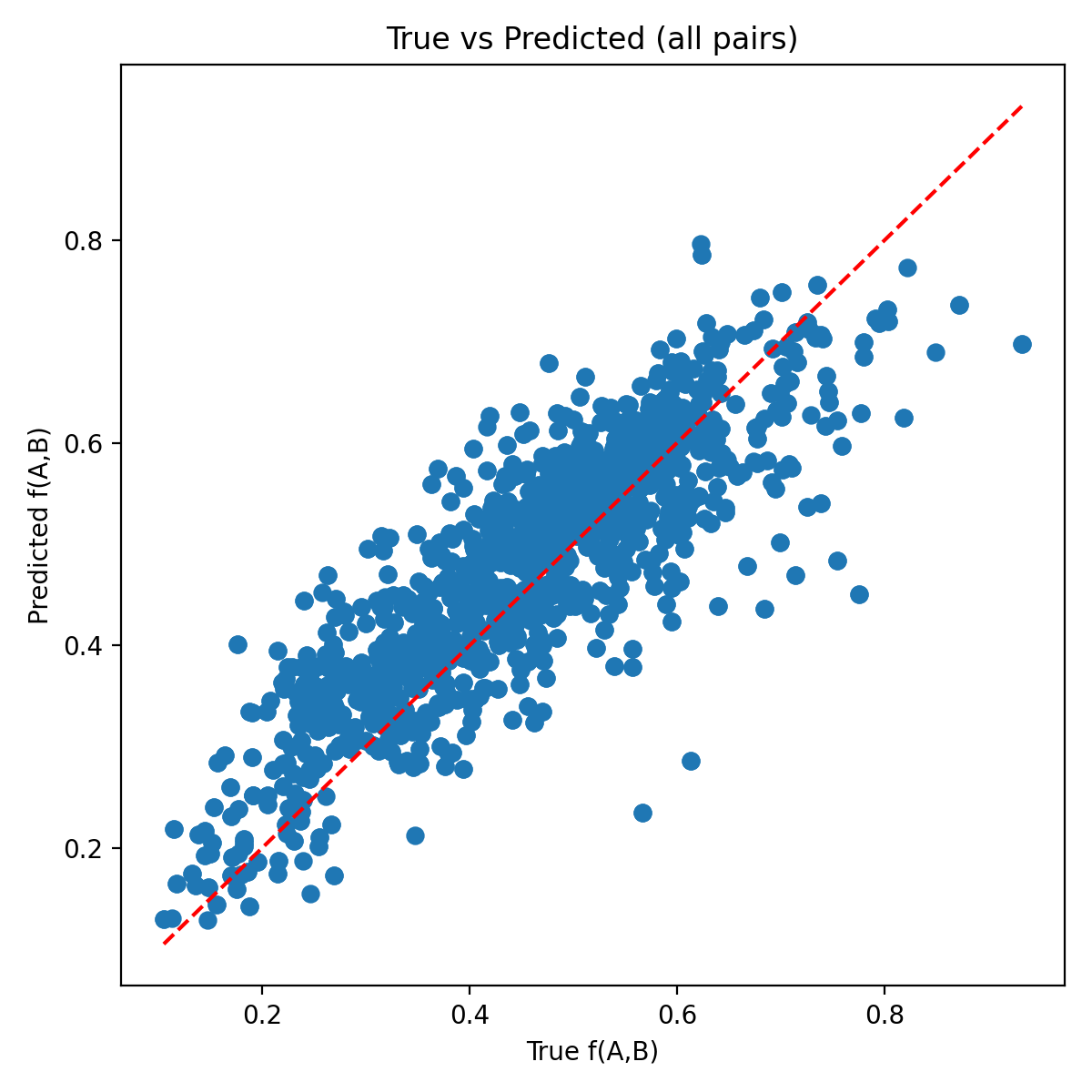}
        \subcaption{True vs. predicted: kinetic}
        \label{fig:true_vs_pred}
    \end{subfigure}
    \hfill
    \begin{subfigure}[t]{0.295\linewidth}
        \centering
        \includegraphics[width=\linewidth]{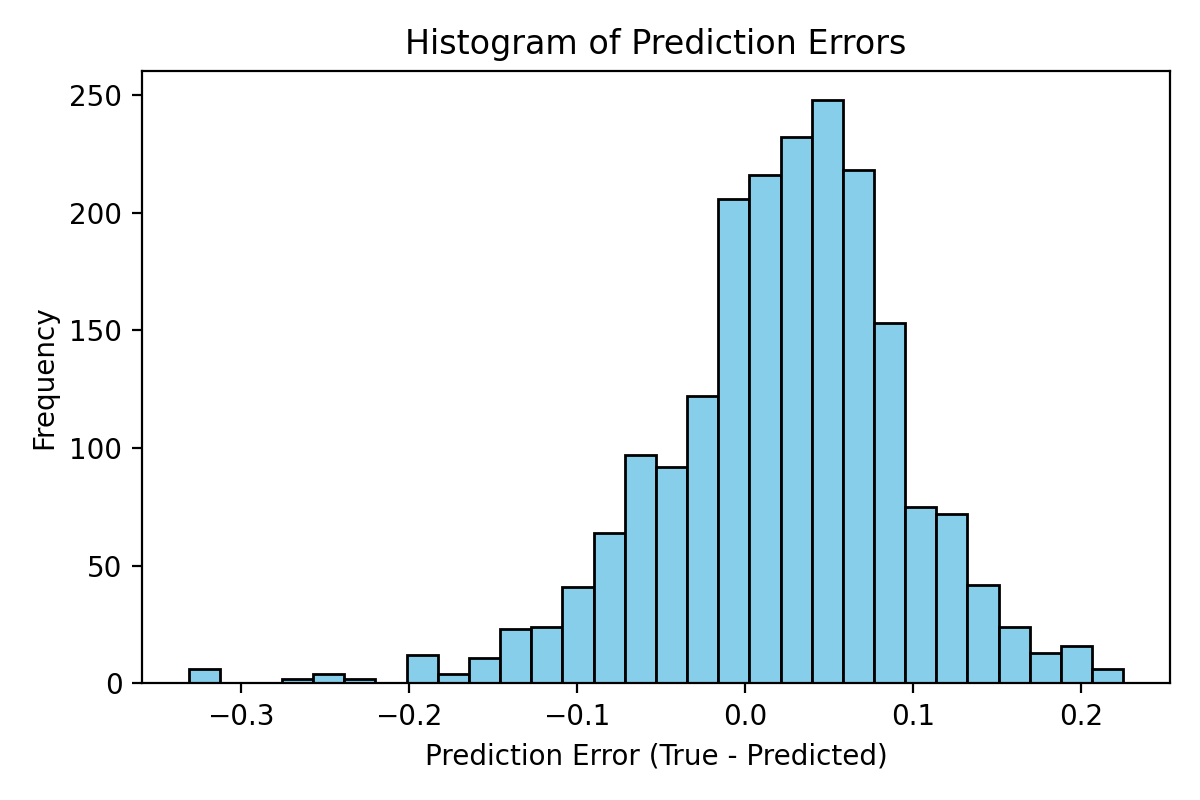}
        \subcaption{Error histogram: kinetic}
        \label{fig:error_histogram}
    \end{subfigure}
    \caption{Joint embedding model trained simultaneously on isotropic static and kinetic friction matrices. (a) Training and validation loss curves exhibit stable, monotonic convergence without evidence of overfitting. (b,c) PCA and t-SNE projections of the shared material embeddings closely replicate the structural organization obtained from regime-specific models. (d,e) Static true–vs.–predicted scatter and residual histogram show high linear correlation with tightly concentrated errors. (f,g) Kinetic counterparts display analogous fidelity. Collectively, the results indicate that a unified latent representation captures regime-dependent frictional variability with high accuracy.}
    \label{fig:joint_embedding_static_kinetic}
\end{figure*}

\paragraph{Kinetic Friction Model}
In addition, we developed a dedicated kinetic friction model that incorporates the unique features of kinetic friction behavior. This model was trained on the kinetic friction matrix $\mathbf{F}_{\text{iso-k}}$ and aims to provide more accurate predictions for kinetic friction across various material pairings. The results, illustrated in Figure~\ref{fig:embeddings_and_predictions-kinect}, demonstrate that the model effectively captures the underlying relationships in kinetic friction data, as evidenced by the coherent embeddings and accurate predictions. This confirms the model's capability to generalize across different material interactions within the kinetic friction regime. As compared to the results shown in Figure~\ref{fig:kinect-on-static}, the dedicated kinetic friction model exhibits improved performance, highlighting the importance of tailoring models to specific friction regimes for optimal accuracy.

Figure~\ref{fig:rrqr_proxies_kinect} illustrates representative proxy sets obtained via rank‑revealing QR (RRQR) applied to \(\mathbf{F}_\text{iso-k}\). The selected material indices (zero‑based) are: size 1: [44]; size 5: [4, 11, 16, 17, 44]; size 11: [0, 4, 8, 11, 13, 16, 17, 25, 30, 42, 44]; size 23: [0, 2, 4, 6, 8, 9, 11, 13, 16, 17, 18, 19, 20, 25, 26, 28, 29, 30, 31, 40, 42, 43, 44]. These selections closely mirror those identified for the static-friction regime, underscoring the consistency of proxy material selection across friction types. Figure~\ref{fig:mask_proxies_kinect} reports the results of the masking‑based optimization applied to the same matrix, which yields an optimal proxy set of two materials (indices 24 and 26, zero‑based), identical to the set identified for the static‑friction regime. Materials 24 and 26 emerge as critical proxies in both regimes, suggesting their pivotal role in capturing the essential frictional characteristics of the broader material set. This convergence reinforces the validity of our proxy selection methodology and its applicability to diverse frictional contexts.

\paragraph{Joint Embedding Model for Friction Analysis}
We further investigated the relationship between static and kinetic friction within the isotropic framework by jointly embedding the two types of friction data. As demonstrated in Figure~\ref{fig:joint_embedding_static_kinetic}, the joint embedding model effectively captures the shared characteristics of static and kinetic friction, as evidenced by the coherent embeddings and accurate predictions for both regimes. This approach allows us to analyze the interplay between static and kinetic friction more comprehensively, providing insights into their underlying mechanisms.

\begin{figure*}
    \centering
    \begin{subfigure}[t]{0.395\linewidth}
        \centering
        \includegraphics[width=\linewidth]{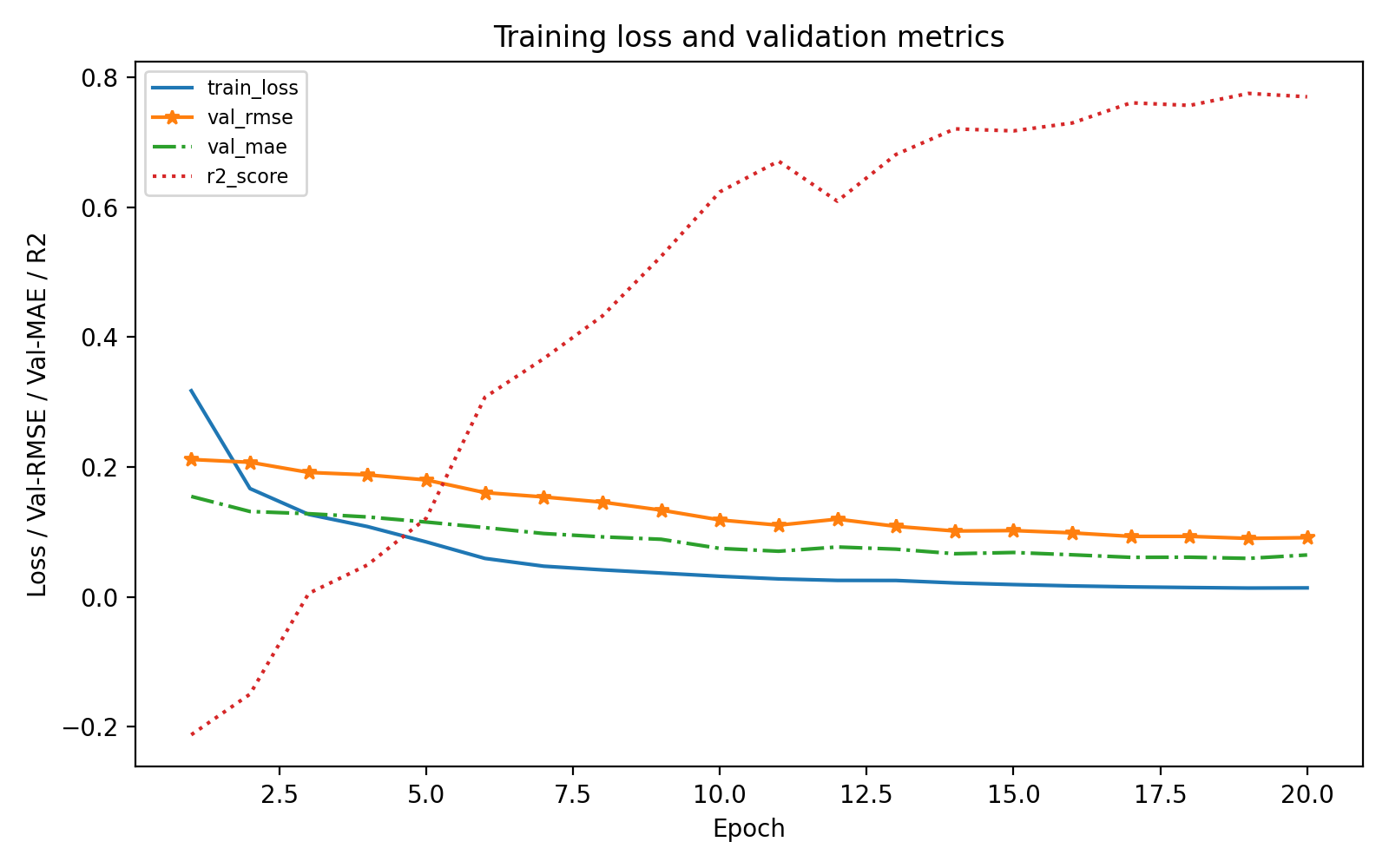}
        \subcaption{Training and validation loss curves}   
        \label{fig:true_vs_pred}
    \end{subfigure}
    \hfill
    \begin{subfigure}[t]{0.295\linewidth}
        \centering
        \includegraphics[width=\linewidth]{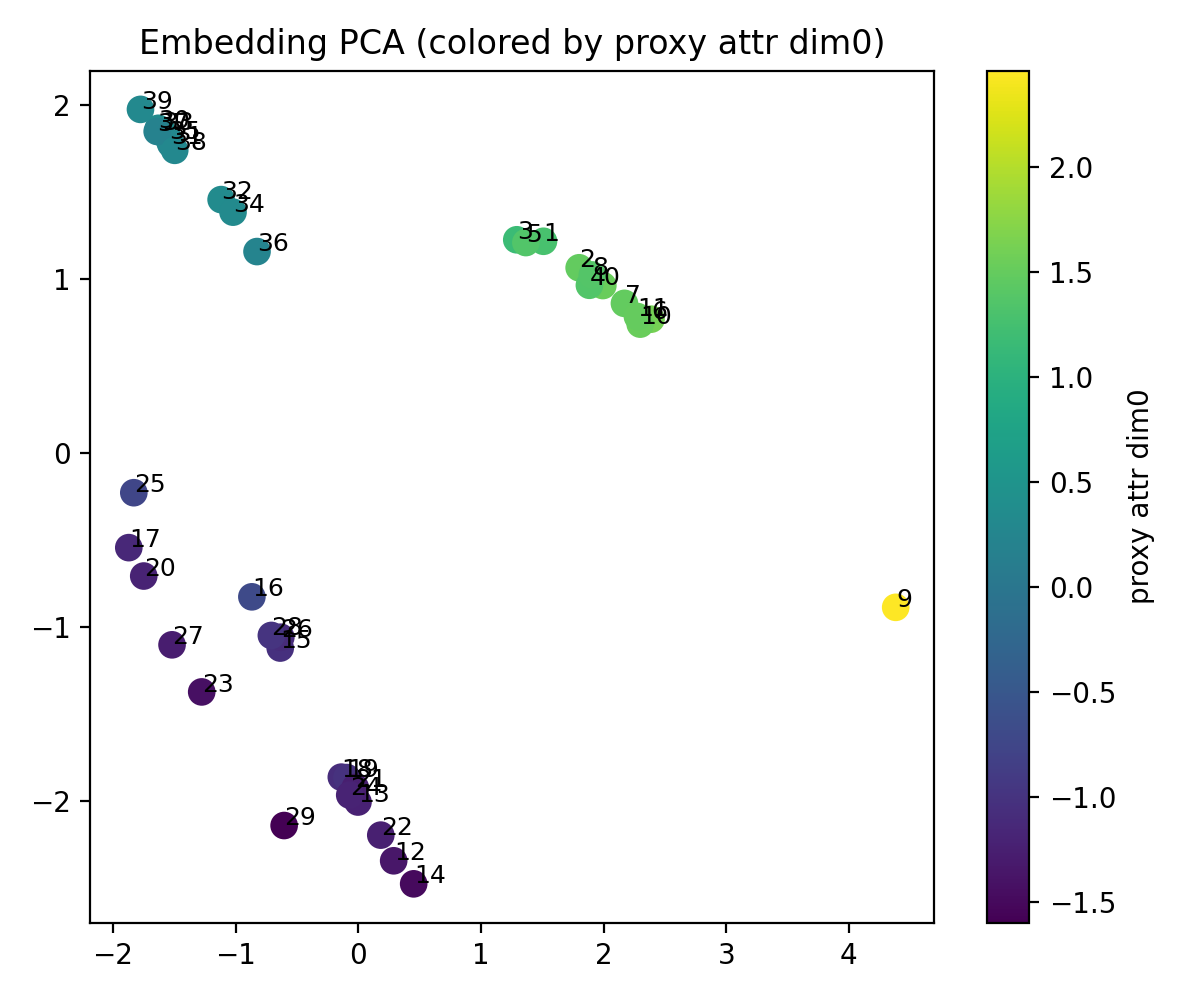}
        \subcaption{PCA embedding}
        \label{fig:emb_pca}
    \end{subfigure}
    \hfill
    \begin{subfigure}[t]{0.295\linewidth}
        \centering
        \includegraphics[width=\linewidth]{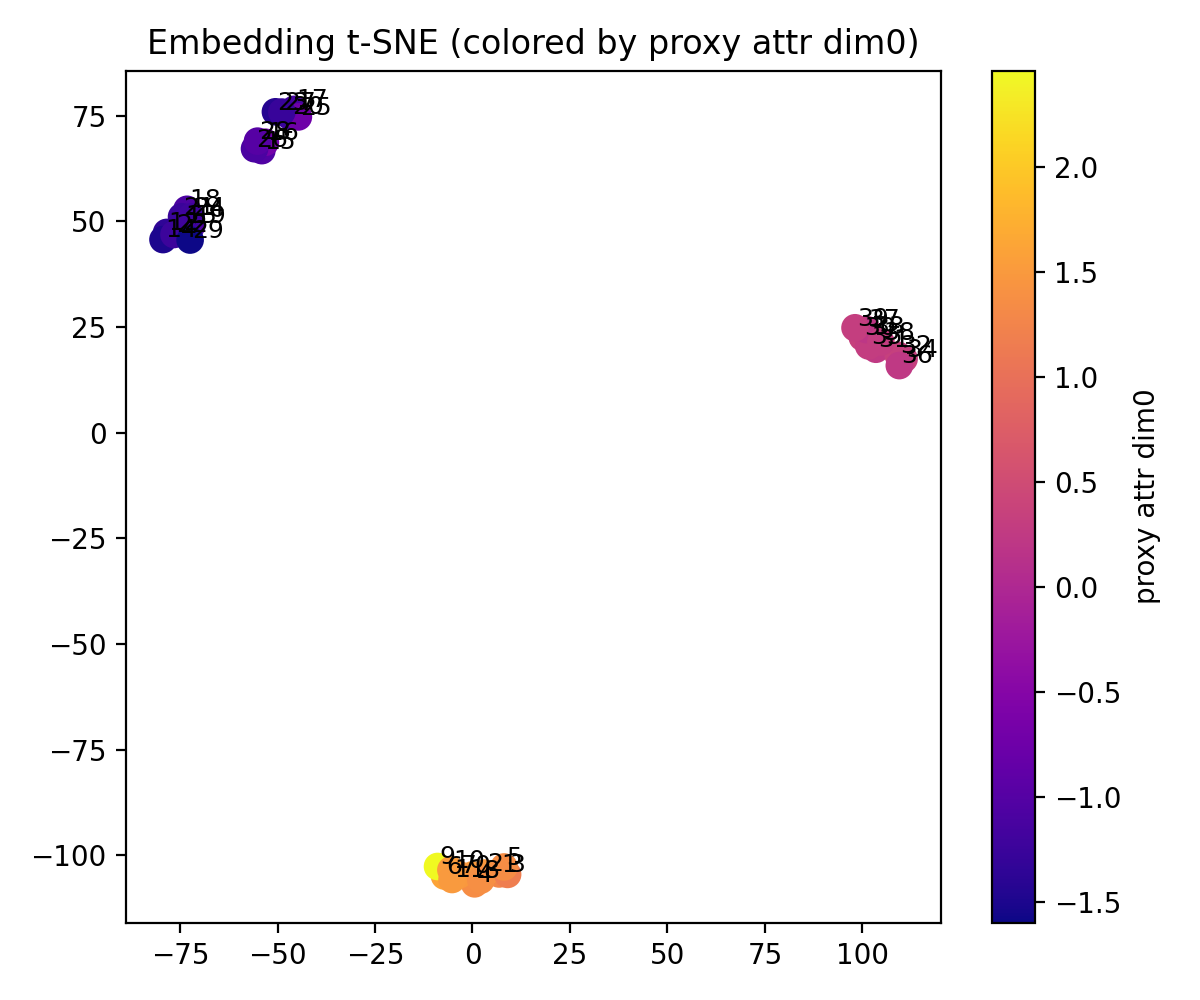}
        \subcaption{t-SNE embedding}
        \label{fig:emb_tsne}
    \end{subfigure}
    \hfill
    \begin{subfigure}[t]{0.195\linewidth}
        \centering
        \includegraphics[width=\linewidth]{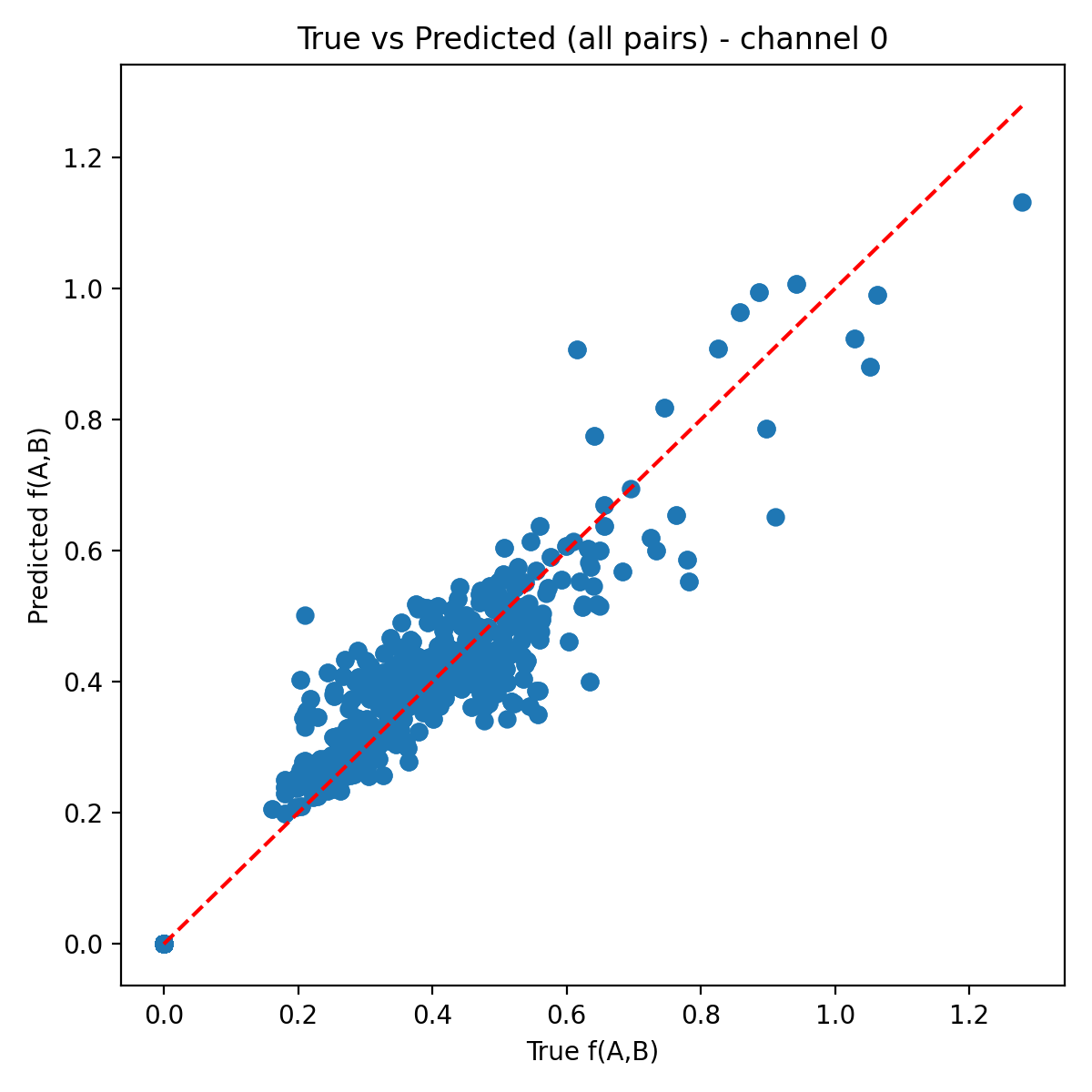}
        \subcaption{True vs. predicted: 0}
        \label{fig:emb_pca}
    \end{subfigure}
    \hfill
    \begin{subfigure}[t]{0.295\linewidth}
        \centering
        \includegraphics[width=\linewidth]{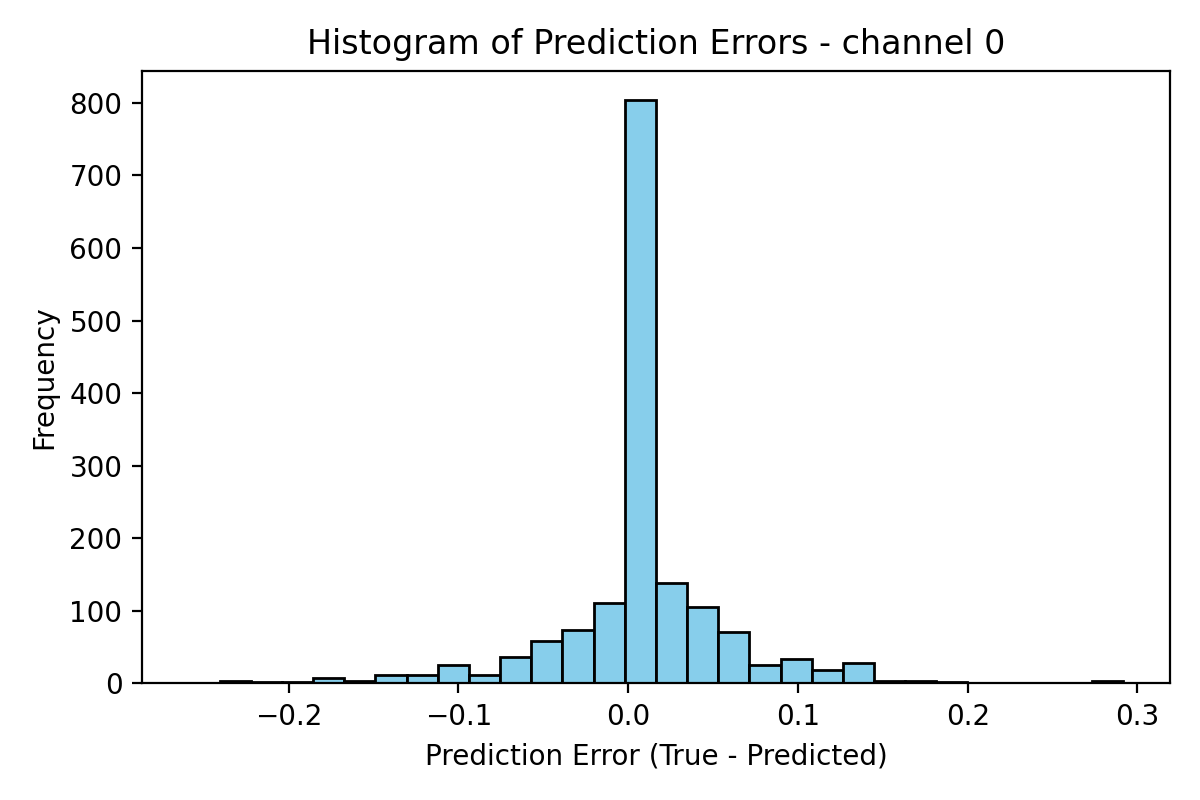}
        \subcaption{Error histogram: 0}
        \label{fig:emb_tsne}
    \end{subfigure}
    \hfill
    \begin{subfigure}[t]{0.195\linewidth}
        \centering
        \includegraphics[width=\linewidth]{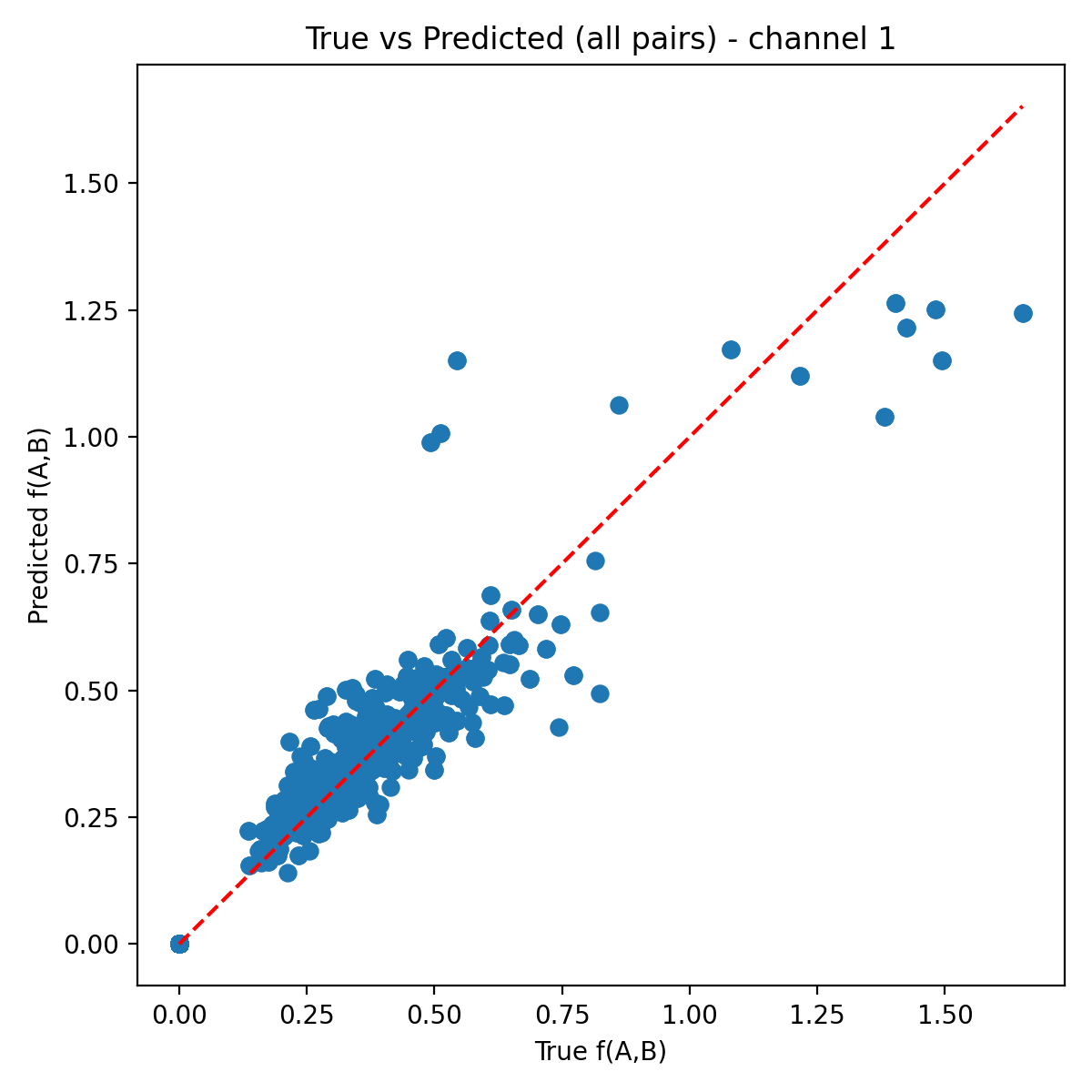}
        \subcaption{True vs. predicted: 1}
        \label{fig:true_vs_pred}
    \end{subfigure}
    \hfill
    \begin{subfigure}[t]{0.295\linewidth}
        \centering
        \includegraphics[width=\linewidth]{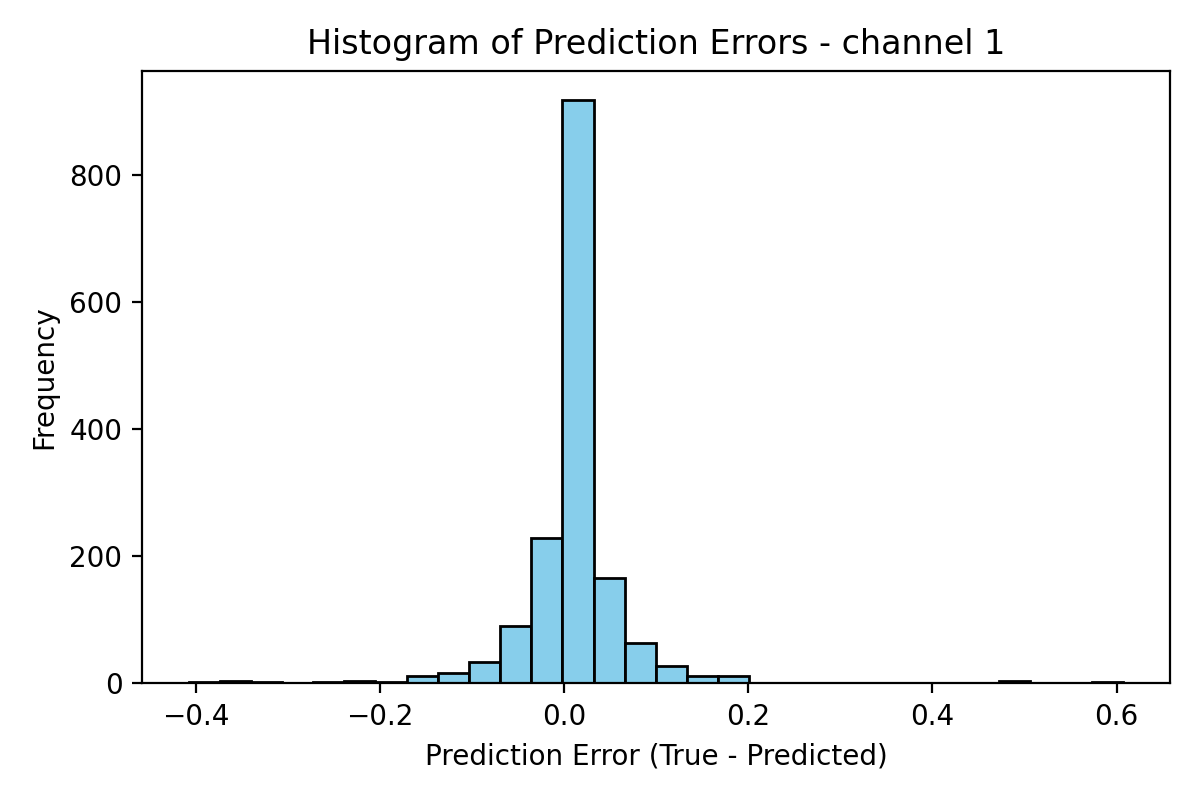}
        \subcaption{Error histogram: 1}
        \label{fig:error_histogram}
    \end{subfigure}
    \hfill
    \begin{subfigure}[t]{0.195\linewidth}
        \centering
        \includegraphics[width=\linewidth]{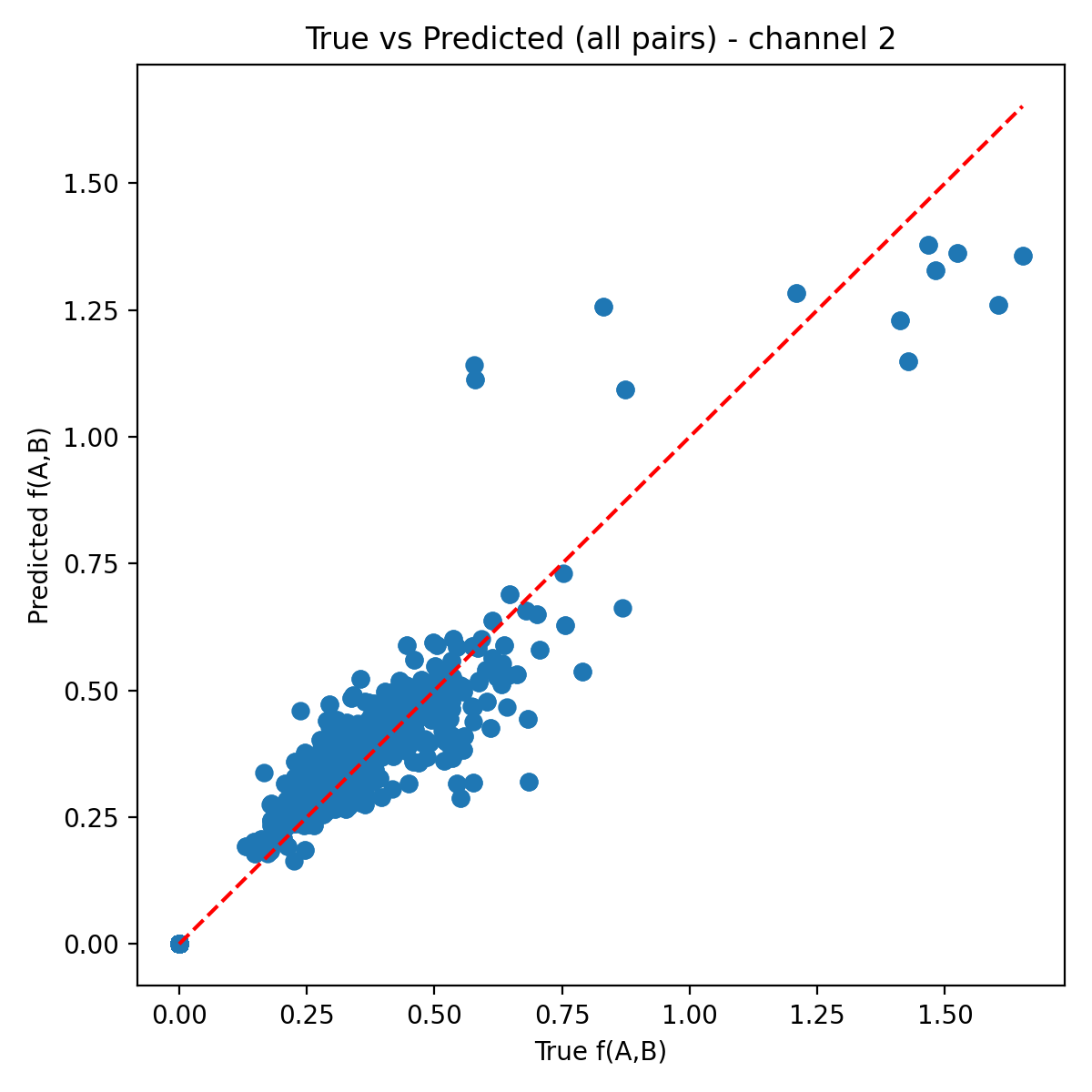}
        \subcaption{True vs. predicted: 2}
        \label{fig:emb_pca}
    \end{subfigure}
    \hfill
    \begin{subfigure}[t]{0.295\linewidth}
        \centering
        \includegraphics[width=\linewidth]{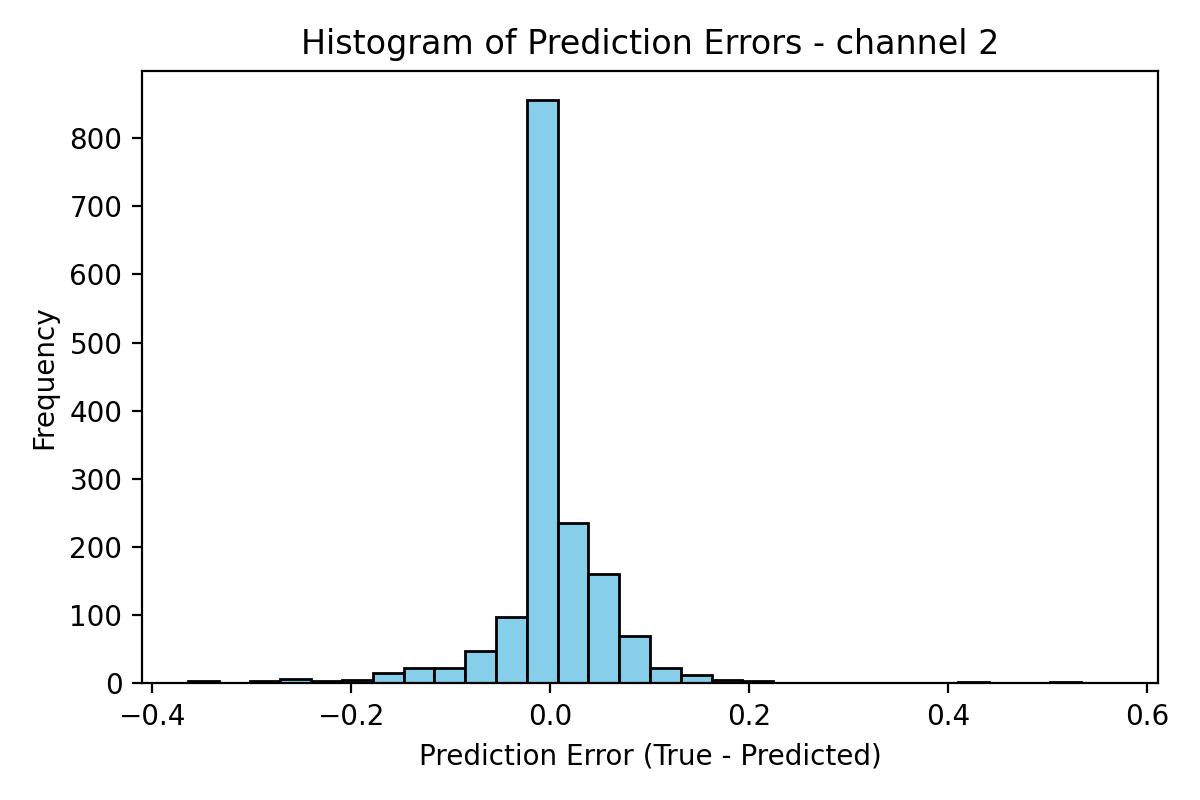}
        \subcaption{Error histogram: 2}
        \label{fig:emb_tsne}
    \end{subfigure}
    \hfill
    \begin{subfigure}[t]{0.195\linewidth}
        \centering
        \includegraphics[width=\linewidth]{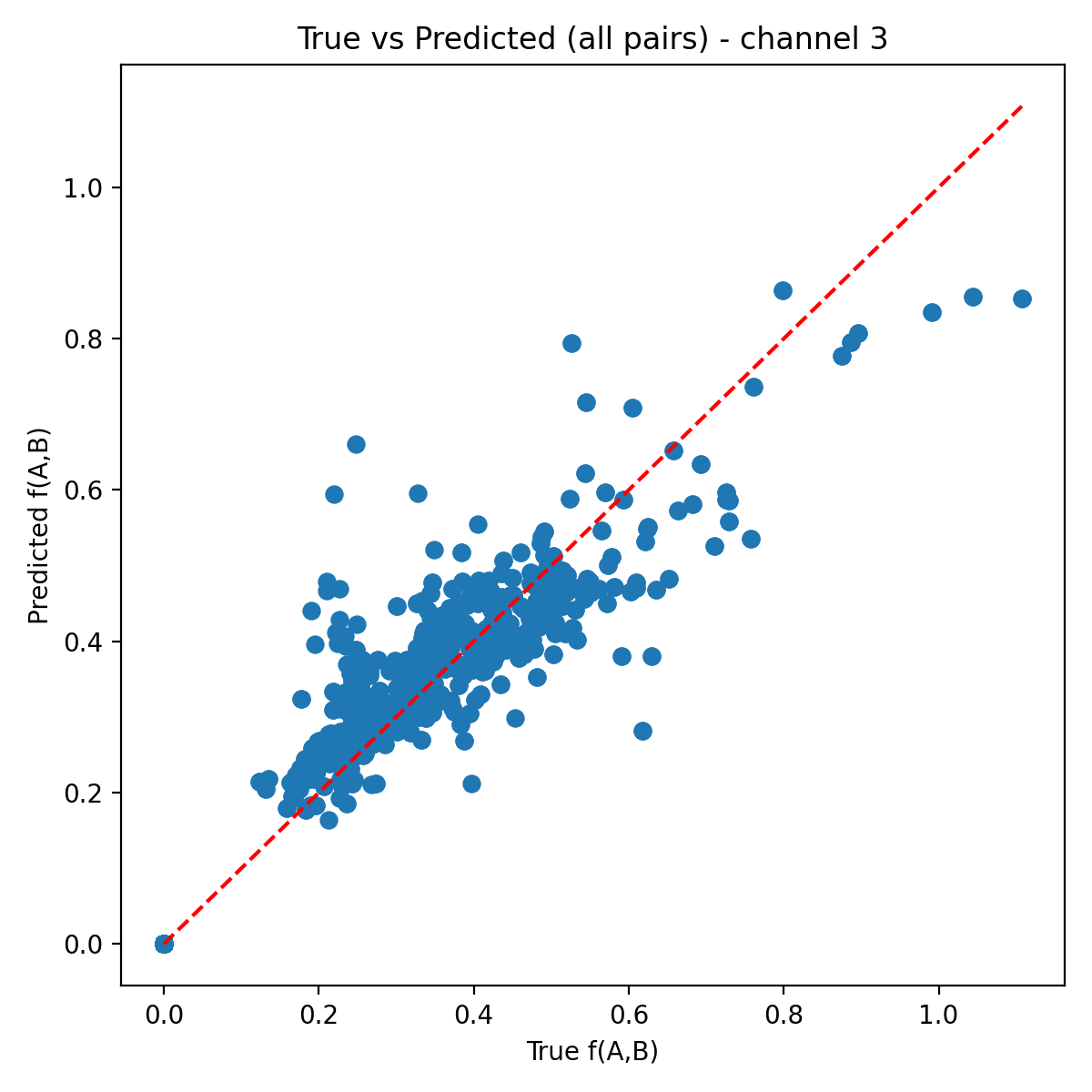}
        \subcaption{True vs. predicted: 3}
        \label{fig:true_vs_pred}
    \end{subfigure}
    \hfill
    \begin{subfigure}[t]{0.295\linewidth}
        \centering
        \includegraphics[width=\linewidth]{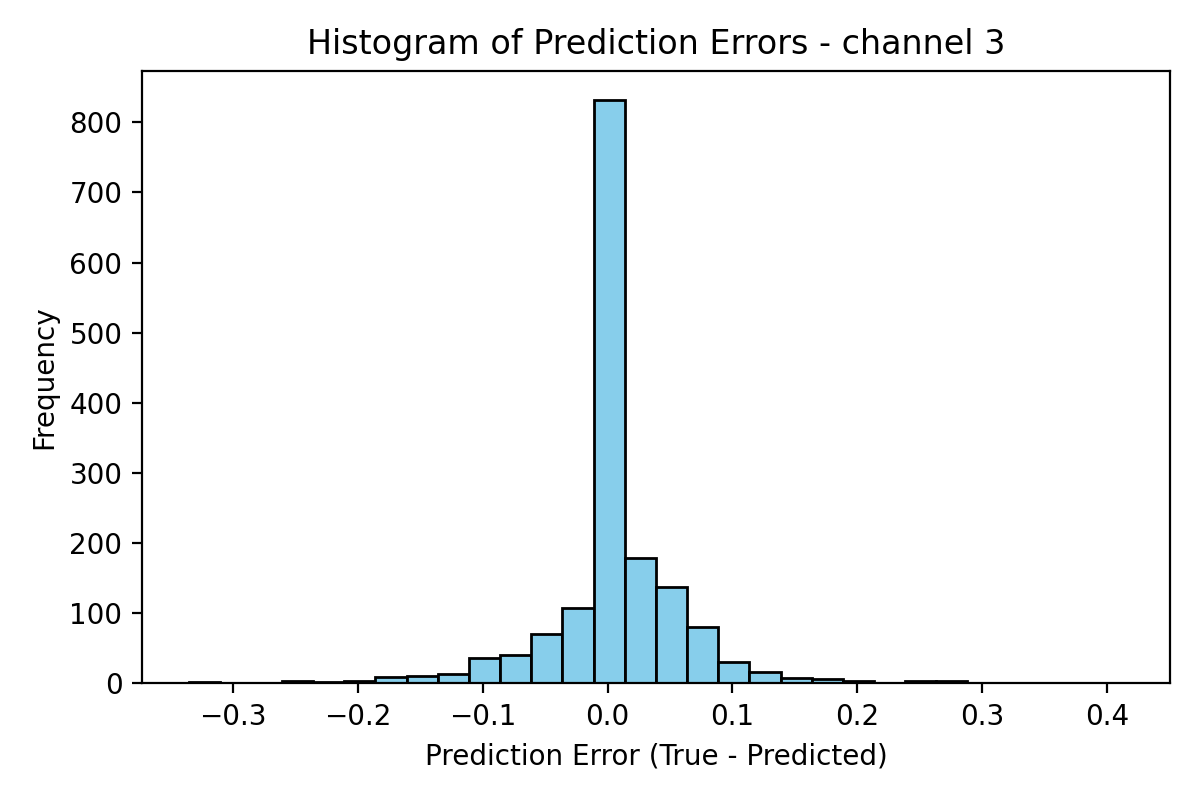}
        \subcaption{Error histogram: 3}
        \label{fig:error_histogram}
    \end{subfigure}
    \caption{Joint embedding model trained simultaneously on anisotropic friction matrices with each entry having four channels. (a) Training and validation loss curves exhibit stable, monotonic convergence without evidence of overfitting. (b,c) PCA and t-SNE projections of the shared material embeddings closely replicate the structural organization obtained from regime-specific models. True–vs.–predicted scatter and residual histogram show high linear correlation with tightly concentrated errors. Collectively, the results indicate that a unified latent representation captures direction-dependent frictional variability with high accuracy.}
    \label{fig:anisotropic_embeddings_and_predictions}
\end{figure*}

\subsection{Anisotropic Friction}

For the anisotropic regime we employ the incomplete block matrix \(\mathbf{F}_\text{aniso}\) described in Section~\ref{sec:dataset_details}. For model training we extract the submatrix
\begin{equation}
\mathbf{F}'_\text{aniso}=\begin{bmatrix}
    \mathbf{F}_\text{knit} & \mathbf{Nan} & \times\\
    \times & \mathbf{F}_\text{woven} & \times\\
    \mathbf{F}_\text{non-fabric-knit} & \mathbf{F}_\text{non-fabric-woven} & \mathbf{Nan}\\
\end{bmatrix}
\label{eq:anisotropic}
\end{equation}
and reserve \(\mathbf{F}_\text{knit-woven}\in\mathbb{R}^{3\times18}\) for out‑of‑distribution evaluation. Although direct knit–woven interaction entries are absent during training, implicit cross‑class coupling arises via their respective interactions with the shared non‑fabric substrates, furnishing a transferable latent representation across fabric classes.

\paragraph{Training Evaluation}

Model training on anisotropic friction data is complicated by direction‑dependent interaction channels and systematic sparsity in the block matrix. Nevertheless, the learned embeddings exhibit well‑separated, coherent clusters (Figure~\ref{fig:anisotropic_embeddings_and_predictions}), evidencing effective latent structuring across fabric classes. Predictive performance remains high, indicating that the model captures salient orientation‑resolved coupling mechanisms despite incompleteness in cross‑class measurements.



\begin{figure*}
    \centering
    \begin{subfigure}[t]{0.195\linewidth}
        \centering
        \includegraphics[width=\linewidth]{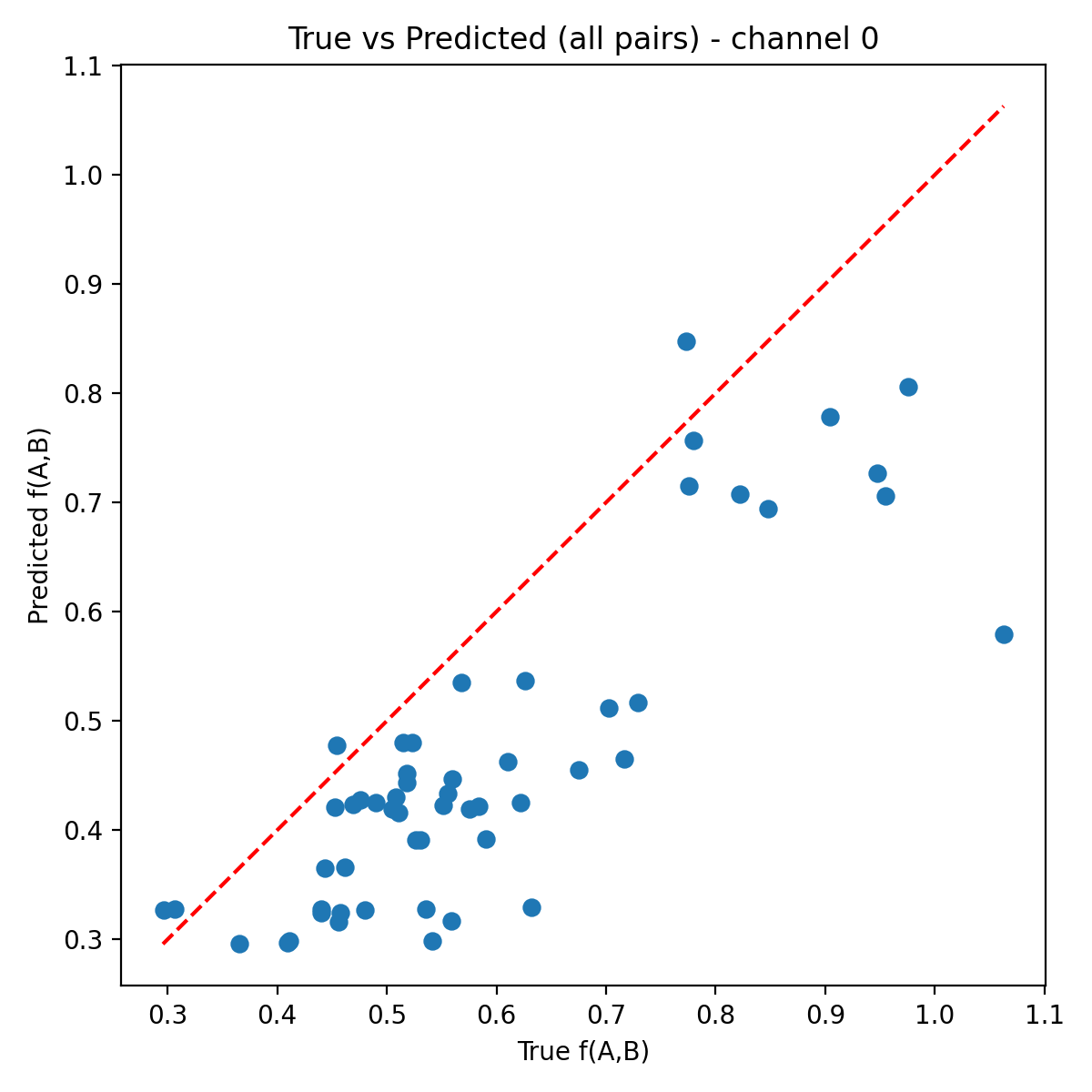}
        \subcaption{True vs. predicted: 0}
        \label{fig:emb_pca}
    \end{subfigure}
    \hfill
    \begin{subfigure}[t]{0.295\linewidth}
        \centering
        \includegraphics[width=\linewidth]{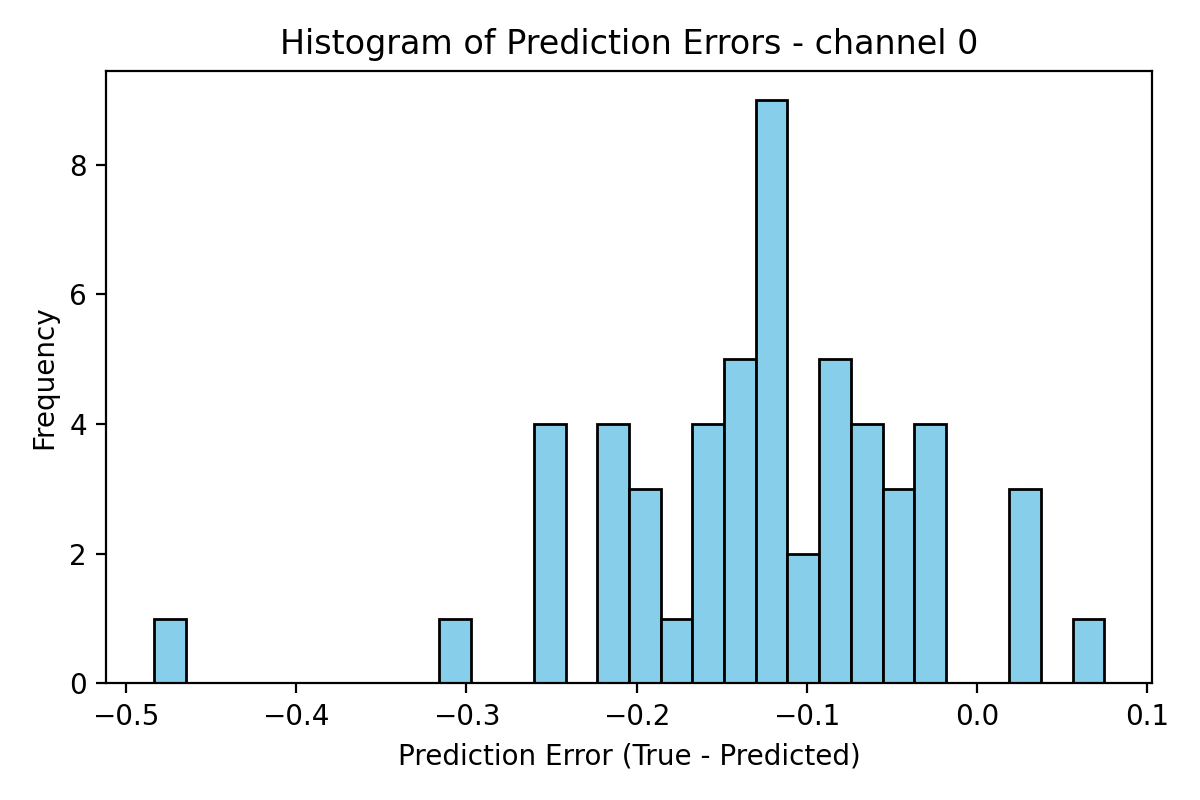}
        \subcaption{Error histogram: 0}
        \label{fig:emb_tsne}
    \end{subfigure}
    \hfill
    \begin{subfigure}[t]{0.195\linewidth}
        \centering
        \includegraphics[width=\linewidth]{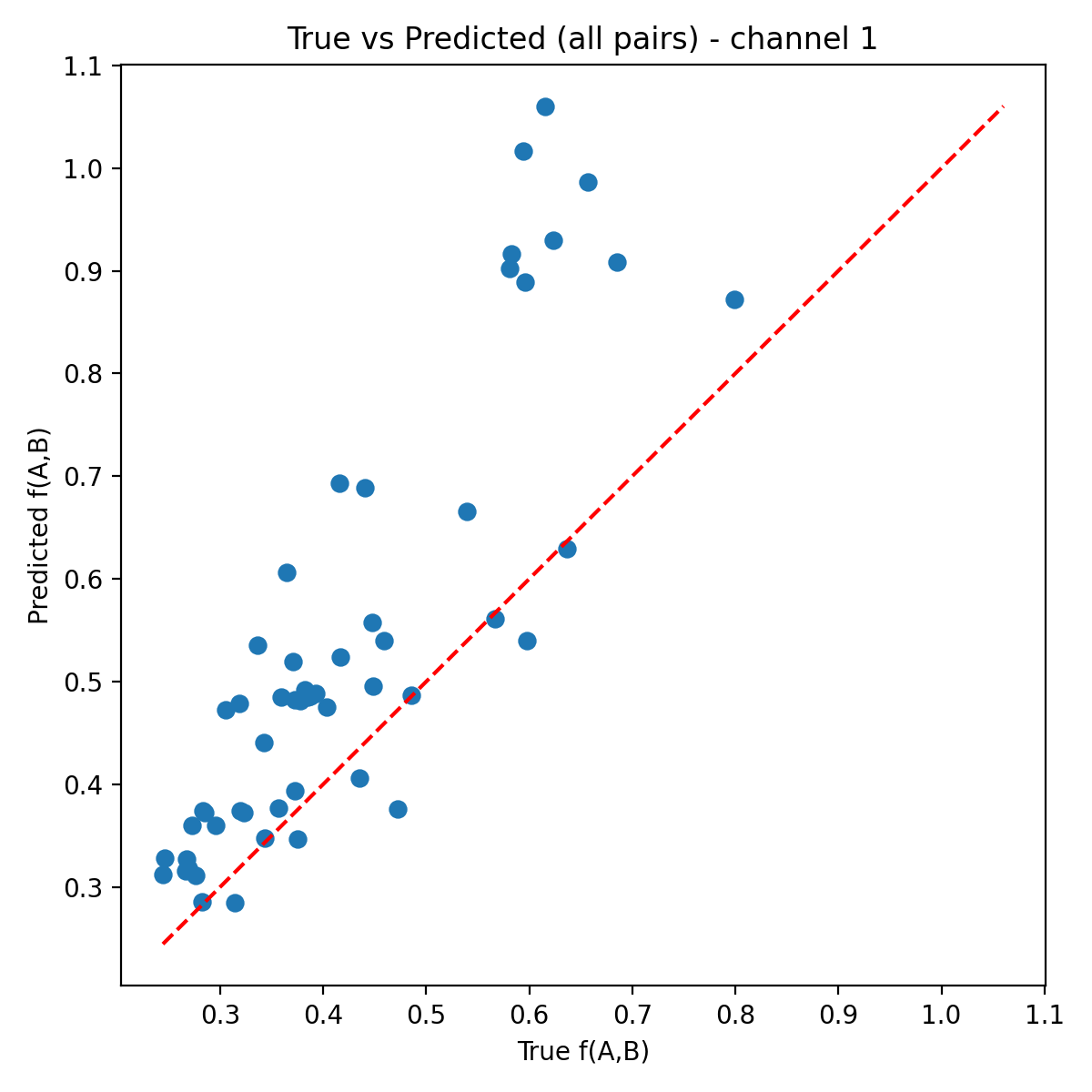}
        \subcaption{True vs. predicted: 1}
        \label{fig:true_vs_pred}
    \end{subfigure}
    \hfill
    \begin{subfigure}[t]{0.295\linewidth}
        \centering
        \includegraphics[width=\linewidth]{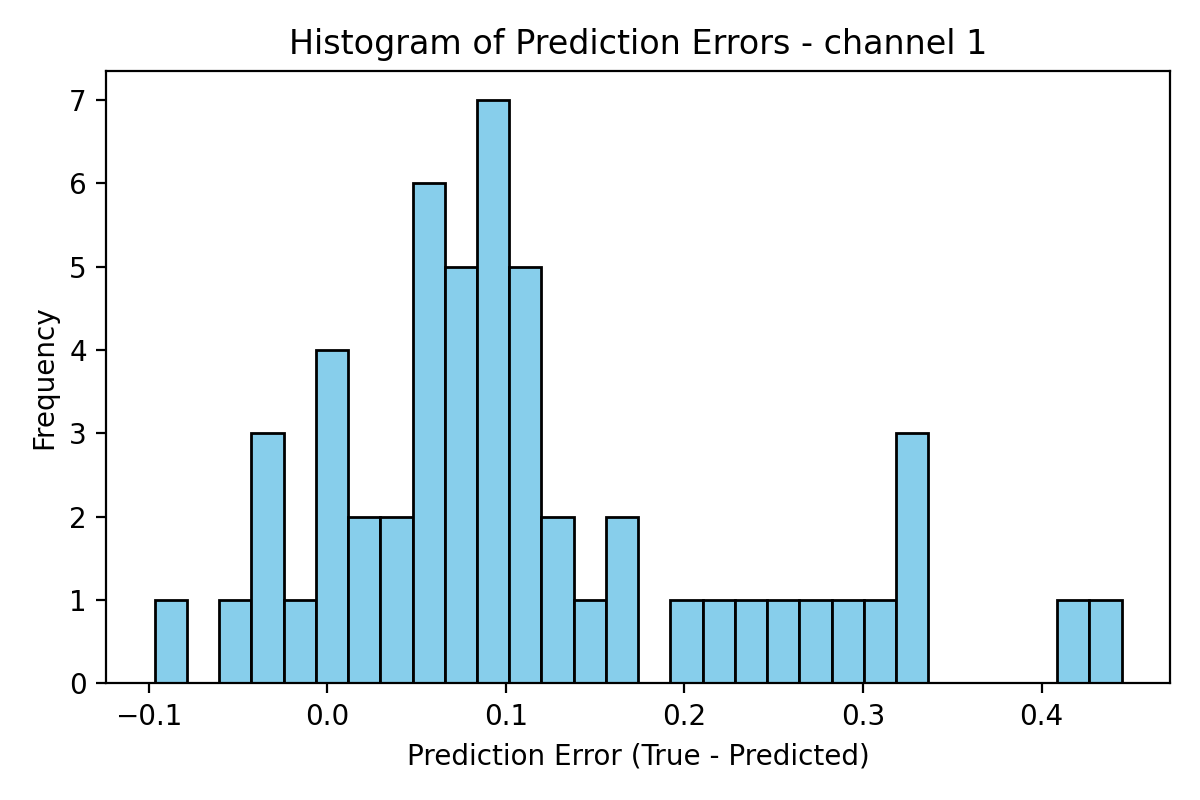}
        \subcaption{Error histogram: 1}
        \label{fig:error_histogram}
    \end{subfigure}
    \hfill
    \begin{subfigure}[t]{0.195\linewidth}
        \centering
        \includegraphics[width=\linewidth]{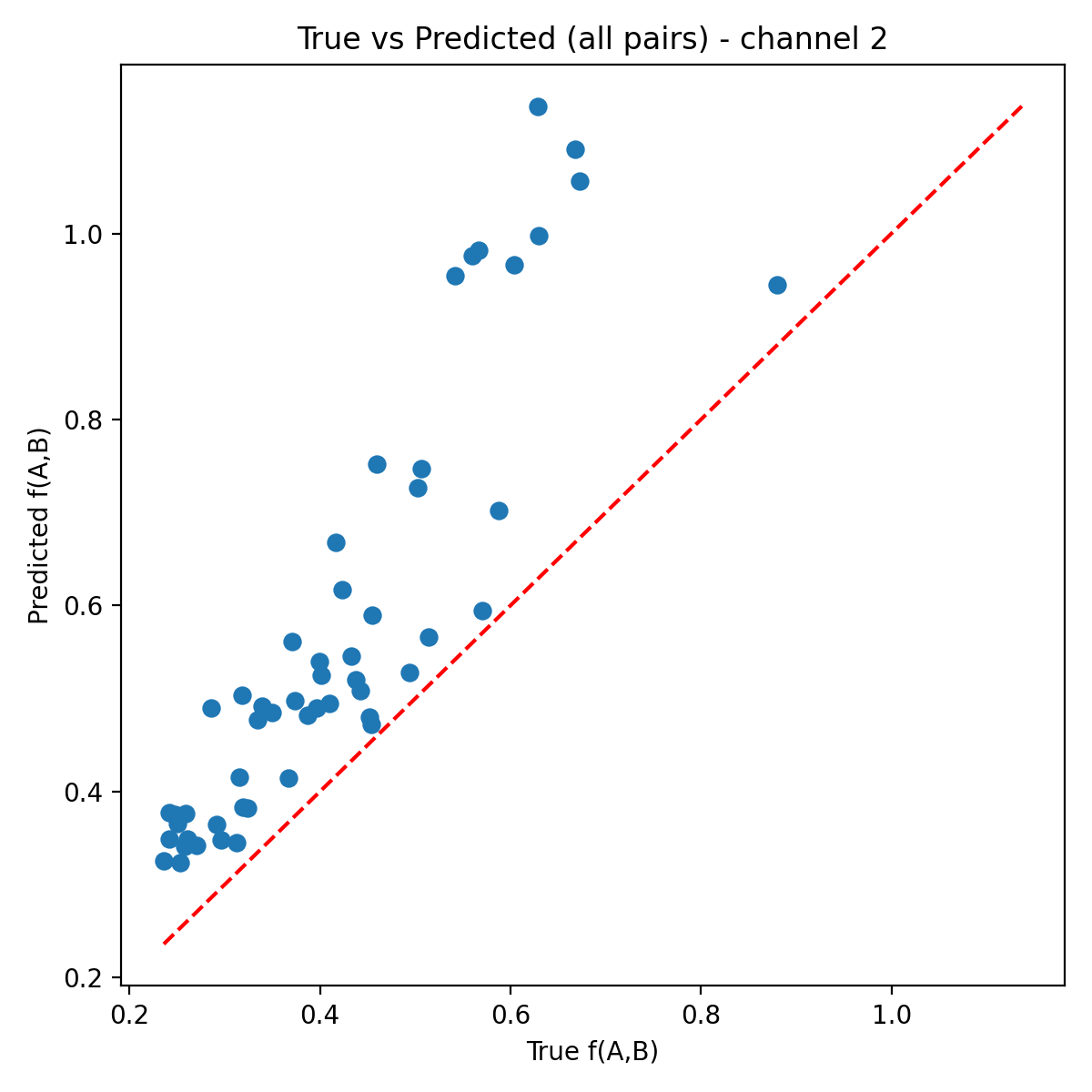}
        \subcaption{True vs. predicted: 2}
        \label{fig:emb_pca}
    \end{subfigure}
    \hfill
    \begin{subfigure}[t]{0.295\linewidth}
        \centering
        \includegraphics[width=\linewidth]{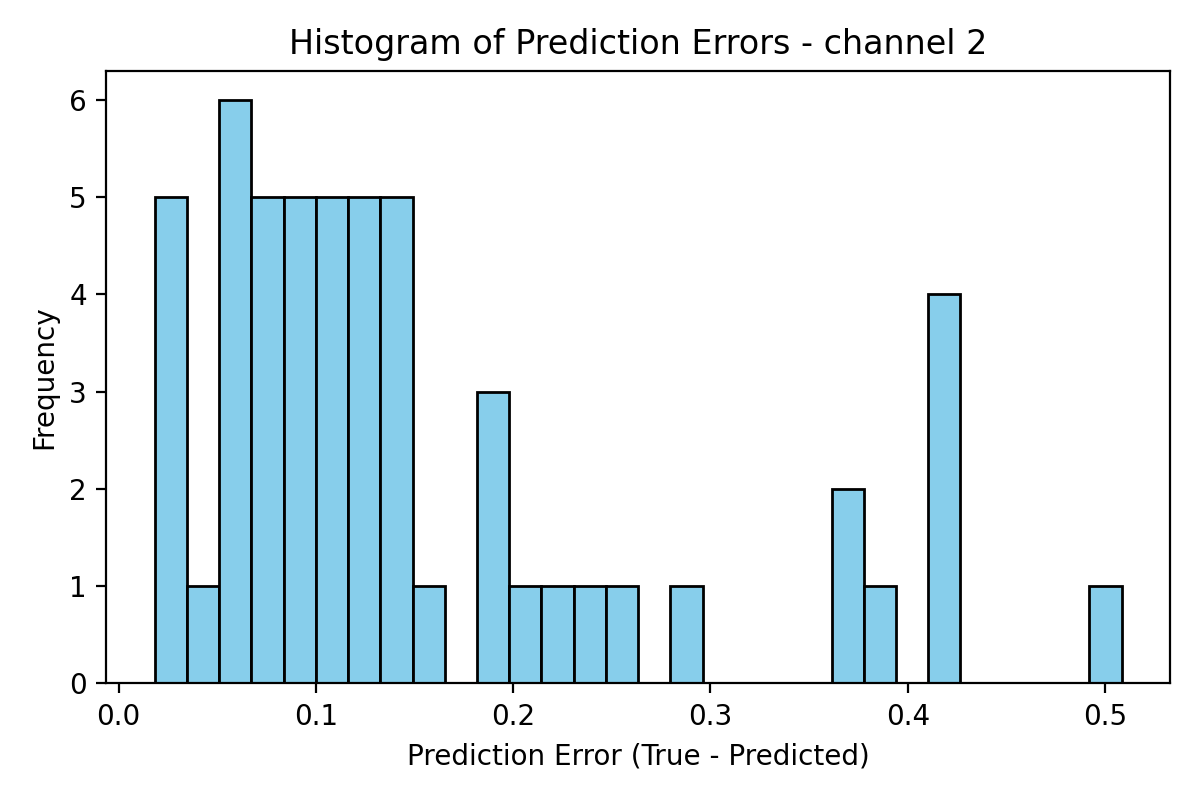}
        \subcaption{Error histogram: 2}
        \label{fig:emb_tsne}
    \end{subfigure}
    \hfill
    \begin{subfigure}[t]{0.195\linewidth}
        \centering
        \includegraphics[width=\linewidth]{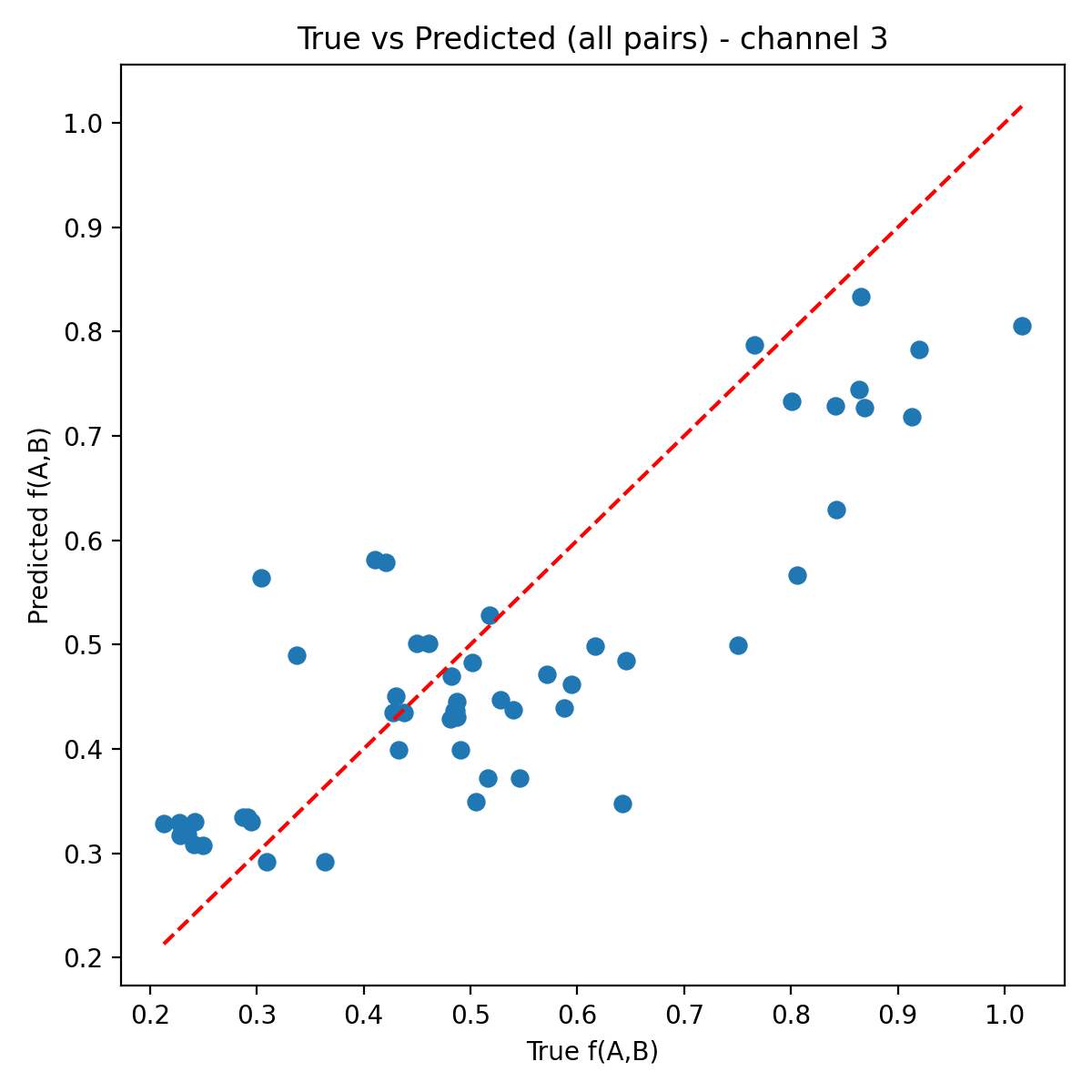}
        \subcaption{True vs. predicted: 3}
        \label{fig:true_vs_pred}
    \end{subfigure}
    \hfill
    \begin{subfigure}[t]{0.295\linewidth}
        \centering
        \includegraphics[width=\linewidth]{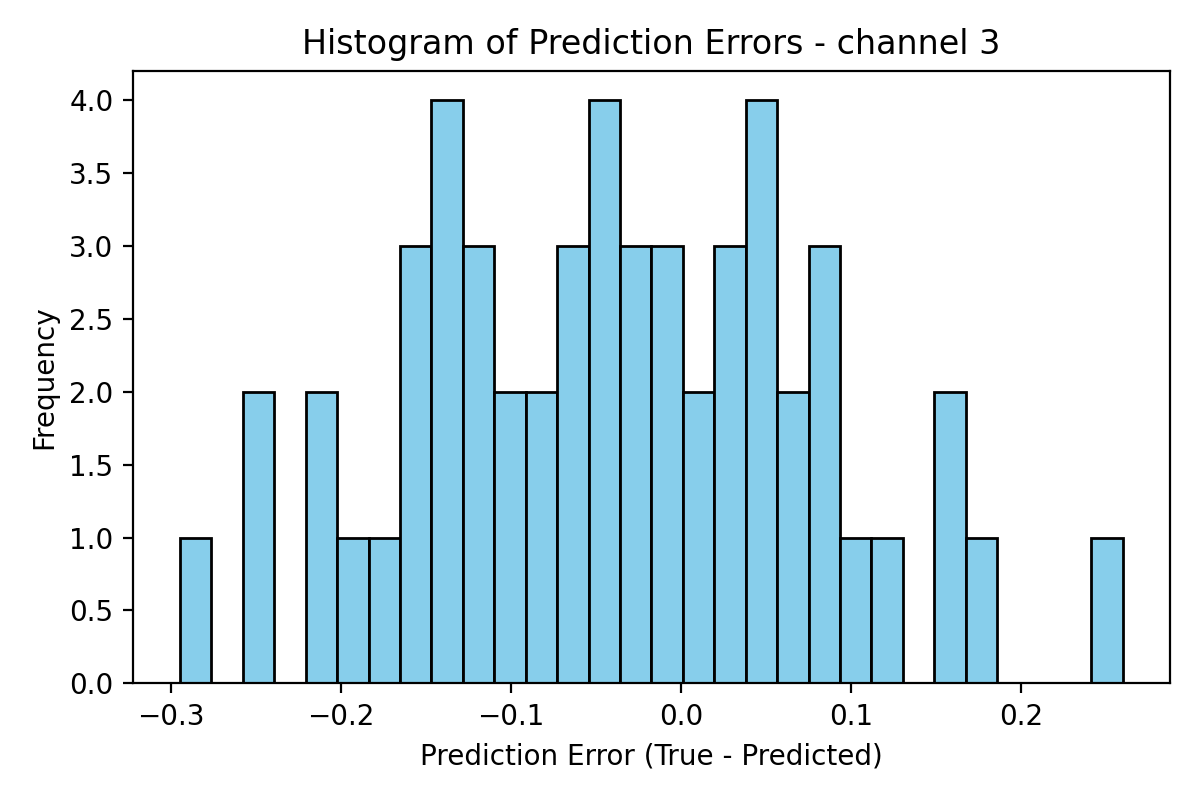}
        \subcaption{Error histogram: 3}
        \label{fig:error_histogram}
    \end{subfigure}
    \caption{Out-of-distribution evaluation on the withheld knit–woven submatrix shows reduced predictive fidelity relative to in-class pairings, attributable to the absence of direct supervisory signals and residual sparsity in the non‑fabric coupling blocks (incomplete \(\mathbf{F}_{\text{non-fabric}}\)). Across the four orientation channels, true–vs.–predicted scatter plots and associated residual histograms exhibit broadened error distributions, evidencing the difficulty of inferring anisotropic knit–woven frictional mechanics without explicit measurements.}
    \label{fig:anisotropic_unseen_testing}
\end{figure*}

\paragraph{Testing for Knit and Woven Fabrics}

We examined anisotropic knit–woven interactions by withholding the direct knit–woven submatrix \(\mathbf{F}_{\text{knit-woven}}\) during training (Eq.~\ref{eq:anisotropic}) to form an out-of-distribution evaluation set. Baseline results (Fig.~\ref{fig:anisotropic_unseen_testing}) show reduced accuracy on knit–woven pairs, attributable to missing direct supervision and residual sparsity in the non‑fabric coupling blocks (incomplete \(\mathbf{F}_{\text{non-fabric}}\)). The present dataset omits most knit–woven interactions and a substantial portion of non‑fabric measurements, resulting in insufficient substrate‑mediated coupling between the two fabric classes. Moreover, it spans only a small fraction of the combinatorial space of admissible material pairings, which constrains generalization to unseen knit–woven pairs. Future work should expand direct knit–woven coverage, densify the non‑fabric coupling blocks, and broaden the material taxonomy to improve model robustness and out‑of‑distribution performance.